\documentclass{article}

\usepackage[frozencache=true, cachedir=minted-cache]{minted}

    \PassOptionsToPackage{numbers, compress}{natbib}

\usepackage[preprint]{neurips_2024}

\usepackage[utf8]{inputenc} 
\usepackage[T1]{fontenc}    
\usepackage{hyperref}       
\usepackage{url}            
\usepackage{booktabs}       
\usepackage{amsfonts}       
\usepackage{nicefrac}       
\usepackage{microtype}      
\usepackage{xcolor}         
\usepackage{amsmath}
\usepackage{amssymb}
\usepackage{caption}
\usepackage{graphicx}
\usepackage{dsfont}
\usepackage{wrapfig} 
\usepackage[ruled, noend]{algorithm2e}

\usepackage{tcolorbox}
\tcbuselibrary{minted,breakable,xparse,skins}
\definecolor{bg}{gray}{0.95}
\DeclareTCBListing{mintedbox}{O{}m m !O{}}{%
  breakable=true,
  listing engine=minted,
  listing only,
  minted language=#2,
  title={#3},
  minted style=default,
  minted options={%
    linenos,
    gobble=0,
    breaklines=true,
    breakafter=,,
    fontsize=\small,
    numbersep=8pt,
    #1},
  boxsep=0pt,
  left skip=0pt,
  right skip=0pt,
  left=25pt,
  right=0pt,
  top=3pt,
  bottom=3pt,
  arc=5pt,
  leftrule=0pt,
  rightrule=0pt,
  bottomrule=2pt,
  toprule=2pt,
  colback=bg,
  colframe=blue!70,
  enhanced,
  overlay={%
    \begin{tcbclipinterior}
    \fill[blue!20!white] (frame.south west) rectangle ([xshift=20pt]frame.north west);
    \end{tcbclipinterior}},
  #4}

\SetCommentSty{mycommfont}

\def\rd{{\textnormal{d}}}
\definecolor{lightblue}{HTML}{AFEEEE}
\newcommand{\better}[1]{\colorbox{lightblue}{#1}}

\title{\looseness=-2 Scaling Up Diffusion and Flow-based XGBoost Models}

\author{%
  Jesse C. Cresswell \\
  Layer 6 AI, Toronto, Canada\\
  \texttt{jesse@layer6.ai} 
  \And
  Taewoo Kim \\
  Layer 6 AI, Toronto, Canada\\
  \texttt{taewoo@layer6.ai} 
}

\begin{document}

\maketitle

\begin{abstract}
\vspace{-8pt}
Novel machine learning methods for tabular data generation are often developed on small datasets which do not match the scale required for scientific applications. We investigate a recent proposal to use XGBoost as the function approximator in diffusion and flow-matching models on tabular data, which proved to be extremely memory intensive, even on tiny datasets. In this work, we conduct a critical analysis of the existing implementation from an engineering perspective, and show that these limitations are not fundamental to the method; with better implementation it can be scaled to datasets 370$\times$ larger than previously used. Our efficient implementation also unlocks scaling models to much larger sizes which we show directly leads to improved performance on benchmark tasks. We also propose algorithmic improvements that can further benefit resource usage and model performance, including multi-output trees which are well-suited to generative modeling. Finally, we present results on large-scale scientific datasets derived from experimental particle physics as part of the Fast Calorimeter Simulation Challenge. Code is available at \href{https://github.com/layer6ai-labs/calo-forest}{\tt{github.com/layer6ai-labs/calo-forest}}.
\vspace{-8pt}
\end{abstract}

\section{Introduction}
\label{sec:introduction}
\vspace{-2pt}

\begin{wrapfigure}[15
]{r}{0.4\textwidth}
\vspace{-20pt}
    \centering
    \includegraphics[width=0.38\textwidth, trim={0 0 0 25}, clip]{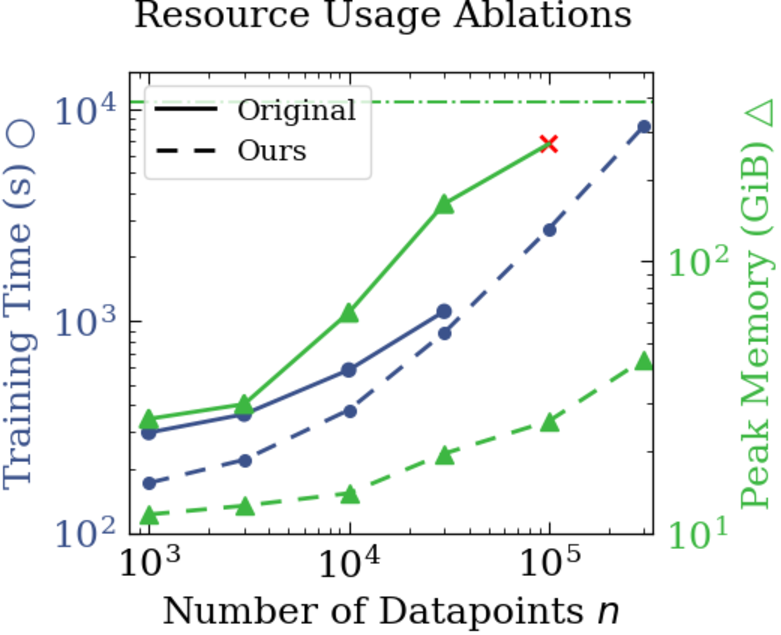}
    \vspace{-5pt}
    \caption{Comparison of training time and memory usage between the original implementation and ours. The \textcolor{red}{$\times$} indicates job failure, and the horizontal line indicates the maximum system memory.}
    \label{fig:single_usage}
\end{wrapfigure}
\normalsize
The design of neural network (NN) architectures with appropriate inductive biases for a given data modality has lead to incredible breakthroughs on text \cite{vaswani2017attention}, audio \cite{hochreiter1997lstm}, image \cite{krizhevsky2012alexnet, he2016resnet}, graph \cite{kipf2016semi}, and many other modalities. However, tabular data stands out in that tree-based architectures still often outperform NNs \cite{grinsztajn2022, mcelfresh2023neural}. This can largely be attributed to the lack of exploitable structure in tabular data that NN design typically relies on.

\vspace{-2pt}
Despite the success of boosted tree architectures like XGBoost \cite{chen2016xgb} on \emph{discriminative} tasks, they are rarely used for \emph{generative} modeling \cite{nock22, nock2023generative}. This is surprising, as XGBoost brings several other meaningful advantages: XGBoost does not require significant data pre-processing (NNs are highly sensitive to data scale and distribution); XGBoost can operate on data that contains null values (NNs require null values to be imputed or entire columns dropped); XGBoost can be trained efficiently on CPU or GPU (NNs usually require GPU training); and XGBoost has superior explainability (Shapley values \cite{shapley1951} are intractable for large NNs, but the TreeSHAP algorithm makes them tractable for trees \cite{lundberg2017, lundberg2018}). Similar to MLP networks, XGBoost is a universal function approximator \cite{friedman2000additive, friedman2001greedy} and can be used to fit any function, at least in principle.

\vspace{-2pt}
Recently, \citet{jolicoeur2023generating} proposed a method for training diffusion \cite{song2021scorebased} and flow-matching \cite{liu2023flow, albergo2023building, lipman2023flow} generative models on tabular data by using XGBoost as the function approximator for a learnable vector field. Given the discussion above, this idea shows great promise. However, the original implementation was only benchmarked on small datasets (up to 11,000 datapoints with 16 features), and proved to be incredibly memory intensive (Figure \ref{fig:single_usage} solid lines). Important scientific and industrial applications of tabular generative modeling typically operate at much larger scales, such as the Fast Calorimeter Simulation Challenge for generative modeling of particle physics interactions \cite{calochallenge} with tabular datasets 370$\times$ larger than those used in \cite{jolicoeur2023generating}.

\vspace{-2pt}
In this work we conduct a deep and critical analysis of the implementation of diffusion and flow-matching models backed by XGBoost using engineering best-practices, and provide a new implementation re-engineered from the ground-up. Our implementation reduces CPU memory requirements from roughly quadratic in dataset size to linear and greatly reduces the memory overhead (Figure \ref{fig:single_usage} dashed lines), showing that the algorithm is much more broadly applicable than previously thought. Our efficient implementation allows models to be scaled up in size which directly leads to model performance gains. In addition, we demonstrate novel techniques to improve generative quality, including the use of multi-output trees for generative modeling, which can more accurately represent high-dimensional joint distributions. Finally, we demonstrate that the methods are feasible in practice by applying them to large-scale scientific datasets.

\section{Background}
\label{sec:background}
\vspace{-4pt}
First, we briefly review diffusion and flow matching models and then describe how to train them with XGBoost function approximators.\footnote{We refer to XGBoost throughout, but other tree-based regressors could be used. Our implementation takes advantage of XGBoost features that are currently unavailable in other gradient-boosted decision tree libraries.} For a more extensive description of diffusion and flow matching models, see \cite{loaiza2024deep}. Finally, we introduce the application of interest from experimental particle physics.

\vspace{-2pt}
\subsection{Diffusion Models}
\vspace{-2pt}
Score-based diffusion models \cite{song2021scorebased} corrupt data $\mathbf{x}_0\sim p_0$ by progressively adding noise as $t\in[0,1]$ increases, in a process modeled by a stochastic differential equation (SDE). Reversing this process enables the generation of data from pure noise. The reverse SDE involves a novel term, the score function $\nabla_{\mathbf{x}_t} \log p_t(\mathbf{x}_t)$, where $p_t$ is the density corresponding to data at noise level $t$. Since the data density $p_0$ is not known in closed form, neither is $p_t$, so $\nabla_{\mathbf{x}_t} \log p_t(\mathbf{x}_t)$ cannot be directly computed. However, it can be estimated using a denoising score matching approach \cite{hyvarinen2005estimation, vincent2011connection} with the loss
\vspace{-1pt}
\begin{equation}
\label{eq:sm_loss}
    L_{\mathrm{SM}}(\theta)=\mathbb{E}_{t\sim \mathcal{U}(0,1)} w(t) \mathbb{E}_{\mathbf{x}_0\sim p_0, \mathbf{x}_t\sim p_t(\cdot\mid\mathbf{x}_0)} \Vert \mathbf{s}_\theta(\mathbf{x}_t, t) - \nabla_{\mathbf{x}_t} \log p_t(\mathbf{x}_t\mid\mathbf{x}_0)\Vert_2^2.
\end{equation}
Here, $\mathbf{s}_\theta$ is a parameterized vector field that is directly regressed on the score function, while $w(t)$ is a positive-valued weighting function that can be chosen freely. In words, for a $t$ sampled uniformly, and $\mathbf{x}_0$ drawn from the data distribution, we sample $\mathbf{x}_t$ from $p_t(\cdot\mid\mathbf{x}_0)$ which is Gaussian \cite{song2021scorebased},
\vspace{-1pt}
\begin{equation}
\label{eq:diffusion_conditional}
    p_t(\mathbf{x}_t\mid \mathbf{x}_0) = \mathcal{N}\Big(\mathbf{x}_t; \sqrt{1 - \sigma_t^2} \mathbf{x}_0, \sigma_t^2 \mathbf{I}_D\Big),
\end{equation}
since $p_t$ is the result of a linear SDE starting with a point mass at $\mathbf{x}_0$. The standard deviation $\sigma_t$ depends on the details of the forward SDE. For generation, $\mathbf{s}_\theta$ replaces the score function in the reverse SDE which is then solved numerically.

\vspace{-4pt}
\subsection{Flow Matching}
\vspace{-2pt}
Like continuous normalizing flows \cite{chen2018neuralode}, flow matching interpolates probability densities $p_t$ for $t{\in}[0,1]$  \cite{liu2023flow, albergo2023building, lipman2023flow}. We consider $p_0$ as the data, and select a simple prior $p_1{=}\mathcal{N}( \mathbf{x}_1 \mid 0, \sigma^2)$. The interpolation is determined by a vector field at each time $\boldsymbol{\mu}_t$, which transports datapoints $\mathbf{x}_t$ via the ODE $\rd\mathbf{x}_t{=}\boldsymbol{\mu}_t(\mathbf{x}_t)\rd t$. When $p_t$ and $\boldsymbol{\mu}_t$ jointly satisfy the continuity equation
\vspace{-1pt}
\begin{equation}\label{eq:continuity}
    \frac{\rd}{\rd t} p_t + \nabla_{\mathbf{x}} \cdot (p_t \boldsymbol{\mu}_t) = 0,
\end{equation}
then $p_t$ will be a properly normalized density at each $t$. To perform flow matching, one would train a model $\boldsymbol{\nu}_\theta(\mathbf{x}_t, t)$ of the vector field $\boldsymbol{\mu}_t(\mathbf{x}_t)$ by direct regression,
\vspace{-1pt}
\begin{equation}\label{eq:fm_loss}
    L_{\text{FM}}(\theta) = \mathbb{E}_{t\sim \mathcal{U}(0,1), \, \mathbf{x}_t\sim p_t}\Vert \boldsymbol{\nu}_\theta(\mathbf{x}_t, t) - \boldsymbol{\mu}_t(\mathbf{x}_t)\Vert^2_2.
\end{equation}
However, in practice neither $p_t$ nor $\boldsymbol{\mu}_t$ is uniquely determined, we can only sample from $p_t$ for $t=0$ (data) and $1$ (prior), and we do not have access to $\boldsymbol{\mu}_t$ for evaluation at $\mathbf{x}_t$.

As a workaround, conditional flow matching (CFM) proposes to use conditional densities \hbox{$p_t(\mathbf{x}_t \mid (\mathbf{x}_0, \mathbf{x}_1))$} and vector fields $\boldsymbol{\mu}_t(\mathbf{x}_t \mid (\mathbf{x}_0, \mathbf{x}_1))$, where $\mathbf{x}_0\sim p_0$ is a training datapoint and $\mathbf{x}_1\sim p_1$ is noise, such that both are tractable. For example, when we define
\vspace{-1pt}
\begin{equation}\label{eq:cfm_prob}
    p_t(\mathbf{x}_t\mid (\mathbf{x}_0, \mathbf{x}_1)) = \mathcal{N}\big(\mathbf{x}_t ; t\mathbf{x}_1 + (1-t)\mathbf{x}_0, \sigma^2\mathbf{I}_D\big), \quad \ \
    \boldsymbol{\mu}_t(\mathbf{x}_t\mid (\mathbf{x}_0, \mathbf{x}_1)) = \mathbf{x}_1 - \mathbf{x}_0,
\end{equation}
for some $\sigma\geq 0$, the continuity equation (Eq. \ref{eq:continuity}) is satisfied \cite{tong2023improving}. Now sampling data conditionally as $\mathbf{x}_t\sim p_t(\cdot\mid (\mathbf{x}_0, \mathbf{x}_1))$, the CFM loss
\vspace{-1pt}
\begin{equation}\label{eq:cfm_loss}
    L_{\text{CFM}} = \mathbb{E}_{t\sim \mathcal{U}(0,1), \, \mathbf{x}_0\sim p_0,  \, \mathbf{x}_1\sim p_1, \, \mathbf{x}_t\sim p_t(\cdot\mid (\mathbf{x}_0, \mathbf{x}_1))}\Vert \boldsymbol{\nu}_\theta(\mathbf{x}_t, t) - \boldsymbol{\mu}_t(\mathbf{x}_t\mid (\mathbf{x}_0, \mathbf{x}_1))\Vert^2_2,
\end{equation}
has the same gradients as Eq. \ref{eq:fm_loss}, and therefore will lead to the same model $\boldsymbol{\nu}_\theta(\mathbf{x}_t, t)$, but is actually tractable. Finally, new datapoints are generated by solving the ODE starting from $\mathbf{x}_1\sim p_1$ but using the learned vector field $\boldsymbol{\nu}_\theta(\mathbf{x}_t, t)$ instead of $\boldsymbol{\mu}_t(\mathbf{x}_t)$.

\subsection{ForestDiffusion and ForestFlow}\label{sec:forest_recap}

\vspace{-2pt}
There is a clear commonality between the two methods: both regress a parameterized vector field. In almost all applications to date, NNs are used to parameterize the vector field. \citet{jolicoeur2023generating} made the interesting observation that an XGBoost regressor \cite{chen2016xgb} could be used instead, potentially harnessing the successes of tree-based learning in a generative setting. However, there are several major differences between how NNs and XGBoost are trained that must be overcome.

\vspace{-2pt}
First, when using NNs Eq. \ref{eq:sm_loss} or \ref{eq:cfm_loss} would be optimized by sampling a minibatch of data $\mathbf{x}_0\sim p_0$, sampling $t\sim \mathcal{U}(0,1)$ independently for each $\mathbf{x}_0$, sampling fresh noise $\mathbf{x}_1\sim p_1$ (Eq. \ref{eq:cfm_loss} only), and then sampling $\mathbf{x}_t$ from the Gaussian in Eq. \ref{eq:diffusion_conditional} or \ref{eq:cfm_prob}, respectively. In particular, the timestep $t$ and random vector $\mathbf{x}_t$ would be sampled anew every batch, eventually leading to good coverage of the distributions in the loss function expectations. XGBoost is not trained with minibatches; it takes an entire dataset at once and minimizes the loss overall. Hence, the random vector $\mathbf{x}_t$ for each training point $\mathbf{x}_0$ would only be sampled once. For better coverage of the distribution, \cite{jolicoeur2023generating} proposed to duplicate each of the $n$ training datapoints $K$ times, and generate different $\mathbf{x}_t$ for each copy. 

\vspace{-2pt}
Second, whereas a NN can easily be designed with a number of outputs equal to the number of features $p$ in $\mathbf{x}$ (the same size as the target vector field), standard decision trees only output a scalar. A brute-force workaround is to train a different XGBoost ensemble to predict each feature \cite{jolicoeur2023generating}.

\vspace{-2pt}
Third, whereas with a NN the time step $t$ could be fed in as an additional input to the network during training and generation, \cite{jolicoeur2023generating} argued that simply adding $t$ as a feature to XGBoost is unlikely to give sufficient emphasis to it, instead proposing to discretize $t$ into $n_t$ uniform steps and train a different XGBoost ensemble for each. The expectation over $t$ is removed in the loss function Eq. \ref{eq:sm_loss} or \ref{eq:cfm_loss}, and $t$ is instead treated as a constant for each of $n_t$ separate loss functions.

\vspace{-2pt}
Fourth, when conditional generation on a class label $y$ is required, a NN can accept $y$ as an input to adapt its behaviour while sharing parameters. Like conditioning on $t$, conditioning on $y$ is better done by training a separate XGBoost ensemble for each of the $n_y$ classes \cite{jolicoeur2023generating}.

\begin{wraptable}[12]{r}{0.45\textwidth}
    \vspace{-10pt}
    \centering
    \setlength{\tabcolsep}{2pt} 
    \caption{\small \textbf{Top:} The largest training datasets from \cite{jolicoeur2023generating} in terms of $n$, $p$, and $np$. N/A means $y$ is continuous. \textbf{Bottom:} We scale to calorimeter datasets which are up to 370$\times$ larger in $np$.}
    \label{tab:dataset_size}
    \vspace{2pt}
    \small
    \begin{tabular}{lrrr}
        Dataset & Datapoints $n$ & Columns $p$ & Classes $n_y$ \\
        \toprule
        calif. & 16,512 & 9 & N/A \\
        libras & 288 & 90 & 15 \\
        bean & 10,888 & 16 & 7 \\
        \midrule
        Photons & 121,000 & 368 & 15 \\
        Pions & 120,800 & 533 & 15 \\
        \bottomrule
    \end{tabular}
\end{wraptable}

Combining these four solutions, \cite{jolicoeur2023generating} proposed ForestDiffusion and ForestFlow, aiming to realize the promises of tree-based generative modeling laid out in Section \ref{sec:introduction}. While \cite{jolicoeur2023generating} reported excellent model performance, there are clear limitations, mainly the memory requirements from data duplication, and inefficient parameter use from predicting each column with separate XGBoost ensembles. In total, on a tabular dataset of size \hbox{[$n$, $p$]}, both methods require training $n_t\cdot n_y\cdot p$ XGBoost ensembles on $n_t$ different datasets of size [$n_i\cdot K$, $p$], where $n_i$ is the number of datapoints with label $i$ such that $\sum_{i=1}^{n_y} n_i=n$. The recommended settings are $n_t\approx 50$ and $K\approx 100$, whereas $n_i\approx n/n_y$ for class-balanced data. To emphasize the scaling issues, the \emph{libras} dataset featured in \cite{jolicoeur2023generating} with $n=288$ training datapoints required 151 GiB of CPU memory using the original implementation. We address these scaling issues below with novel techniques and better implementation. 

\subsection{Calorimeter Simulation}
\vspace{-2pt}
To motivate the need for scalable tabular data generation, we consider an important scientific application - calorimeter simulation. Measuring particle energy with calorimeters is one of the major components in particle accelerator experiments. To understand predictions from theory, physicists simulate interactions within calorimeters, but doing so from first principles is computationally expensive \cite{geant4, geant4-add1, geant4-add2}. Generative models have seen remarkable uptake as surrogates for fast simulation, spurred on by the Fast Calorimeter Simulation Challenge \cite{calochallenge} which provides large tabular datasets of calorimeter measurements, and evaluation metrics that are scientifically relevant. A comparison of the training dataset sizes from \cite{jolicoeur2023generating} and from the Challenge is given in Table \ref{tab:dataset_size}. In Appendix \ref{app:calorimeter} we provide an extensive review of machine learning methods for calorimeter simulation.

\section{Scaling Up}
\label{sec:scaling}

In this section we provide a step-by-step breakdown of the implementation of the main algorithm from \cite{jolicoeur2023generating}, shown in pseudocode in Algorithm \ref{alg:train}, then re-engineer it to scale to calorimeter data. After resolving implementation issues, we offer new techniques to further improve algorithm and model performance, including by completely changing the tree structure used by XGBoost.

\subsection{Limitations of the Existing Implementation}\label{sec:limitations}

\vspace{-5pt}
\begin{tabular}{@{}cc@{}}
\begin{minipage}{.57\textwidth}
\noindent
\begin{algorithm}[H]
    \caption{\small ForestDiffusion and ForestFlow}
    \label{alg:train}
    \small
    \SetNoFillComment
     \textbf{Input:} Dataset $X_0$ of size $[n,p]$, $K$, $n_{t}$.\\

    $X'_{0} \gets$ $K$-fold duplicate of the rows of $X_0$\\
    $X_1 \gets $ Dataset of  $\mathbf{x}_1 \sim \mathcal{N}(0,\mathbf{I}_{p})$ with the size of $X'_{0}$\\
     \For {Timestep $t \in \text{range}(n_t)$}
     {
         \For {Class $y \in \text{range}(n_y)$}
        {
            \For {Feature $p_i \in \text{range}(p)$}
            {
            $X'_{0,y} \gets$ rows of $X'_{0}$ with label $y$\\
            $X_{1,y} \gets$ corresponding rows of $X_{1}$\\
            \tcp{Create $\mathbf{x}_t$ and regression targets}
            $X'_{t,y}, Z_{t,p_i} \gets \text{Forward}(X'_{0,y}, X_{1,y}, t, p_i)$
    
            $f_{t, y, p_i}\!\! \gets$\! Regress on $Z_{t,p_i}$ given $X'_{t,y}$
            }
        }
     }
     
     \Return $\{f_{t, y, p_i} \}$\,\tcp{\!Set\,of\,$n_t n_y p$\:XGBoost\,ensembles}
\end{algorithm}
\end{minipage} 
&
\begin{minipage}{.41\textwidth}
\centering
    \includegraphics[width=0.9\textwidth, trim={5 5 5 5}, clip]{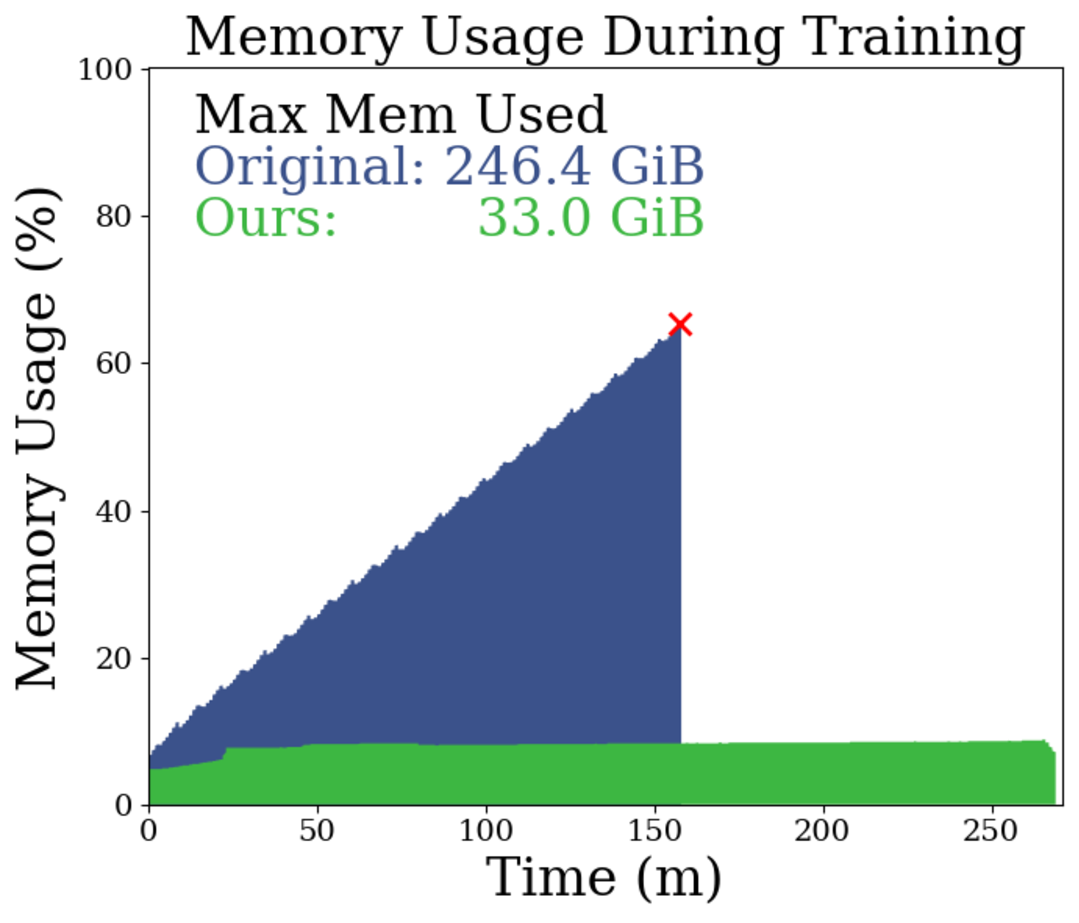}
    \vspace{-5pt}
    \captionof{figure}{\looseness=-2Memory usage during training for the original implementation and ours. The red cross \textcolor{red}{$\times$} indicates job failure.}
    \label{fig:usage_1}
\end{minipage}
\end{tabular}

\normalsize 

\looseness=-2 In Algorithm \ref{alg:train} the triple loop must be parallelized well for the method to scale to large datasets (see Appendix \ref{app:implementation} for discussion). Since the training dataset $X_0$ is duplicated $K$ times and $n_t$ different versions are generated to represent $\mathbf{x}_t$ at different timesteps, memory can become a severe issue which is compounded by the number of XGBoost ensembles to be trained. Unfortunately, these challenges are not handled well in the implementation published by \cite{jolicoeur2023generating}. To demonstrate the need for rework, we show the resource usage during training for a small dataset with $n=1000$, $p=100$, and $n_y=10$ in Figure \ref{fig:usage_1}. Despite our computer having 385 GiB of CPU memory available, the training job failed. There are three clear problems that this example shows: (1) even on a modest dataset the absolute amount of memory consumed is unexpectedly high (250 GiB); (2) memory usage grows at a constant rate during training, potentially leading to out-of-memory errors hours into a job; (3) training can fail due to memory issues even when the system maximum has not been reached.

\subsection{Outlining the Existing Implementation}\label{sec:outline}

To begin unravelling the causes of the three undesirable behaviours above, we consider in detail the Python implementation by \cite{jolicoeur2023generating}.\footnote{We base our discussion on \href{https://github.com/SamsungSAILMontreal/ForestDiffusion}{\tt{github.com/SamsungSAILMontreal/ForestDiffusion}}, commit hash \texttt{855281b} dated Nov. 2, 2023. After this point significant changes involving the XGBoost data iterator were added. We discuss the data iterator and reasons for not using a more recent version in Appendix \ref{app:data_iterator}.} First, the dataset $X_0$ of size $[n, p]$ is given as a Numpy array, and if the dataset has discrete labels for conditioning, these are denoted as $y$. A min-max scaler applied to $X_0$ fits all data into the range $[-1, 1]$, after which the dataset is duplicated $K$ times, giving $X'_{0}$ of size $[nK, p]$, and noise $X_1$ is sampled with the same shape. For conditioning on the $n_y$ distinct classes in $y$, $n_y$ Boolean masks are created over $X'_0$. With this preparatory work done, the regression inputs and targets are created, denoted by $X'_{t}$ and $Z_{t}$ respectively. $X'_{t}$ represents samples $\mathbf{x}_t$ from the distribution in Equation \ref{eq:diffusion_conditional} or \ref{eq:cfm_prob}, while $Z_{t}$ is either the score function $\nabla_{\mathbf{x}_t} \log p_t(\mathbf{x}_t\mid\mathbf{x}_0)$ from Equation \ref{eq:sm_loss} or the conditional vector field $\boldsymbol{\mu}_t(\mathbf{x}_t\mid (\mathbf{x}_0, \mathbf{x}_1))$ from Equation \ref{eq:cfm_loss}. Finally, all $n_t\cdot n_y\cdot p$ models are trained in a parallel triple loop with the widely used Python parallelization library Joblib. These steps are shown in the following code snippet, which has been compressed to show only the crucial information, and uses ForestFlow as a representative example.

\vspace{-5pt}
\begin{mintedbox}{python}{Python Implementation of ForestFlow from \cite{jolicoeur2023generating}}
from sklearn.preprocessing import MinMaxScaler
import numpy, xgboost
from joblib import delayed, Parallel
X0, y, K, n_t, xgb_kwargs, n_jobs = inputs()
n, p = X0.shape
# Scale data so that range matches noise variance
X0 = MinMaxScaler(feature_range=(-1, 1)).fit_transform(X0)
# Duplicate data and generate noise
X0 = numpy.tile(X0, (K, 1)) # [n*K, p]
X1 = numpy.random.normal(size=X0.shape) # [n*K, p]
# Create Boolean masks for class conditioning
mask = {} # Boolean mask for which rows of X0 have label y_i
y_uniq = numpy.unique(y)
for y_i in y_uniq:
  mask[y_i] = numpy.tile(y == y_i, K)
# Create regression targets (ForestFlow)
t = numpy.linspace(0, 1, num=n_t) # Discretize time into n_t equal steps
X_train = t*X1 + (1-t)*X0 # [n_t, n*K, p]
Z_train = X1 - X0 # [n*K, p]
# Train models in parallel triple loop over timesteps, classes, and features
def train_parallel(X_tr_i, Z_tr_i):
  model = xgboost.XGBRegressor(**xgb_kwargs)
  return model.fit(X_tr_i, Z_tr_i)
regressors = Parallel(n_jobs)(
  delayed(train_parallel)(
    X_train[t_i][mask[y_i], :], Z_train[mask[y_i], p_i]
  ) for t_i in range(n_t) for y_i in y_uniq for p_i in range(p)
) # regressors is list of n_t*n_y*p XGBoost ensembles
\end{mintedbox}
\subsection{Analysis and Improvement of the Implementation}\label{sec:analysis}
\vspace{-2pt}
While the implementation above looks innocuous, Figure \ref{fig:usage_1} shows that there are serious engineering issues lurking. We aim to answer the following specific questions based on the observations above: (1) Why are memory requirements high for tiny datasets? (2) Why does memory usage increase during training? (3) Why do jobs fail before memory usage reaches 100\%? We proceed by identifying issues, proposing solutions, and quantifying the benefits, starting with a simple observation.\\
\looseness=-2 \textbf{Issue 1:} Regression inputs \texttt{X\_train} for all timesteps are created in memory at once, which is a large array of shape \texttt{[n\_t, n*K, p]} (Line 18). Using the recommended values of \texttt{n\_t} and \texttt{K} creates an array 5000 times the size of the training dataset, making even small datasets burdensome (Question 1).\\
\textbf{Solution 1:} Each XGBoost training call requires only the information at a single timestep \texttt{X\_train[t\_i]}, and so this data should be generated on-the-fly within the \texttt{n\_t} loop.\\
\textbf{Benefit 1:} We avoid holding an array of size \texttt{[n\_t, n*K, p]} in memory. This is in fact the first issue encountered when trying to apply the implementation from \cite{jolicoeur2023generating} to calorimeter data, since a \texttt{numpy.float64} array of size \texttt{[50, 120800*100, 533]} (Table \ref{tab:dataset_size}) requires \textbf{2.34 TiB} of memory. \\
\begin{minipage}[t]{0.49\linewidth}%
\vspace{0pt}
\begin{mintedbox}{python}{Issue 1: Original}
t = numpy.linspace(0, 1, num=n_t)
X_train = t*X1 + (1-t)*X0
# [n_t, n*K, p] array in memory
\end{mintedbox}
\end{minipage}%
\hfill
\begin{minipage}[t]{0.51\linewidth}%
\vspace{0pt}
\begin{mintedbox}{python}{Issue 1: Improvement}
def train_parallel(X0,X1,Z_tr_i,t_i):
  X_tr_i = t_i*X1 + (1-t_i)*X0
  model = xgboost.XGBRegressor()
  return model.fit(X_tr_i, Z_tr_i)
\end{mintedbox}
\end{minipage}%

\textbf{Issue 2:} Parallelization of the many XGB training calls is necessary for an efficient implementation. We discuss this requirement and other approaches to parallelization in Appendix \ref{app:implementation}. Due to the Python global interpreter lock, parallelization is often handled through multiprocessing with Joblib, wherein the main process spawns worker processes and sends them copies of data. When a large array is assigned to multiple workers, Joblib automatically puts it into shared memory as a memory-mapped file, and passes only the reference to the workers. When an indexed array like \texttt{X\_train[t\_i][mask[y\_i], :]} is passed to Joblib’s \texttt{Parallel} call (Line 26), it is treated as a distinct array, even though the same indexed array might appear for many jobs. This is because ``advanced indexing'' in Numpy, such as when the array \texttt{mask[y\_i]} is used as the selection object, always creates a copy of the data. Hence, Joblib creates a new array in shared memory for every copy throughout every call in the triple parallel loop. By default, Joblib stores the memory-mapped arrays in RAM disk, a virtual disk on RAM, and does not free that memory until all parallel jobs are done. This continuously increases RAM disk usage during training (Question 2) and can lead to out-of-memory errors if RAM disk usage reaches its maximum capacity. For example, the maximum shared memory size on our Ubuntu 22.04 machine had been set to 189 GiB by default, and the failure in Figure \ref{fig:usage_1} was caused by the RAM disk reaching this limit, even though RAM itself (at 385 GiB) was not at 100\% usage (Question 3).\\
\textbf{Solution 2:} \looseness=-2 Instead of passing indexed arrays in the \texttt{Parallel} call, pass the entire array and index it inside worker processes. Upon the first call, Joblib puts the array into shared memory, but in subsequent calls, Joblib identifies the same array being requested and passes only a reference to workers.\\
\textbf{Benefit 2:} In the original implementation, each \texttt{train\_parallel} call in the triple loop over (\texttt{n\_t}, \texttt{n\_y}, \texttt{p}) creates a copy of the inputs in shared memory. Looking only at \texttt{X\_train}, the triple loop consumes \texttt{n\_t*p*sizeof(X\_train[t\_i])=n\_t*p*(n*K*p*8)} bytes in shared memory, which is \texttt{p} times more than Issue 1 (Question 1). For the Pions dataset (Table \ref{tab:dataset_size}), this would amount to \textbf{1.22 PiB}. Our solution holds only one copy of \texttt{X0} and \texttt{X1} in shared memory, a factor of \texttt{n\_t*p} less.
\vspace{-2pt}
\begin{mintedbox}{python}{Issue 2: Improvement}
def train_parallel(X0, X1, Z_train, t_i, mask_i, p_i):
  X_tr_i = t_i*X1[mask_i, :] + (1-t_i)*X0[mask_i, :]
  Z_tr_i = Z_train[mask_i, p_i]
  model = xgboost.XGBRegressor(**xgb_kwargs)
  return model.fit(X_tr_i, Z_tr_i)
regressors = Parallel(n_jobs)(
  delayed(train_parallel)(
    X0, X1, Z_train, t_i, mask[y_i], p_i
  ) for t_i in t for y_i in y_uniq for p_i in range(p) )
\end{mintedbox}
\vspace{-2pt}

\textbf{Issue 3:} All trained XGBoost models are held in memory until the end of training, causing a steady increase of consumed memory as training progresses (Question 2). The memory consumed by these models is independent of \texttt{n}, but increases with \texttt{p} and \texttt{n\_y} (Question 1).\\
\textbf{Solution 3:} After a model is trained, write it to disk, and delete it from memory. Use the Universal Binary JSON (UBJ) format as it is compatible across XGBoost versions, and is the fastest format overall for reading and writing with the best compression on disk.\\
\textbf{Benefit 3:} Workers writing their trained model to disk prevents the growth of memory usage over training, and bypasses the need to return models from the worker to the main process via pickling. Furthermore, it serves as a checkpoint so that training can be easily resumed upon system failure. ForestFlow requires \texttt{n\_t*n\_y*p} XGBoost ensembles, each made of \texttt{n\_tree} trees which themselves have up to \texttt{2**(d+1)-1} nodes where \texttt{d} is the maximum depth. Each node of an XGBoost tree uses 53 bytes to store parameters, metadata, and training statistics. Using the recommended defaults of \texttt{n\_tree=100}, \texttt{d=7}, and no regularization, essentially all trees will have the maximum 255 nodes, and would require in total \textbf{503 GiB} on the Pions dataset (Table \ref{tab:dataset_size}).\\
\begin{minipage}[t]{0.5\linewidth}%
\vspace{0pt}
\begin{mintedbox}{python}{Issue 3: Original}
def train_parallel(...):
  ...
  return model.fit(X_tr_i, Z_tr_i)
\end{mintedbox}
\end{minipage}%
\hfill
\begin{minipage}[t]{0.5\linewidth}
\begin{mintedbox}{python}{Issue 3: Improvement}
def train_parallel(...):
  ...
  model.fit(X_tr_i, Z_tr_i)
  model.save_model(f"{path}.ubj")
\end{mintedbox}
\end{minipage}

At this point we have reviewed the most significant training issues which together explain the three problematic observations from Figure \ref{fig:usage_1}, and when fixed, provide the bulk of resource improvements that we report. In Appendix \ref{app:continued_issues} we list four additional issues and solutions of smaller magnitude, and present a Python implementation with all of our changes. Figure \ref{fig:usage_1} shows that our implementation solves the three observed problems. However, training is only one part of the story. In Appendix \ref{app:generation} we investigate improvements to generation speed, with our method proving to be more than an order of magnitude faster. 

\subsection{Algorithmic Modifications}\label{sec:alg_improvements}
\vspace{-2pt}
Beyond improving the implementation of Algorithm \ref{alg:train}, we also offer modifications that can improve model performance and resource efficiency. Our two most significant changes are described below, with further minor recommendations in Appendix \ref{app:performance}, along with ablation studies to understand the impact of the proposed changes and key hyperparameters.

\textbf{Multi-output trees:}  First, we propose a significant change to the structure of trees to make them more suited to the high-dimensional outputs required in generative modelling. Instead of training $p$ single-output (SO) trees, one for each feature, we propose to use multi-output (MO) trees \cite{zhang2021gbdtmo, ying2022mt, marz2022multi, iosipoi2022sketchboost, schmid2023tree}, where a single tree outputs $p$ values. This is clearly advantageous for parameter efficiency: $p$ times fewer XGBoost ensembles are required to represent the vector field, which is a massive benefit for generation speed and trained model memory requirements. However, multi-output trees also have the potential to better model the joint distribution of data. Consider how a set of $p$ single-output trees generates a vector output. From identical inputs, each tree independently identifies the appropriate leaf node and outputs a scalar – there is no dependence between elements during generation. This is clearly not desirable for generative models where we aim to model the joint distribution of the data, not merely the marginals. Multi-output trees can better represent these correlations since generated elements come from a single tree with interactions during generation.

\begin{figure}[t]
    \centering
    \includegraphics[width=0.97\linewidth]{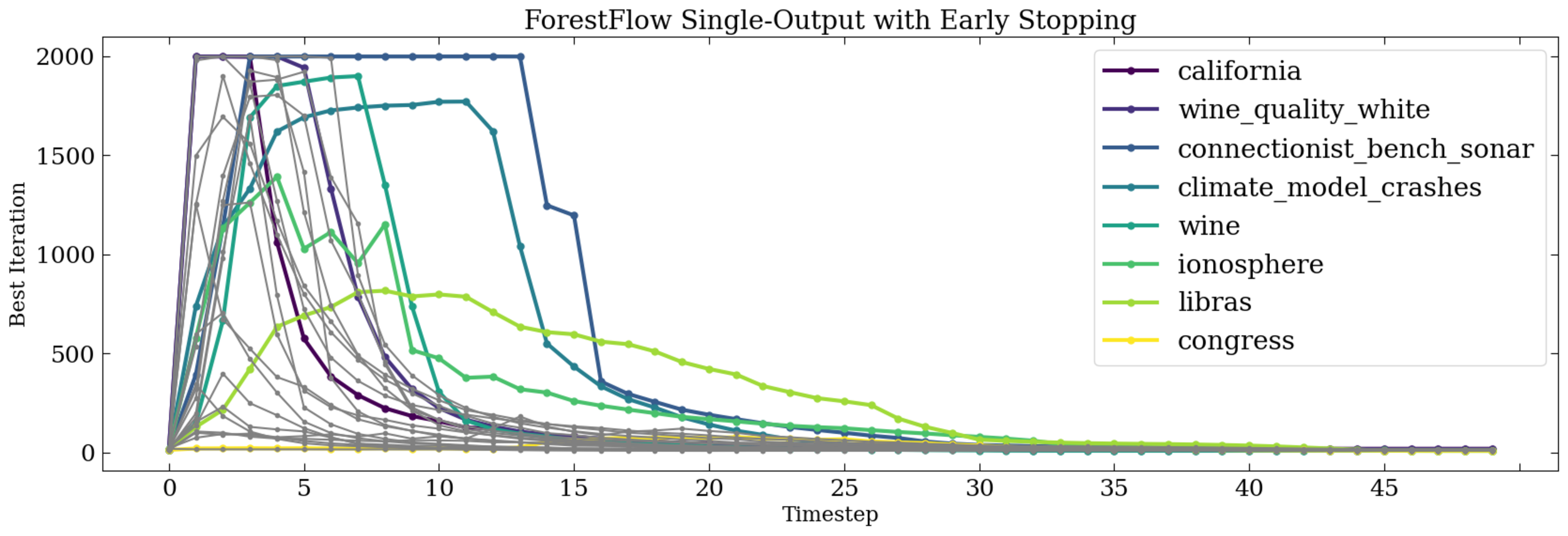}
    \caption{Number of trees at the best iteration of the validation loss by timestep and dataset. Selected datasets from all 27 are highlighted for comparison to MO in App. \ref{app:hypers}. Early stopping after $n_{\text{ES}}=20$ rounds with no improvement prevents wasteful training where no progress is being made.}
    \label{fig:n_tree_SO_flow}
    \vspace{-8pt}
\end{figure}

\textbf{Early stopping:}  Second, our improved implementation unlocks the ability to greatly scale up model size by increasing $K$ for better coverage of the expectations in Eq. \ref{eq:sm_loss} or \ref{eq:cfm_loss}, and by increasing the number of trees per ensemble, $n_\text{tree}$, for more model capacity. \citet{jolicoeur2023generating} noted that underfitting was a major concern, and hence used no regularization. Our findings are more nuanced; underfitting is a concern, but moreso for values of $t$ near $0$ (data) than $1$ (noise). Models trained for large values of $t$ require much less capacity. This is related to the well-known phenomenon that noise scheduling has a strong impact on diffusion model performance \cite{pmlr-v139-nichol21a, chen2023importance}. Rather than tune $n_{\text{tree}}$ for different $t$, we propose to adaptively regularize each model using early stopping. This could be done with a held-out validation set, but since many benchmark tabular datasets are small to begin with, reducing the training set further is disadvantageous. Instead, because the regression targets for both diffusion and CFM are generated by adding noise to the training data, we validate using the training set $X_0$ but with fresh noise $X_1$. An example of how the best iteration varies with timestep is shown in Figure \ref{fig:n_tree_SO_flow}, stopping after $n_{\text{ES}}=20$ trees without improvement. Coupling a very large $n_{\text{tree}}=2000$ with early stopping proves to be an effective regularizer as it allows wide ensembles for the more challenging training iterations while preventing overfitting on the easier steps. We show below in Section \ref{sec:resource-scaling} that adding early stopping to the default settings greatly speeds up training, as it avoids wasted computation for ensembles which have converged.
\vspace{-4pt}

\section{Experimental Results}
\label{sec:results}
\vspace{-4pt}
In this section we demonstrate the improved resource scaling of our implementation, provide benchmarking on small tabular datasets for model performance improvements, then present our results on the much larger calorimeter datasets. Complete details on datasets and hyperparameters are given in Appendix \ref{app:experiments} for reproducibility.

\subsection{Resource Usage Scaling}\label{sec:resource-scaling}

\looseness=-2 We quantify resource usage using synthetic datasets of controllable size. Features and labels are randomly generated, with one of $n$, $p$, or $n_y$ altered from base values of $n\!=\!1000$, $p\!=\!10$, and $n_y\!=\!10$. Increasing $n$ should increase training time and memory usage, but at most linearly due to XGBoost's \texttt{hist} method \cite{chen2016xgb}. Increasing $n_y$ with $n$ held fixed means increasing the number of ensembles trained, but with each on a smaller dataset, hence memory usage should slightly decrease. Meanwhile, increasing $p$ leads to quadratic scaling in time since it increases the number of ensembles required, and the size of each dataset, the latter of which implies linear memory requirements.

\begin{figure}
    \centering
    \includegraphics[scale=0.39, trim = {0 0 0 0}, clip]{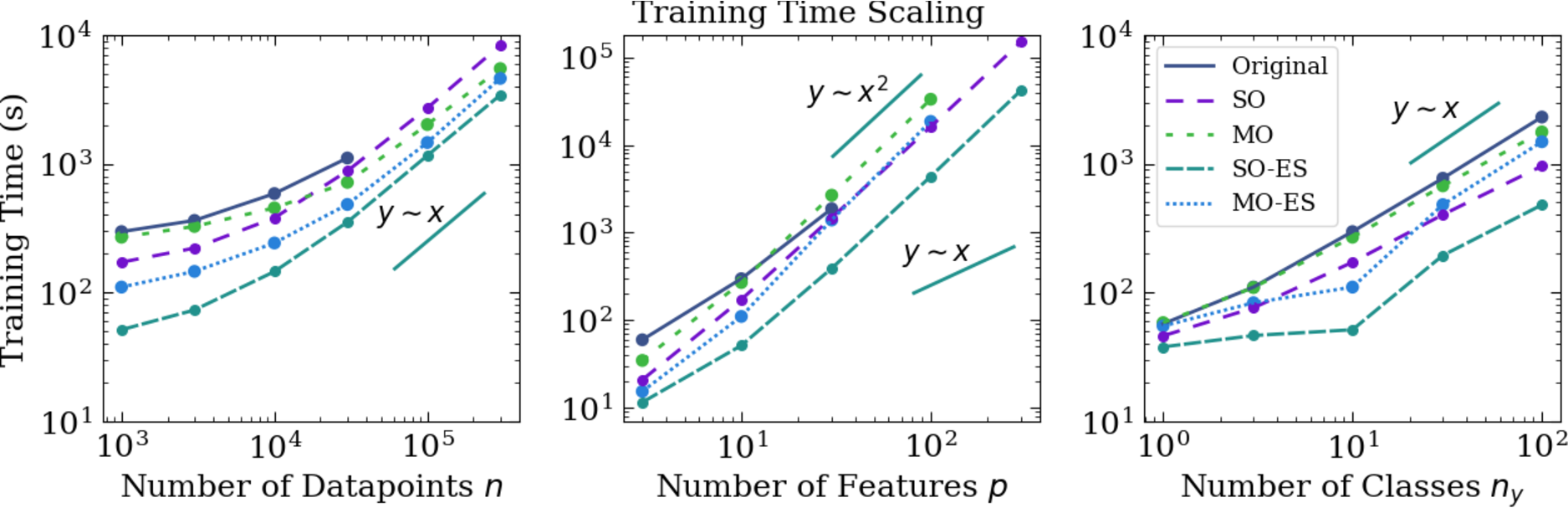}
    \includegraphics[scale=0.39, trim = {0 0 0 0}, clip]{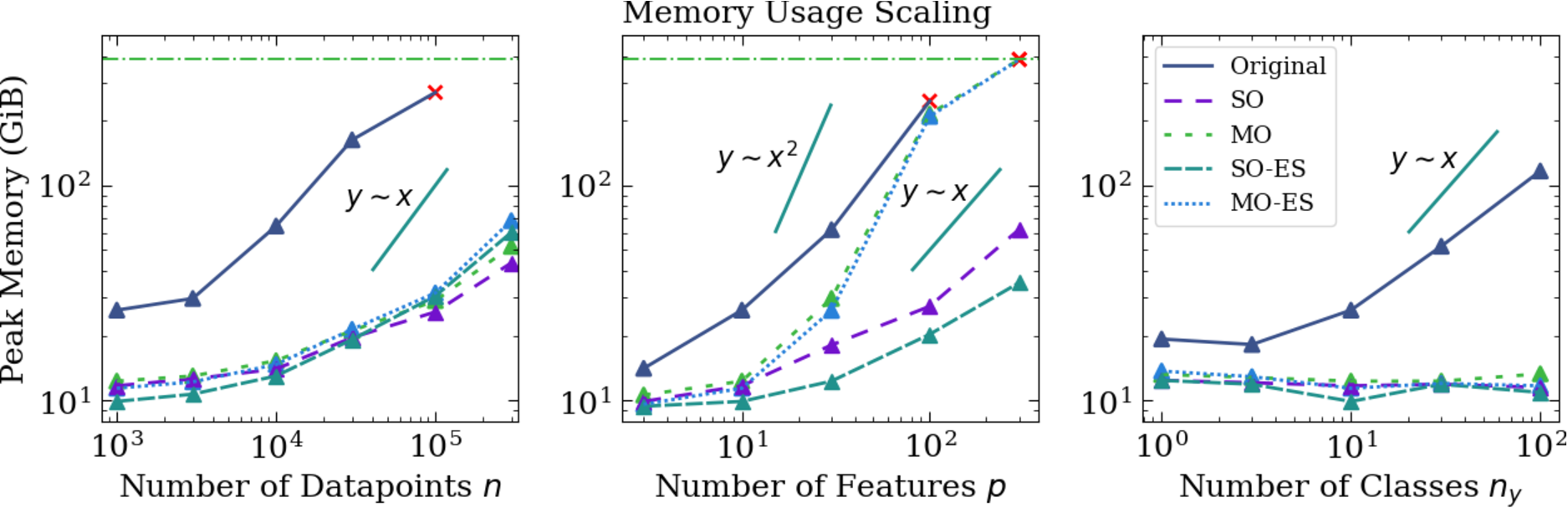}
        \includegraphics[scale=0.39, trim = {0 0 0 0}, clip]{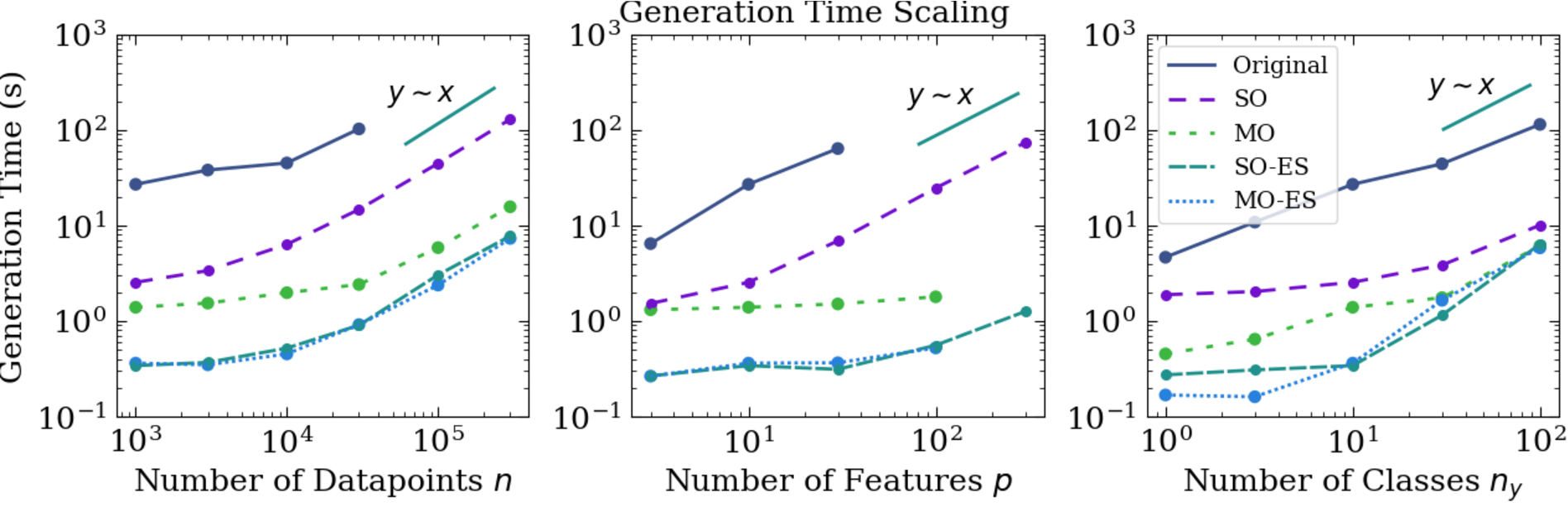}
    \caption{Resource usage of the ForestFlow implementation from \cite{jolicoeur2023generating}, compared to our implementation (SO), including with multi-output trees (MO), and early stopping (ES). \textbf{Top:} Training time. \textbf{Middle:} Peak memory usage. \textbf{Bottom:} Generation time. A red cross \textcolor{red}{$\times$} for memory indicates job failure, and hence corresponding points in other plots are unavailable. A horizontal line indicates the maximum system memory used for all models at 385 GiB.}
    \label{fig:resources-random}
    \vspace{-10pt}
\end{figure}

Figure \ref{fig:resources-random} shows our controlled experiments for training time, memory, and generation time scaling. We compare the Original implementation of ForestFlow to ours (SO) which produces the same models. Additionally, we show the multi-output (MO) variant with the same learning hyperparameters, and our proposal for adaptive early stopping (SO-ES, MO-ES) which trains much smaller models. Each plot is on a log-log scale, and we display line segments with linear or quadratic behaviour so that scaling exponents can easily be read off.\footnote{In some cases the constant overhead can be significant, and the independent variable cannot be made large enough to reach the asymptotic regime. For example, in the second row, first column, memory scaling may appear sublinear for methods other than Original due to the constant, but is actually linear.}

All methods show linear training time in $n$ as nearly all compute time is spent in calls to XGBoost. However, SO reduces overhead which can be up to half the training time for small $n$. Meanwhile, early stopping provides up to a 3$\times$ speedup for SO-ES over SO, and a smaller but still useful speedup for MO-ES over MO (see Figure \ref{fig:n_tree} in Appendix \ref{app:hypers} for an illustration of early stopping's effect).

Our implementation greatly reduces memory overhead, and peak memory usage scales linearly with $n$. The Original implementation shows worse-than-linear scaling (after accounting for the overhead), and leads to job failures with as few as $n=10,000$ datapoints. The number of features $p$ has the biggest impact on resource use, but our implementation reduces the memory scaling from quadratic to linear. Finally, for $n_y$ our implementation uses a constant amount of memory, whereas the Original scales linearly. While MO behaves similarly to SO in $n$ and $n_y$, MO suffers from worse scaling in $p$. Multi-output trees reduce the number of ensembles required by a factor of $p$, but each tree's training is more memory intensive as XGBoost must search over a higher dimensional leaf space.

The bottom of Figure \ref{fig:resources-random} shows the time required to generate 5 batches of $n$ datapoints using the corresponding models trained for the other plots. Not only is our implementation more than an order of magnitude faster for most settings, we see an even greater improvement for multi-output trees. The near-constant scaling in $p$ showcases the benefit of generating all $p$ outputs with a single ensemble making multi-output trees a strong candidate for applications that require large volumes of generated data. Since the ES variants drastically reduce the trained model size, generation time is commensurately reduced.

In summary, our implementation and algorithmic improvements together improve the efficiency of ForestDiffusion and ForestFlow for all major resources -- training time, memory, and generation time -- which allows us to scale up to much larger models and datasets in the following sections.
\vspace{-6pt}

\begin{table}[t]
\caption{Average rank (standard error) of generated data quality over 27 datasets. Lower is better.}  
\label{tab:main}
\small
\centering
\resizebox{\textwidth}{!}{
    \setlength{\tabcolsep}{3pt} 
\begin{tabular}{lrrrrrrrr|r}
  \toprule
 & $W1_\textrm{train}$ & $W1_\textrm{test}$ & $\textrm{Cov}_\textrm{train}$ & $\textrm{Cov}_\textrm{test}$ & $R^2_\textrm{gen}$ & $F1_\textrm{gen}$ & $P_\textrm{bias}$ & $\textrm{cov}_\textrm{rate}$ & Avg.\\ \hline
  GaussianCopula \cite{joe2014dependence} & 10.1{\tiny$\pm$0.3} & 10.2{\tiny$\pm$0.3} & 10.2{\tiny$\pm$0.3} & 10.3{\tiny$\pm$0.3} & 9.1{\tiny$\pm$0.1} & 9.6{\tiny$\pm$0.3}& 8.6{\tiny$\pm$1.4}& 10.6{\tiny$\pm$0.5} & 9.8{\tiny$\pm$0.1} \\  
  TVAE \cite{xu2019tvae} & 8.1{\tiny$\pm$0.3} & 8.0{\tiny$\pm$0.3} & 8.5{\tiny$\pm$0.3} &  8.5{\tiny$\pm$0.3} & 9.7{\tiny$\pm$0.6} & 9.0{\tiny$\pm$0.5}& 10.7{\tiny$\pm$0.5}& 10.1{\tiny$\pm$0.3} & 9.1{\tiny$\pm$0.0} \\  
  CTGAN \cite{xu2019tvae} & 11.4{\tiny$\pm$0.1} & 11.4{\tiny$\pm$0.2} & 11.3{\tiny$\pm$0.2} & 11.1{\tiny$\pm$0.2} & 11.6{\tiny$\pm$0.2} & 11.4{\tiny$\pm$0.2}& 8.4{\tiny$\pm$1.2}& 10.6{\tiny$\pm$0.5} & 10.9{\tiny$\pm$0.1} \\  
  CTAB-GAN+ \cite{zhao2024ctabganp} & 9.8{\tiny$\pm$0.3} & 9.7{\tiny$\pm$0.3} & 10.1{\tiny$\pm$0.3} & 9.9{\tiny$\pm$0.3} & 10.0{\tiny$\pm$0.2} & 10.1{\tiny$\pm$0.3}& 10.7{\tiny$\pm$0.6}& 9.3{\tiny$\pm$0.9} & 10.0{\tiny$\pm$0.1} \\  
  STaSy \cite{kim2023stasy} & 9.1{\tiny$\pm$0.2} & 9.3{\tiny$\pm$0.2} & 8.1{\tiny$\pm$0.3} & 8.0{\tiny$\pm$0.4} & 9.0{\tiny$\pm$1.0} & 8.0{\tiny$\pm$0.3}& 7.4{\tiny$\pm$1.1}& 7.4{\tiny$\pm$1.1} & 8.3{\tiny$\pm$0.2} \\  
  TabDDPM \cite{kotelnikov2023tabddpm} & 4.9{\tiny$\pm$0.9} & 6.4{\tiny$\pm$0.7} & 4.9{\tiny$\pm$0.7} & 5.4{\tiny$\pm$0.6} & \better{2.0{\tiny$\pm$0.7}}  & 6.2{\tiny$\pm$0.8}& \better{3.3{\tiny$\pm$1.4}}& 3.4{\tiny$\pm$0.6} & 4.6{\tiny$\pm$0.1} \\  
  \hline
  FD-Original \cite{jolicoeur2023generating} & 6.7{\tiny$\pm$0.1} & 6.3{\tiny$\pm$0.2} & 6.7{\tiny$\pm$0.2} & 5.9{\tiny$\pm$0.5} & 5.7{\tiny$\pm$0.7} & 4.1{\tiny$\pm$0.5}& 6.7{\tiny$\pm$1.1}& 5.7{\tiny$\pm$1.2} & 6.0{\tiny$\pm$0.1} \\  
  FD-SO-Scaled & 3.8{\tiny$\pm$0.3} & 4.1{\tiny$\pm$0.3} & 4.3{\tiny$\pm$0.3} & 4.4{\tiny$\pm$0.3} & 5.6{\tiny$\pm$0.4} & 3.4{\tiny$\pm$0.5}& 4.6{\tiny$\pm$0.7}& 4.6{\tiny$\pm$1.1} & 4.4{\tiny$\pm$0.1} \\  
  FD-MO-Scaled & 4.5{\tiny$\pm$0.2} & 4.2{\tiny$\pm$0.3} & 5.0{\tiny$\pm$0.3} & 4.8{\tiny$\pm$0.4} & 5.7{\tiny$\pm$0.4} & 4.6{\tiny$\pm$0.4}& 5.1{\tiny$\pm$0.8}& 5.9{\tiny$\pm$0.7} & 5.0{\tiny$\pm$0.1} \\ 
  FF-Original \cite{jolicoeur2023generating} & 5.0{\tiny$\pm$0.2} & 4.3{\tiny$\pm$0.3} & 4.0{\tiny$\pm$0.3} & 3.9{\tiny$\pm$0.4} & 4.3{\tiny$\pm$0.8} & 4.8{\tiny$\pm$0.5}& 3.6{\tiny$\pm$0.8}& 3.8{\tiny$\pm$0.7} & 4.2{\tiny$\pm$0.1} \\  
  FF-SO-Scaled & \better{1.7{\tiny$\pm$0.2}} & 2.3{\tiny$\pm$0.2} & \better{2.1{\tiny$\pm$0.3}} & \better{2.9{\tiny$\pm$0.3}} & 2.1{\tiny$\pm$0.1} & \better{3.1{\tiny$\pm$0.5}}& 4.4{\tiny$\pm$0.5}& \better{3.0{\tiny$\pm$0.5}} & \better{2.7{\tiny$\pm$0.0}} \\  
  FF-MO-Scaled & 2.8{\tiny$\pm$0.2} & \better{1.9{\tiny$\pm$0.2}} & 2.7{\tiny$\pm$0.3} & 3.0{\tiny$\pm$0.3} & 3.1{\tiny$\pm$0.7} & 3.8{\tiny$\pm$0.4}& 4.4{\tiny$\pm$0.7}& 3.6{\tiny$\pm$0.8} & 3.2{\tiny$\pm$0.1} \\  
   \hline
\bottomrule 
\end{tabular}
}
\vspace{-14pt}
\end{table}

\subsection{Model Performance on Benchmark Datasets}\label{sec:benchmarking}
\vspace{-2pt}
We scaled up models by increasing the duplication factor $K$ from 100 to 1000, and the maximum number of trees $n_\text{tree}$ from 100 to 2000. In this section, we directly compare scaled-up models using our proposed algorithmic improvements from Section \ref{sec:alg_improvements} to the original ForestDiffusion (FD) and ForestFlow (FF) on 27 datasets \cite{muzellec2020missing}, across 8 metrics, which we averaged over 5 generated datasets for each of 3 seeds. The metrics convey the quality of generated samples along four dimensions: distributional distance (Wasserstein-1 distance to the training or test set), diversity (Coverage \cite{naeem2020prdc} of the training or test set), usefulness for training discriminative models ($R^2_{\textrm{gen}}$ and $F1_{\textrm{gen}})$, and usefulness for statistical inference ($P_\textrm{bias}$ and $\textrm{cov}_\textrm{rate}$). For comparison, 6 baseline generative models are shown ranging from statistical methods to tabular diffusion models. The 27 datasets, 8 evaluation metrics, and 6 baseline methods are repeated from \cite{jolicoeur2023generating}, and described fully in Appendix \ref{app:experiments} along with exact hyperparameter settings in Table \ref{tab:hypers}, and experimental details.

Table \ref{tab:main} shows the average rank that each method obtained on each metric where the average and standard error are computed over the 27 datasets, similar to previous tabular generation papers \cite{gorishniy2021revisiting, gorishniy2023tabr}. We also include ``Avg.'', the row-wise average and standard error to summarize the table, and highlight the single best method on each metric. While the original ForestFlow already outperforms advanced methods like TabDDPM \cite{kotelnikov2023tabddpm} on several metrics, our performance improvements and scaling-up of the single-output case further establish it as a state-of-the-art tabular generative model. Multi-output trees are also competitive when scaled up, and notably achieve the lowest Wasserstein distance to the test set, showing the best ability to capture the underlying data distribution. Raw metric values are plotted in Appendix \ref{app:extra_benchmarking_plots}.

\subsection{CaloForest - Flow-based XGBoost Models for Calorimeter Data}\label{sec:calo}
\vspace{-2pt}
To show that our implementation scales to much larger datasets, we model Photons and Pions (Table~\ref{tab:dataset_size}) from the Fast Calorimeter Simulation Challenge \cite{calochallenge}. Each dataset comes with training and test splits of size $\approx\!121,000$, where each datapoint represents the energies deposited in voxels of a calorimeter by an incident particle. 

Competitive NN-based approaches perform extensive pre-processing to the data to facilitate training \cite{krause2021caloflow2, Mikuni:2022xry, cresswell2022caloman}. Since XGBoost is robust to features on different scales and with different distributions we need only perform min-max scaling on each class. We used our SO variant of ForestFlow as the MO variant still has high memory costs for large $p$ (Figure \ref{fig:resources-random}). We discretized time into $n_t=100$ steps, and duplicated each datapoint $K=20$ times. Each XGBoost ensemble had $n_{\text{tree}}=20$ trees of maximum depth 7, a learning rate of 1.5, and all other XGBoost hyperparameters left as defaults. We trained up to 20 XGBoost ensembles in parallel, each with 2 CPUs, on a single desktop workstation with 250 GiB RAM and 40 CPUs (Intel Xeon Silver 4114T). In total, for the Photons model 552,000 XGBoost ensembles were trained in 135 hours with a peak memory burden of 54 GiB, while the Pions model used 799,500 ensembles, completed in 281 hours, and required 78 GiB of memory. Generation of $n$ datapoints (matching the number in the training and test sets) took 231 s for Photons (1.91 ms per datapoint), and 347 s for Pions (2.87 ms per datapoint), which can be compared to 40 ms per datapoint for score-based NNs on a GPU \cite{Mikuni:2022xry}, or anywhere from 100 ms to 3 s for the widely used Geant4 simulator \cite{aad2021atlfast3}.

\begin{figure}[t]
    \centering
        \includegraphics[width=0.158\textwidth]{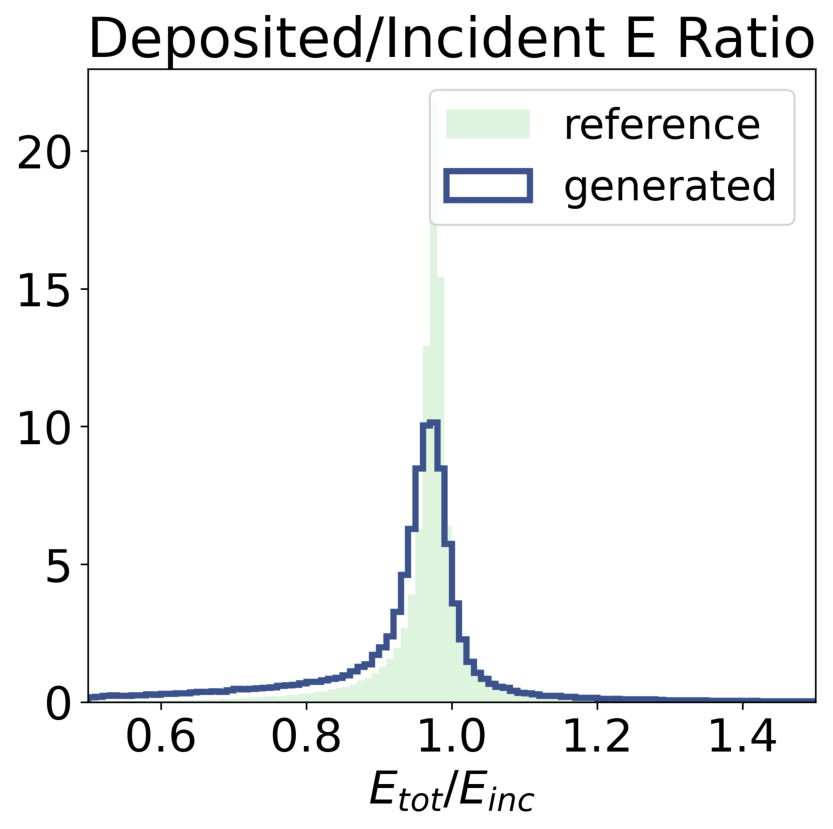}
        \includegraphics[width=0.16\textwidth]{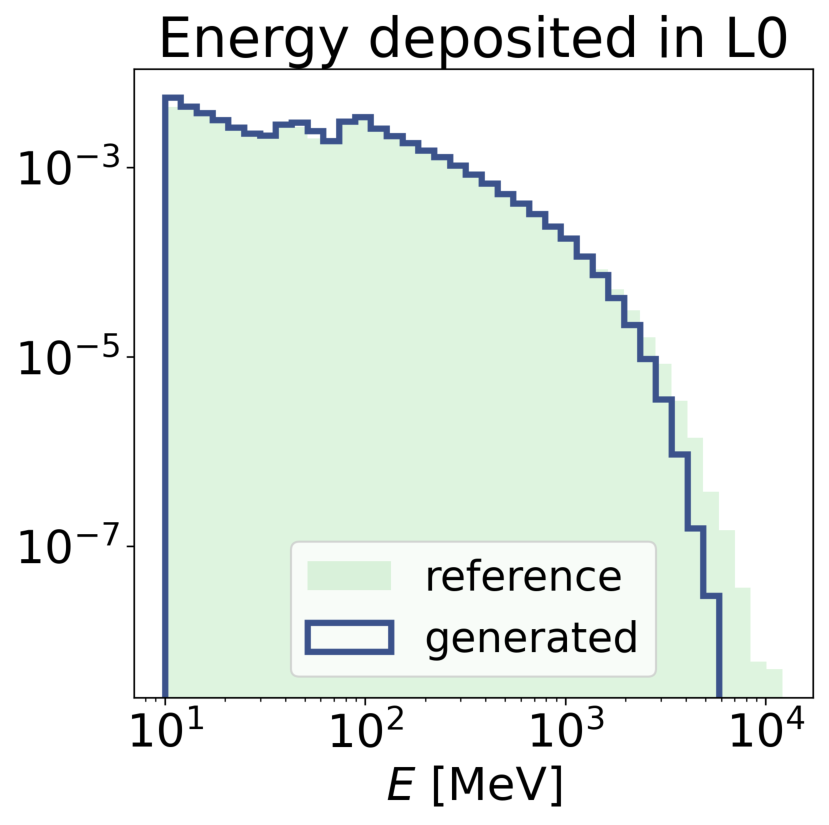}        \includegraphics[width=0.16\textwidth]{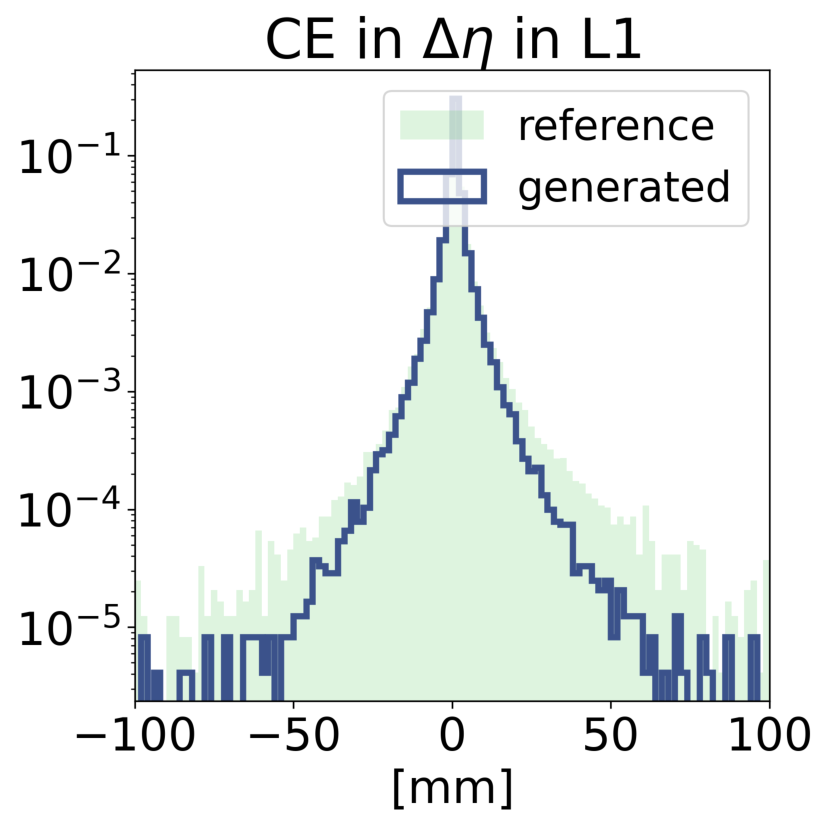}        \includegraphics[width=0.16\textwidth]{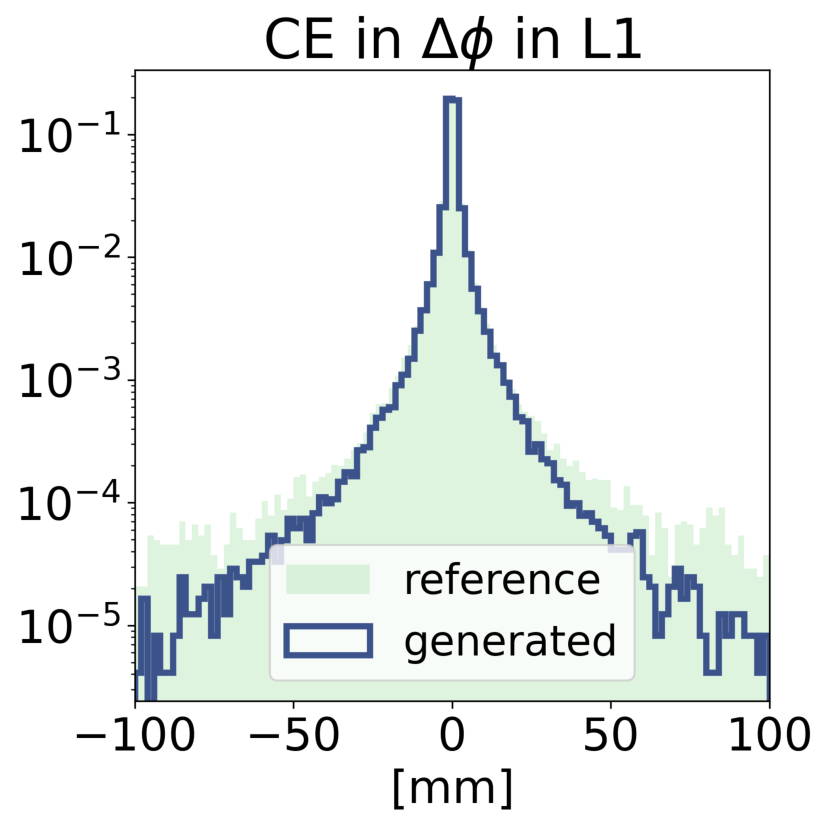}        \includegraphics[width=0.16\textwidth]{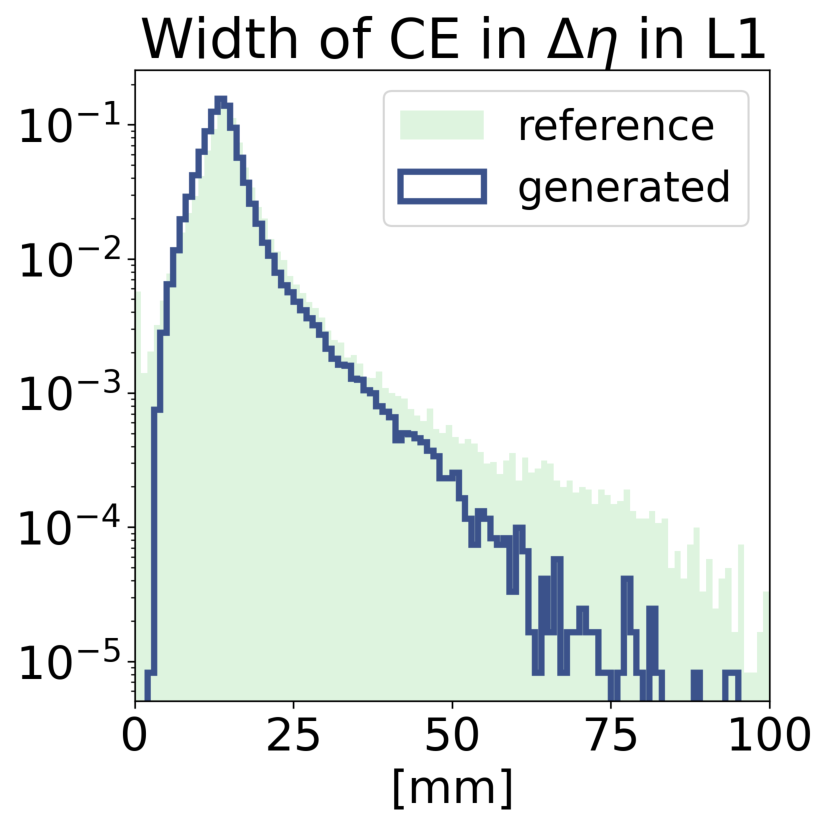}        \includegraphics[width=0.16\textwidth]{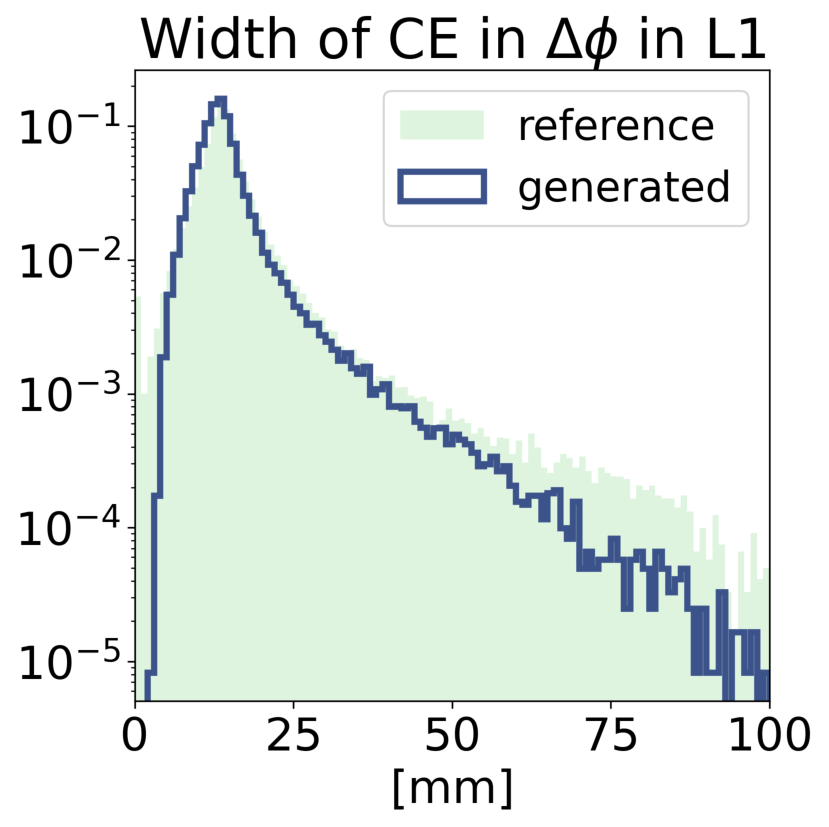}
    \caption{Histograms of high-level features comparing generated Photons samples to the test set. Note the log scale of the y-axis for all but the first plot.}
    \label{fig:histograms_photons}
    \vspace{-15pt}
\end{figure}

The Challenge uses three types of metrics to evaluate models: resource usage, especially generation time as discussed above; distributional closeness to the test set judged by the $\chi^2$ separation power between histograms in features crafted by domain experts; and ROCAUC of a binary classifier trained on a mix of real and generated data. We describe these metrics in full detail in Appendix \ref{app:calo_metrics}. The latter two types of metric are shown in Table \ref{tab:calo_photons} as compared to a NN-based approach \cite{cresswell2022caloman}. We see that ForestFlow produces more ``realistic'' datapoints in that they are harder for a classifier to distinguish from the test set. Feature histograms are shown in Figure \ref{fig:histograms_photons} which confirms an accurate representation of the true distribution. Complete results for all metrics are given in Appendix \ref{app:calo_results}).

\begin{table}[t!]
\caption{Model performance on calorimeter data. Lower is better.}  
\vspace{-2pt}
\label{tab:calo_photons}
\small
\centering
    \setlength{\tabcolsep}{3pt} 
\begin{tabular}{lrrrrrrr}
  \toprule
 \textbf{Photons} & AUC & $E_{\text{dep}}/ E_{\text{inc}}$ & $E_{\text{dep,L}0}$ & CE$_{\eta,\text{L}1}$ &  CE$_{\phi,\text{L}1}$ & Width$_{\eta,\text{L}1}$ & Width$_{\phi,\text{L}1}$  \\ \hline
\rule{0pt}{1.1\normalbaselineskip}CaloMan \cite{cresswell2022caloman} & 0.9998 & \better{0.0020} & \better{0.0001} & 0.0462 & 0.0394 & 0.0366 & 0.0865 \\
  CaloForest (Ours) & \better{0.8392} & 0.0778 & 0.0033 & \better{0.0056} & \better{0.0029} & \better{0.0241} & \better{0.0228}  \\
  \bottomrule
\toprule
 \textbf{Pions} & AUC & $E_{\text{dep}}/ E_{\text{inc}}$ & $E_{\text{dep,L}0}$ & CE$_{\eta,\text{L}1}$ &  CE$_{\phi,\text{L}1}$ & Width$_{\eta,\text{L}1}$ & Width$_{\phi,\text{L}1}$  \\ \hline
\rule{0pt}{1.1\normalbaselineskip}CaloMan \cite{cresswell2022caloman} & 0.9986 & \better{0.0404} & \better{0.0002} & 0.0477 & 0.0282 & 0.2380 & 0.2183\\
  CaloForest (Ours) & \better{0.9119} & 0.0625 &  0.0384 & \better{0.0268} & \better{0.0266} & \better{0.1935} &  \better{0.1978}    \\
  \bottomrule
  \end{tabular}
  \vspace{-5pt}
\end{table}

\section{Conclusions and Limitations}
\label{sec:conclusion}
\vspace{-5pt}
In this work, we have pushed the boundaries of tabular data generation backed not by neural networks, but by XGBoost. As discussed, XGBoost offers several tantalizing advantages over NNs for generative modelling: XGBoost's better performance on discriminative tabular tasks may translate to better tabular generation; it is robust without data pre-processing; it natively handles missing values; it can be efficiently trained on CPU; and finally it offers improved explainability. However, the great differences in the mechanics of XGBoost training compared to NNs led \citet{jolicoeur2023generating} to propose overparameterized models that do achieve state-of-the-art performance, but potentially at the cost of practicality and scalability. Our contributions re-engineered the inner workings of these models leading to faster training and generation, as well as much lower peak memory burdens and better scaling with dataset size, allowing us to train on datasets 370$\times$ bigger than previously tested. The improved implementation unlocked the ability to scale up model size which directly led to improved performance when properly regularized. Finally, we proposed the use of multi-output trees which are more suited to the high-dimensional outputs required in generative modelling, and showed that generation time can be reduced by an additional order of magnitude for applications that require large volumes of generated data. Still, it is clear that the methods we discuss have limitations. Models trained with ForestDiffusion and ForestFlow are highly overparameterized (Section \ref{sec:calo}), still require significant computational resources especially when $p$ is large (Figure \ref{fig:resources-random}), and our proposal to use multi-output trees comes at a cost of somewhat slower and more memory-intensive training (Figure \ref{fig:resources-random}).


\bibliographystyle{plainnat}
\bibliography{bib}


\newpage
\appendix

\section{Fast Calorimeter Simulation}
\label{app:calorimeter}

Particle accelerator experiments in high energy physics utilise several components in their detectors to measure properties of particles created in collisions. Calorimeters are one component that measure the energy of particles. Upon entering the calorimeter, the incident particle begins interacting with the material of the calorimeter and progressively deposits its energy. The interactions form a branching tree-like structure called a shower. Energy deposits are measured in an array of voxels allowing the 3d reconstruction of shower shape. Since nature is inherently probabilistic, a given incident particle with fixed energy gives rise to a probability distribution of possible showers.

Physicists desire to sample from these distributions as one component of detector simulation. By simulating detector responses using known theory, physicists can define their prior for what is expected to be measured when the actual experiment is run. Measured deviations from the prior expectation may indicate new physics, leading to a deeper understanding of nature.

However, sampling calorimeter showers using precise simulation of physical processes from first-principles is incredibly slow. Currently, simulations at the largest particle accelerator, the Large Hadron Collider (LHC), are done with Geant4 \cite{geant4, geant4-add1, geant4-add2}, which is CPU-based and can take upwards of ten minutes per shower. Since billions of simulations are needed to provide accurate background statistics, the computational burden of exact simulation is immense.

Generative modelling heralds a solution by directly generating showers using surrogate models instead of simulating them from first-principles. The first method in this line of research used GANs \cite{paganini2018calogan}, eventually leading to actual deployment of GAN-based generators in the experimental pipeline of the LHC \cite{aad2021atlfast3}. Following work explored other deep generative techniques like normalizing flows \cite{krause2021caloflow, krause2021caloflow2}.

\begin{figure}[t]
\centering
  \includegraphics[width=0.48\textwidth, trim={0 0 0 0}, clip]{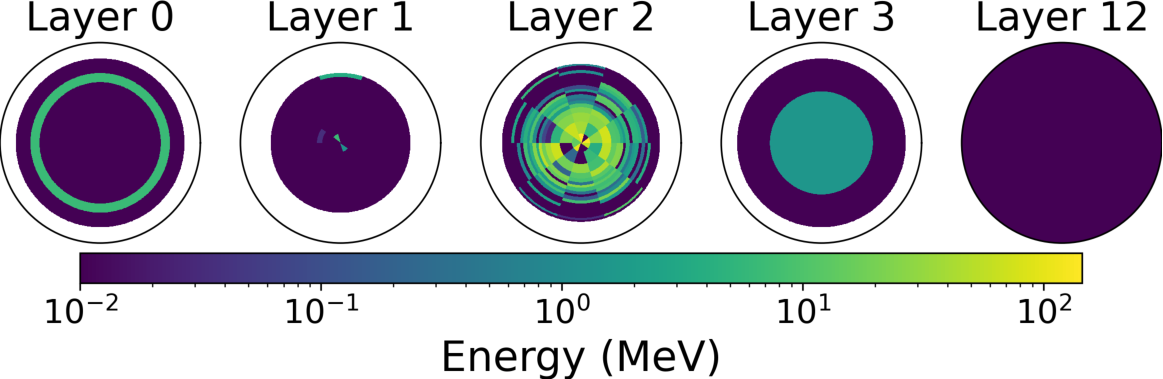} \quad 
  \includegraphics[width=0.48\textwidth, trim={0 0 0 0}, clip]{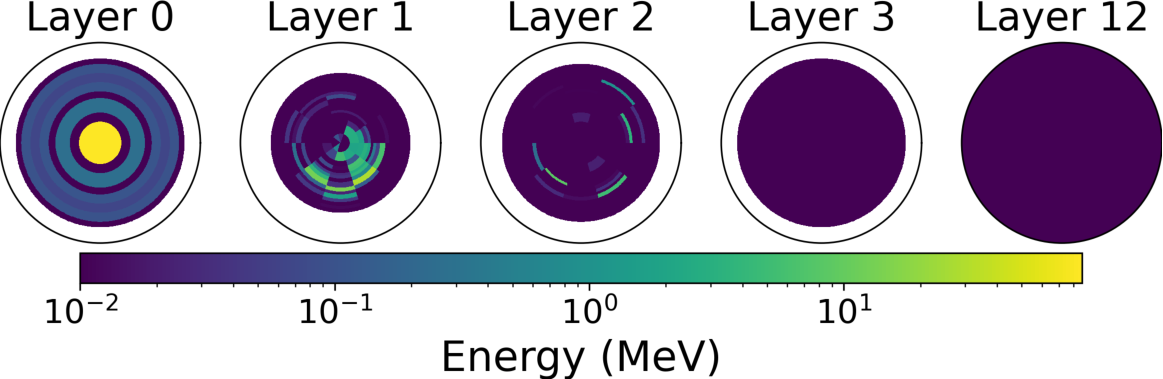}\\
  \vspace{4pt}
  \includegraphics[width=0.48\textwidth, trim={0 0 0 0}, clip]{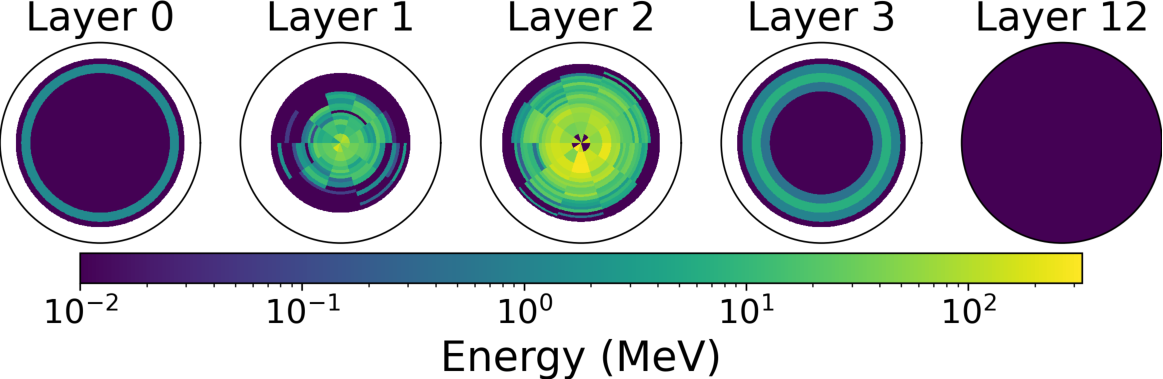} \quad 
  \includegraphics[width=0.48\textwidth, trim={0 0 0 0}, clip]{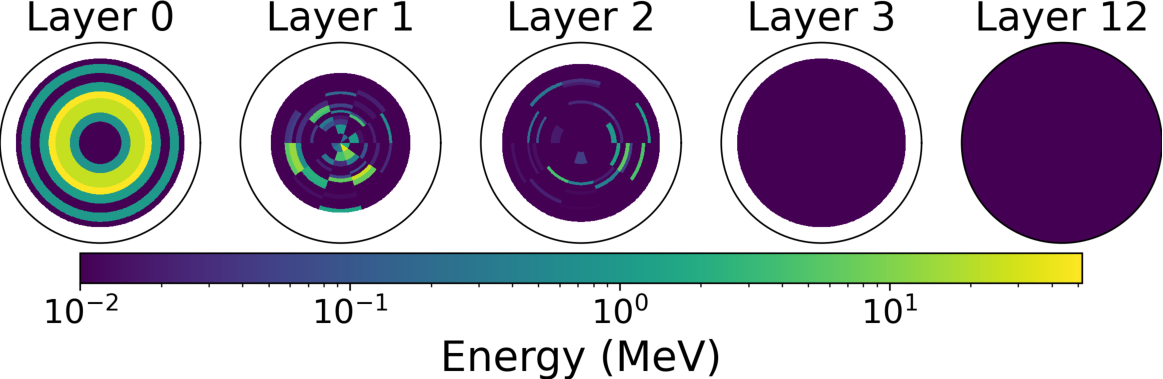}\\
    \vspace{4pt}
  \includegraphics[width=0.48\textwidth, trim={0 0 0 0}, clip]{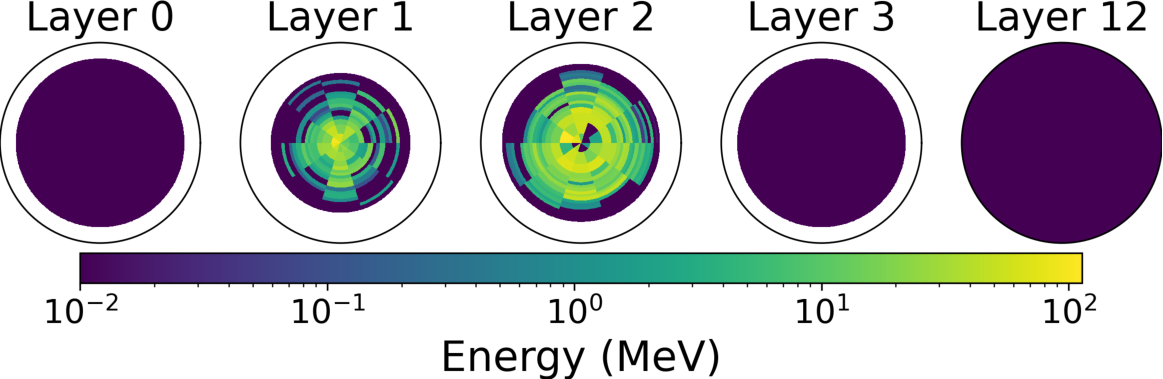} \quad 
  \includegraphics[width=0.48\textwidth, trim={0 0 0 0}, clip]{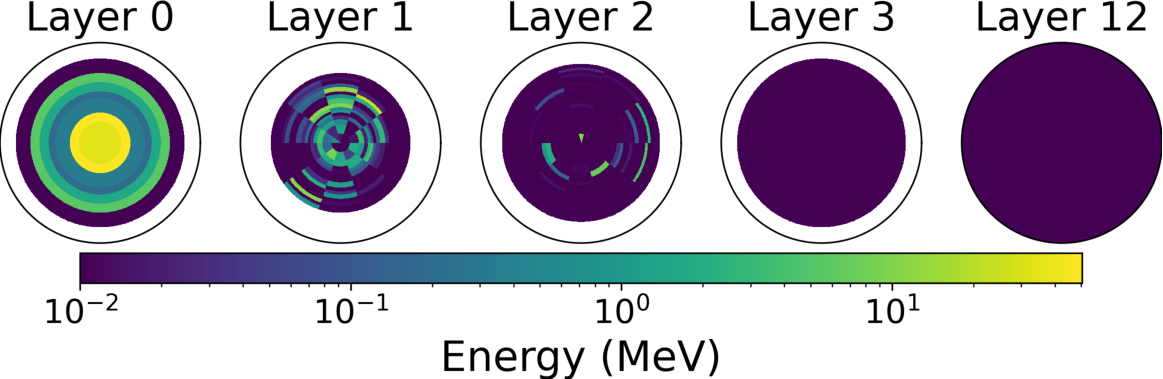}
    \caption{Individual showers shown as energy deposited per voxel for the Photons test dataset (left), and generated by CaloForest (right). Note the nested cylindrical geometry of voxels which is inconsistent across layers, meaning the data must be treated as tabular, rather than as images.}
    \label{fig:individual_showers}
    \vspace{-10pt}
\end{figure}

\begin{figure}[t]
\centering
  \includegraphics[width=0.4\textwidth, trim={0 0 0 0}, clip]{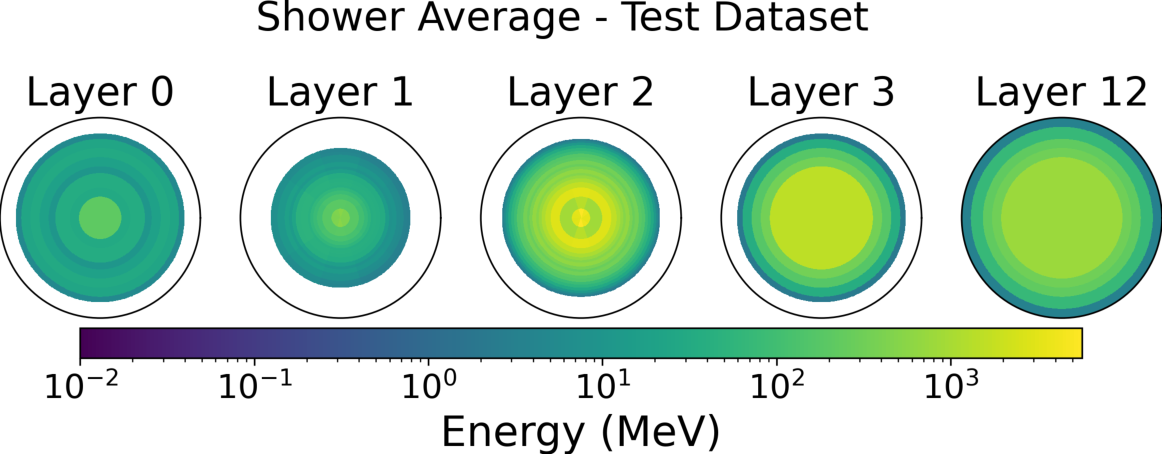} \quad
  \raisebox{0.018\height}{
  \includegraphics[width=0.56\textwidth, trim={0 0 0 0}, clip]{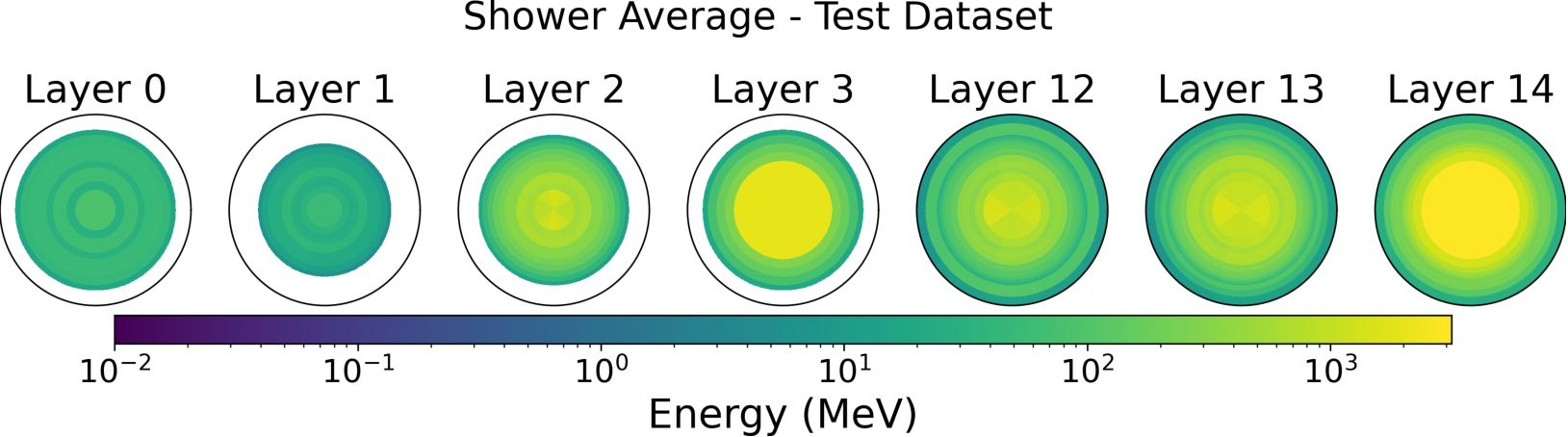}} \\
    \vspace{4pt}
    \includegraphics[width=0.4\textwidth, trim={0 0 0 0}, clip]{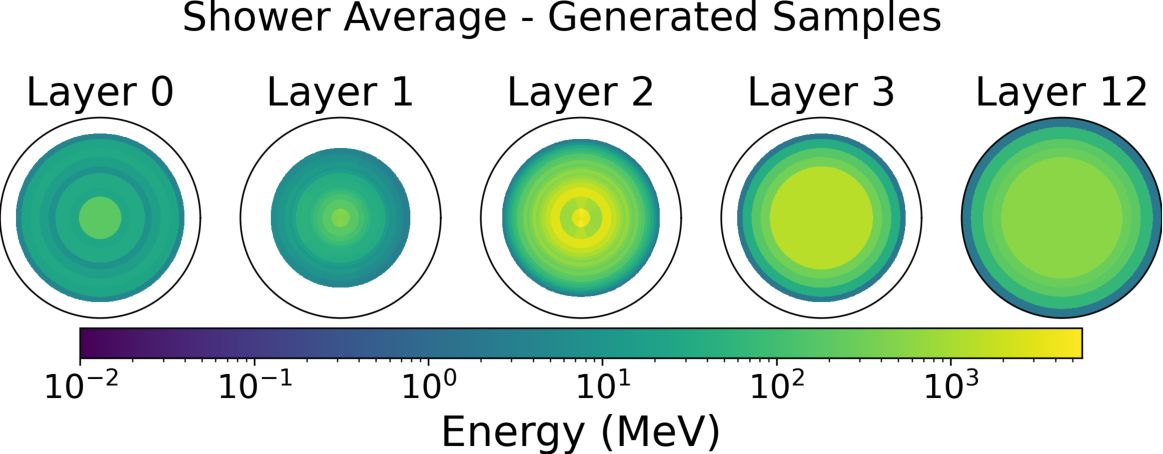}\quad
      \raisebox{0.016\height}{
  \includegraphics[width=0.56\textwidth, trim={0 0 0 0}, clip]{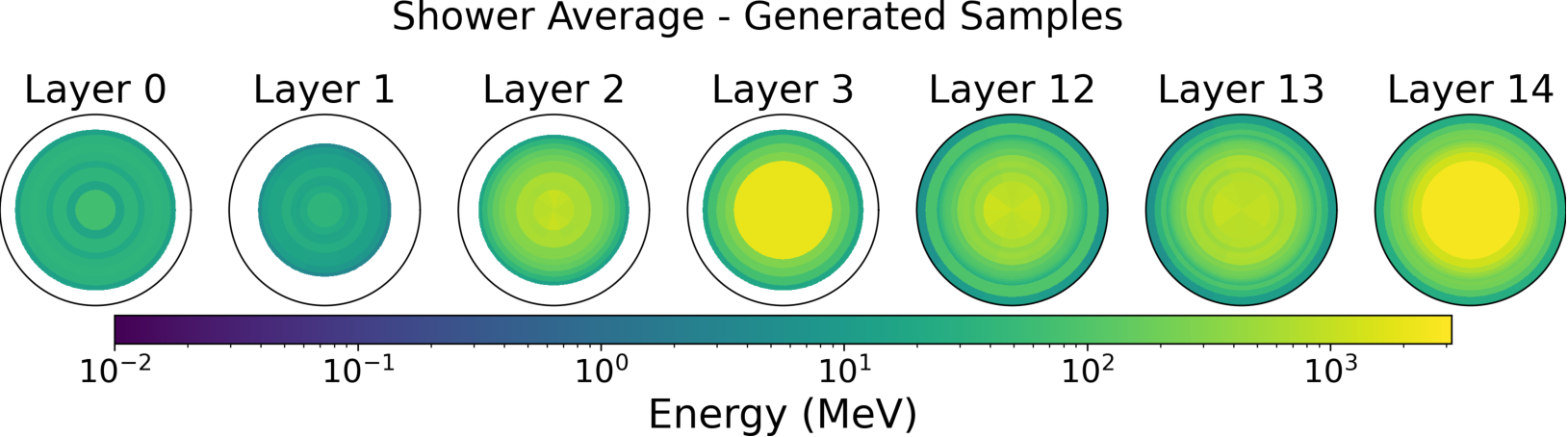}
  }
    \caption{Average deposited energy per voxel for Photons (left) and Pions (right) on the test dataset (top), and samples generated conditionally using the test set class distribution (bottom).}
    \label{fig:average_showers}
\end{figure}

The initial success of these methods at reducing simulation time, while accurately representing the distribution of showers, led to the public release of large-scale training datasets and a call to the community to explore new methods in the Fast Calorimeter Simulation Challenge \cite{calochallenge}. The four datasets involve different types of particles incident on the calorimeter. The datasets represent the calorimeters with a cylindrical pattern of voxels, and each datapoint's features represent the energy deposited in one voxel. This allows the visualization of individual showers (Figure \ref{fig:individual_showers}), as well as averages across the dataset (Figure \ref{fig:average_showers}).

Early submissions to the Challenge branched out to test score-based methods \cite{Mikuni:2022xry} and models that learned the low-dimensional structure of showers \cite{cresswell2022caloman}. Overall, submissions can be classified by the generative modelling paradigm they build off of, with GANs \cite{FaucciGiannelli:2023fow, Kach:2023rqw, scham2023deeptreegan}, normalizing flows \cite{Diefenbacher:2023vsw, Buckley:2023rez, Ernst:2023qvn, Schnake:2024mip}, diffusion models \cite{Amram:2023onf, Buhmann:2023bwk, kobylianskii2024calograph}, and conditional flow matching \cite{favaro2024calodream} being popular choices.

Notably, every prior submission to the Fast Calorimeter Simulation Challenge uses deep neural networks as function approximators. This is despite the need for GPU resources to train and generate with NNs, whereas existing scientific computing infrastructure for shower simulation is largely CPU-based. Our method, CaloForest, provides an alternative, as it is the only attempt to use tree-based approximators for the Challenge's large-scale tabular datasets.

\subsection{Evaluation Metrics}\label{app:calo_metrics}

Since the generated showers are meant to be used in actual scientific experiments, custom evaluation metrics have been defined for the Challenge using domain knowledge.

First, computational resources are important (hence ``Fast'' in the Challenge's title). Training should be accomplished with as little time and memory as possible, but the most important resource metric is shower generation time, as billions of generated showers will be needed in practice. Hence, we track the training time, generation time, and peak memory usage during training. These results are given in Section \ref{sec:calo}.

Second, generated showers must accurately represent the actual distribution of showers predicted by theory. Calculating this distribution in closed form from theory is not feasible, so instead the ground truth is taken from theory-based simulations using Geant4. A test dataset of showers generated in the same way as the training data is provided with each of the Challenge's datasets. Using domain knowledge, physicists defined high-level features from voxel-level information. The one-dimensional distributions of each feature can be compared between the test set and a generated set using the $\chi^2$ separation power between histograms which is defined as
\begin{equation}
    \chi^2(h_1, h_2) = \frac{1}{2}\sum_i \frac{(h_{1, i} - h_{2, 1})^2}{h_{1, i} + h_{2, i}},
\end{equation}
where $h_{j, i}$ is the fraction of all datapoints falling into bin $i$ of histogram $j$, such that $\sum_i h_{j, i}=1$. The metric is normalized such that $\chi^2(h_1, h_2)=0$ if and only if the histograms are the same, $h_1 = h_2$, whereas when the histograms have no overlap $\chi^2(h_1, h_2)=1$. The high-level features denote the ratio of deposited energy to incident energy, the total deposited energy in each layer of the calorimeter, the center of energy in angular directions $\eta$ and $\phi$ per layer, and the width of the center of energy in angular directions per layer. Example $\chi^2$ metrics are given in Table \ref{tab:calo_photons} with example histogram plots in Figures \ref{fig:histograms_photons} and \ref{fig:histograms_pions}, while the complete lists of metrics are shown below in Tables \ref{tab:hist_photons_full} and \ref{tab:hist_pions_full}.

Third, a binary classifier is trained to distinguish generated showers from the test set using the architecture and training details provided by the Challenge \cite{calochallenge}. The ROCAUC of the classifier on a balanced, held-out set of samples should be as low as possible, indicating that generated samples are indistinguishable from test datapoints. We present the ROCAUC metrics in Table \ref{tab:calo_photons}.

\subsection{Extended Results}\label{app:calo_results}

\begin{table}[t]
\caption{Photons dataset histogram $\chi^2$ separation powers in domain expert features. L denotes layer. CE is the center of energy. Lower is better.}\label{tab:hist_photons_full}
\centering
    \begin{tabular}{lll}
        \toprule
        \textsc{Feature} & CaloMan \cite{cresswell2022caloman} & CaloForest  \\
        \midrule
        $E_{\text{dep}}/ E_{\text{inc}}$ & 0.0020 & 0.0778 \\ 
        $E_{\text{dep}}$, \textsc{L}$0$ & 0.00005 & 0.0033 \\ 
        $E_{\text{dep}}$, \textsc{L}$1$ & 0.00008 & 0.0036 \\ 
        $E_{\text{dep}}$, \textsc{L}$2$ & 0.00002 & 0.0031 \\ 
        $E_{\text{dep}}$, \textsc{L}$3$ & 0.00001 & 0.0018 \\ 
        $E_{\text{dep}}$, \textsc{L}$12$ & 0.00002 & 0.0037 \\ 
        CE in $\eta$, L1 & 0.0462 & 0.0056 \\ 
        CE in $\eta$, L2 & 0.0419 & 0.0014 \\ 
        CE in $\phi$, L1 & 0.0394 & 0.0029 \\ 
        CE in $\phi$, L2 & 0.0367 & 0.0017 \\ 
        Width in CE in $\eta$, L1 & 0.0366 & 0.0241 \\ 
        Width in CE in $\eta$, L2 & 0.0696 & 0.0108 \\ 
        Width in CE in $\phi$, L1 & 0.0865 & 0.0228 \\ 
        Width in CE in $\phi$, L2 & 0.0649 & 0.0097 \\ 
        \bottomrule
    \end{tabular}
\end{table}

\begin{table}[t]
\caption{Pions dataset histogram $\chi^2$ separation powers in domain expert features. L denotes layer. CE is the center of energy. Lower is better.}\label{tab:hist_pions_full}
\centering
    \begin{tabular}{lll}
        \toprule
        \textsc{Feature} & CaloMan \cite{cresswell2022caloman} & CaloForest  \\
        \midrule
        $E_{\text{dep}}/ E_{\text{inc}}$ & 0.0404 & 0.0625 \\ 
        $E_{\text{dep}}$, \textsc{L}$0$ & 0.0002 & 0.0384 \\ 
        $E_{\text{dep}}$, \textsc{L}$1$ & 0.0347 & 0.1440 \\ 
        $E_{\text{dep}}$, \textsc{L}$2$ & 0.0052 & 0.0532 \\ 
        $E_{\text{dep}}$, \textsc{L}$3$ & 0.0001 & 0.0178 \\ 
        $E_{\text{dep}}$, \textsc{L}$12$ & 0.0008 & 0.0046 \\
        $E_{\text{dep}}$, \textsc{L}$13$ & 0.0001 & 0.0102 \\
        $E_{\text{dep}}$, \textsc{L}$14$ & 0.0002 & 0.0085 \\
        CE in $\eta$, L1 & 0.0477 & 0.0268 \\ 
        CE in $\eta$, L2 & 0.0808 & 0.0168 \\ 
        CE in $\eta$, L12 & 0.0477 & 0.0641 \\ 
        CE in $\eta$, L13 & 0.0808 & 0.1377 \\
        CE in $\phi$, L1 & 0.0282 & 0.0266 \\ 
        CE in $\phi$, L2 & 0.0240 & 0.0155 \\ 
        CE in $\phi$, L12 & 0.0282 & 0.0573 \\ 
        CE in $\phi$, L13 & 0.0240 & 0.1203 \\ 
        Width in CE in $\eta$, L1 & 0.2380 & 0.1935 \\ 
        Width in CE in $\eta$, L2 & 0.2074 & 0.1121 \\
        Width in CE in $\eta$, L12 & 0.2380 & 0.1758 \\ 
        Width in CE in $\eta$, L13 & 0.2074 & 0.2384 \\ 
        Width in CE in $\phi$, L1 & 0.2183 & 0.1978 \\ 
        Width in CE in $\phi$, L2 & 0.2067 & 0.1141 \\
        Width in CE in $\phi$, L12 & 0.2183 & 0.1788 \\ 
        Width in CE in $\phi$, L13 & 0.2067 & 0.2399 \\ 
        \bottomrule
    \end{tabular}
\end{table}

Here we present complete information on the histogram evaluation metrics for the Challenge that are obtained with our method CaloForest. Tables \ref{tab:hist_photons_full} and \ref{tab:hist_pions_full} show the $\chi^2$ separation powers for histograms of the generated and test set samples. Compared to a NN-based approach designed for the challenge called CaloMan \cite{cresswell2022caloman}, CaloForest better captures the distribution of Centers of Energy and their Widths. CaloMan was designed with a separate module to predict the deposited energy in each layer, and thus has better performance in those metrics.

\begin{figure}[t]
    \centering
        \includegraphics[width=0.155\textwidth]{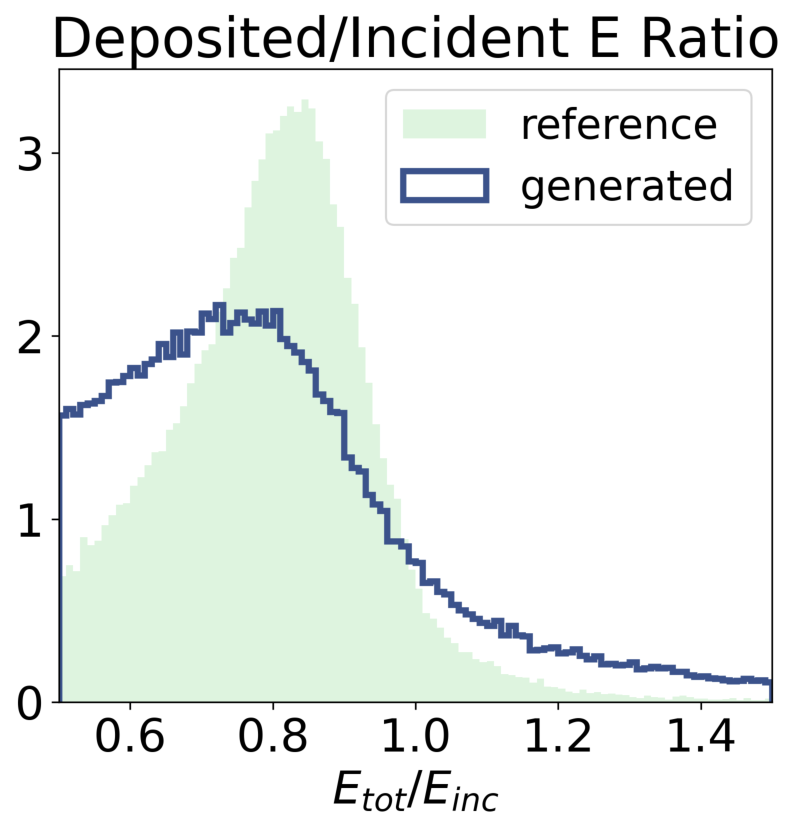}
        \includegraphics[width=0.16\textwidth]{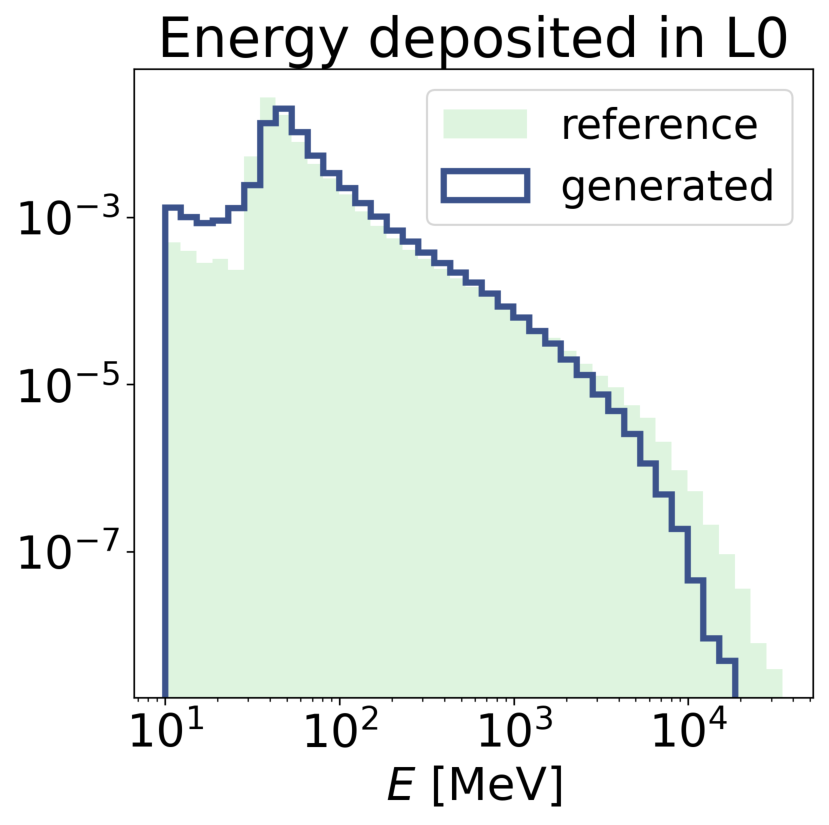}        \includegraphics[width=0.16\textwidth]{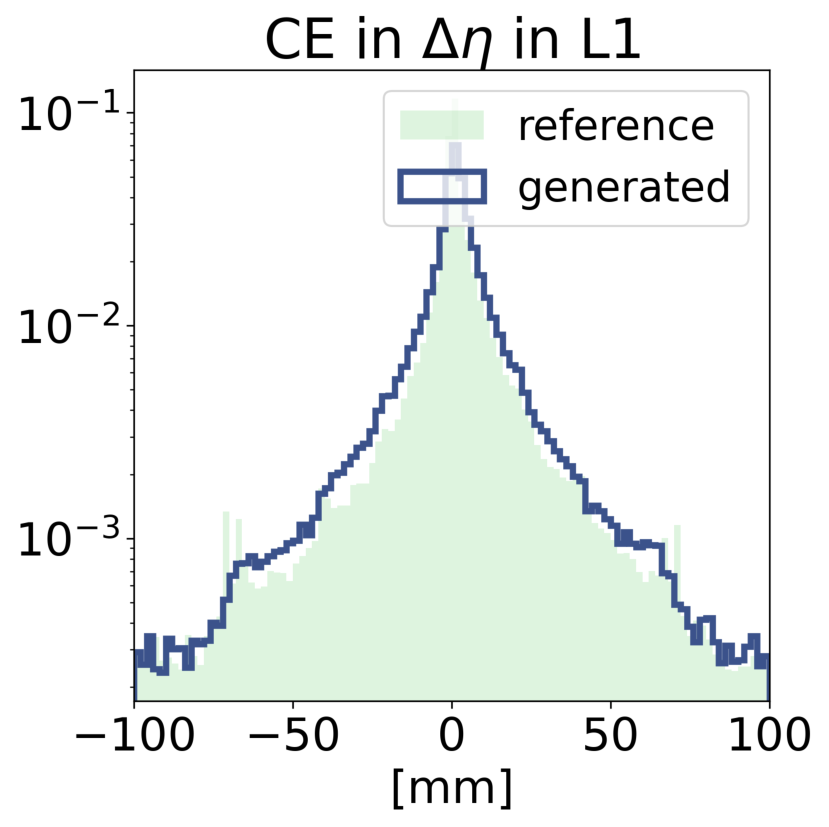}        \includegraphics[width=0.16\textwidth]{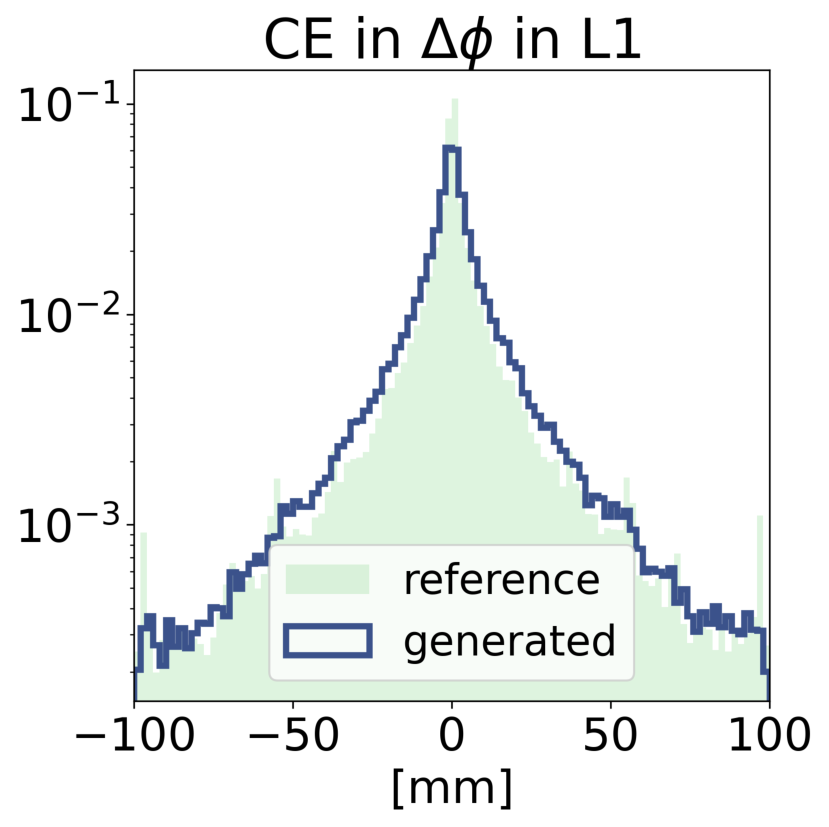}        \includegraphics[width=0.16\textwidth]{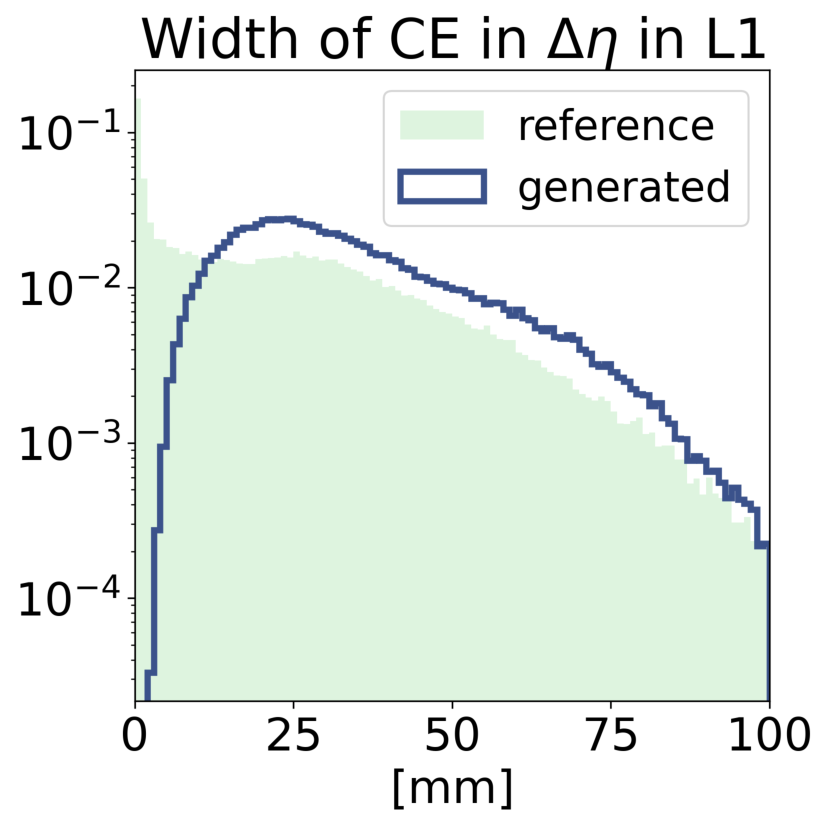}        \includegraphics[width=0.16\textwidth]{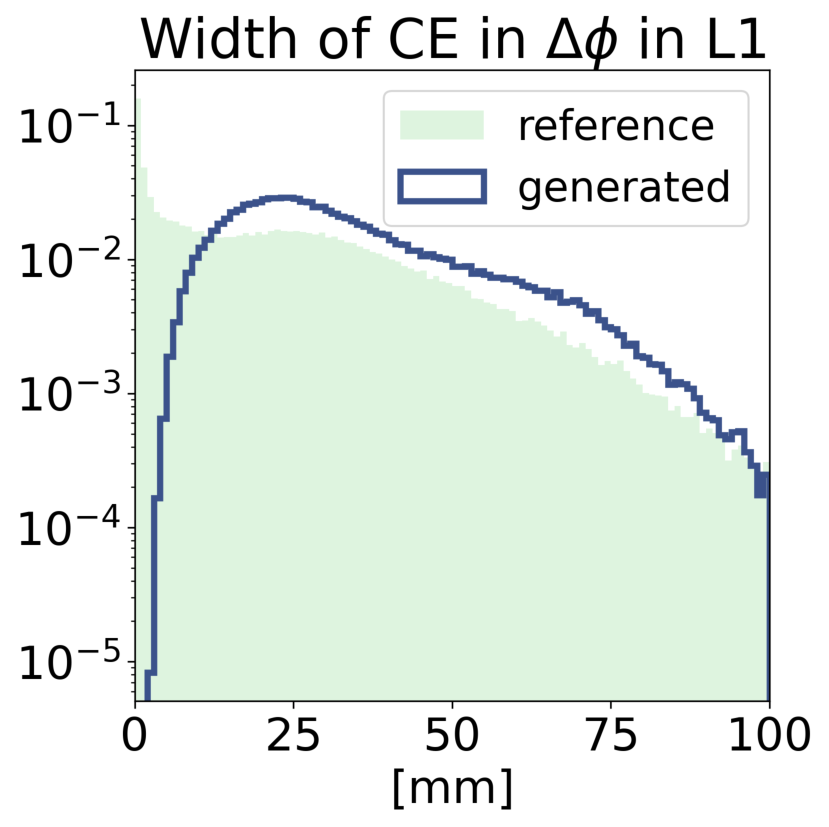}
    \caption{Histograms of high-level features comparing generated Pions samples to the test set. Note the log scale of the y-axis for all but the first plot.}
    \label{fig:histograms_pions}
\end{figure}

We also add feature histogram plots for the Pions dataset in Figure \ref{fig:histograms_pions} to complement the Photons results shown in the main text (Figure \ref{fig:histograms_photons}).

As mentioned above, average per-voxel energy deposits are shown for the models trained on both datasets in Figure \ref{fig:average_showers} where it is clear that these distributions are learned almost perfectly.

\section{Re-engineering the ForestDiffusion and ForestFlow Implementation}
\label{app:implementation}

In this Appendix we continue our analysis of the original implementation of ForestDiffusion and ForestFlow training (Algorithm \ref{alg:train}) provided by  \cite{jolicoeur2023generating}, and present a summary of our improvements as a unified implementation. We further consider how to optimize data generation.

First, we comment on the pros and cons of parallelization. Most of the memory issues experienced when using ForestDiffusion and ForestFlow are a result of training many XGBoost ensembles in parallel. Parallelization using multiprocessing requires copying data arrays to worker processes, so one may ask about alternatives. Apart from multiprocessing, parallelism in Python can also be achieved through multithreading. This avoids spawning new processes with their own memory spaces and can allow threads to share memory in the main process, however, due to the Python global interpreter lock (GIL), multithreading can only be done on tasks that release the GIL while running. In fact, calls to XGBoost training do release the GIL, as XGBoost runs native C++ code, so multithreading is a potential solution for ForestDiffusion and ForestFlow. However, in our preliminary tests we found that multithreading was more prone to memory not being properly released, causing increased usage over training. We were not able to find a definitive reason for this but we suspect that the Python garbage collector in a multithreaded process  does not effectively free up used memory. On the other hand, we observed that multiprocessing is very effective for releasing all used memory when the corresponding job is completed.

\begin{wrapfigure}[15
]{r}{0.45\textwidth}
\vspace{-15pt}
    \centering\includegraphics[width=0.34\textwidth, trim={0 0 0 0}, clip]{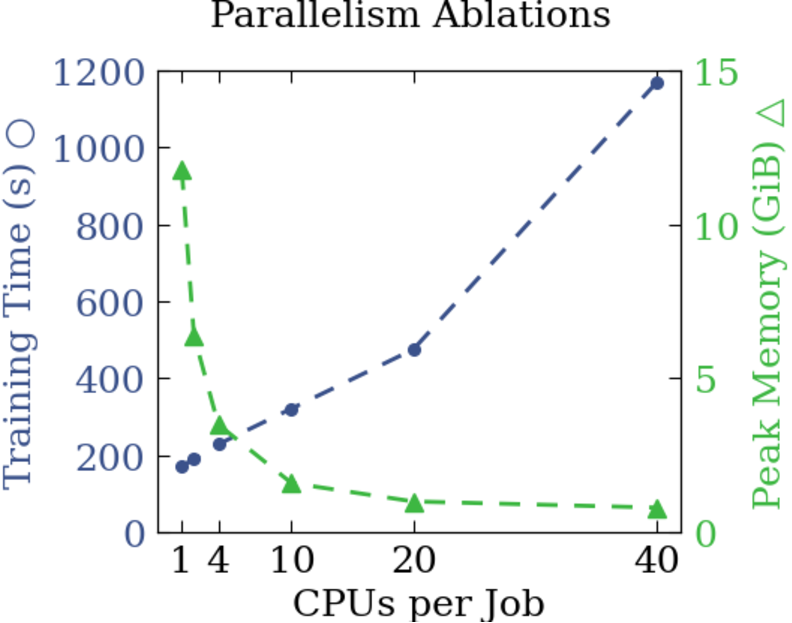}
    \vspace{-5pt}
    \caption{XGBoost is not efficient with multiple CPUs, especially for small datasets. Running single-CPU jobs in parallel is the most time-efficient method as long as adequate memory is available.}
    \label{fig:cpu_efficiency}
\end{wrapfigure}
Alternatively, one may wonder why creating parallel jobs is necessary at all when a single XGBoost training job can make use of multiple CPUs. In fact XGBoost is not perfectly efficient in its use of additional CPUs, especially on small datasets like those from Table \ref{tab:datasets}; training time is reduced by less than 50\% when two CPUs are used instead of one, and this efficiency becomes even worse as more CPUs are provided. Hence, there is a tradeoff between speed and memory: assigning all CPUs to a single job uses the least memory but may be slow, whereas assigning one CPU to $N$ training jobs will be much faster, while using roughly $N$ times as much memory. Figure \ref{fig:cpu_efficiency} demonstrates this by training our implementation of ForestFlow on a dataset with $n=1000$, $p=10$, and $n_y=10$ for various assignments of CPUs per job (cf. Figure \ref{fig:resources-random}). On our machine with 40 CPUs, we set the number of CPUs assigned to each job to the values \{1, 2, 4, 10, 20, 40\} (and correspondingly set the number of parallel jobs to \{40, 20, 10, 4, 2, 1\}). Generally, the number of parallel jobs times the number of CPUs assigned per job should not exceed the number of CPUs available in total, otherwise thread contention can degrade performance. When memory is a limitation, Figure \ref{fig:cpu_efficiency} shows that assigning a few CPUs per job and reducing the number of jobs can greatly decrease peak memory requirements at a marginal increase in training time. Hence, we used two CPUs per job when training on large-scale calorimeter data.

For the sake of our resource benchmarking across methods, we always use multiprocessing, and assign one CPU per worker, with the number of workers equal to the available CPUs.

\subsection{Continued Analysis and Improvement of the Implementation}\label{app:continued_issues}

In Section \ref{sec:analysis} we began our analysis with the most impactful issues and solutions. We recommended to avoid creating large arrays in memory and instead create slices on-the-fly as needed within parallel loops (Issue 1). We found that improper use of multiprocessing could lead to excessive duplication of arrays in shared memory that could not be freed by the system, and explained how to properly share an array across processes (Issue 2). Finally, we recommended to write XGBoost models to disk as their training completed to prevent them piling up in active memory (Issue 3). These three issues and solutions accounted for the vast majority of memory improvements we observed and explained the three problematic behaviours pointed out in Section \ref{sec:limitations}. Nevertheless, we pick up where we left off and present several additional improvements that further optimize memory usage and runtime while adhering to engineering best-practices.

\textbf{Issue 4:} Since worker processes created by Joblib access data saved to shared memory, the main process does not need to occupy memory by holding on to its copy of \texttt{X0}, \texttt{X1}, and \texttt{Z\_train}. These costly array objects are merely used as keys for Joblib to identify the arrays in shared memory.
\\
\textbf{Solution 4:} Explicitly save the arrays in shared memory as memory-mapped files, delete them from the main process, and retain only a reference which can be passed to worker processes.\\
\textbf{Benefit 4:} The \texttt{X0}, \texttt{X1}, and \texttt{Z\_train} objects can be freed from the main process, amounting to \textbf{144 GiB} for the Pions dataset. Technically, we save memory-mapped files on a disk instead of a RAM disk. While a RAM disk occupies RAM space, saving to an actual disk does not, which leaves more available memory during training. Nonetheless, this does not cause slow downs from disk I/O. When a file is saved to a disk, it is first saved to cache memory (part of RAM). When the file is accessed again (potentially by a different process), if it is already in cache, the file in cache is reused. Unlike in-use RAM disk memory, this kind of cache memory can be freed upon memory pressure as it merely serves as cache for a disk.

\begin{mintedbox}{python}{Issue 4: Improvement}
import tempfile, os
from joblib import dump, load
# Create memmap files
temp_folder = tempfile.mkdtemp()
def create_memmap(array, file_name):
  file = os.path.join(temp_folder, file_name)
  dump(array, file)
  return load(file, mmap_mode='r')
X0_mmap = create_memmap(X0, "X0.mmap")
X1_mmap = create_memmap(X1, "X1.mmap")
Z_mmap = create_memmap(Z_train, "Z.mmap")
# Free memory of X0, X1, Z_train
del X0, X1, Z_train
\end{mintedbox}

\textbf{Issue 5:} Using \texttt{n\_y} Boolean masks across the duplicated dataset to index each class’s data requires \texttt{n*K*n\_y} bytes, since the \texttt{numpy.bool} datatype uses one byte, not one bit. Moreover, indexing a Numpy array with the mask array (advanced indexing) creates a copy of the underlying data, as done for example in Line 26 of the original implementation (\texttt{X\_train[t\_i][mask[y\_i], :]}).\\
\textbf{Solution 5:} First sort the data by class, then use Python's \texttt{slice(start, end)} function with the beginning and end indices of each class.\\
\textbf{Benefit 5:} On the Pions dataset these Boolean masks would occupy \textbf{173 MiB} of space. Our solution only requires \texttt{2*n\_y} integers, and when used for indexing only creates a view, not a copy of underlying data. However, our solution does involve sorting, though the time involved is minuscule compared to the training time for hundreds of thousands of XGBoost ensembles.

\begin{mintedbox}{python}{Issue 5: Original}
# Create Boolean masks for class conditioning
mask = {} # Boolean mask for which rows of X0 have label y_i
y_uniq = numpy.unique(y)
for y_i in y_uniq:
  mask[y_i] = numpy.tile(y == y_i, K)
\end{mintedbox}

\begin{mintedbox}{python}{Issue 5: Improvement}
# Sort by label and slice for class conditioning
y_arg_sort = numpy.argsort(y)
y, X0 = y[y_arg_sort], X0[y_arg_sort]
y_uniq, y_counts = numpy.unique(y, return_counts=True)
mask = {} # Slice of X0's rows that have label y_i
csum = 0
for y_i, count in zip(y_uniq, y_counts):
  mask[y_i] = slice(csum, csum + count)
  csum += count
y_slice = {} # adjust slices for duplicated data
for y_i, sl in mask.items():
    y_slice[y_i] = slice(sl.start*K, sl.stop*K)
\end{mintedbox}

\textbf{Issue 6:} In XGBoost training, input data is converted to a \texttt{DMatrix}, XGBoost's native data structure, and is reformatted and cached for faster access. For example, features are converted to histograms when using \texttt{hist} training, as we do. The histogram computations are redundant across jobs since the same \texttt{X\_tr\_i} is used for all \texttt{p\_i}.\\
\textbf{Solution 6:} Ever since XGBoost version 1.6, multiple regressors trained with the same features but different targets can be encapsulated in a single \texttt{Booster} object. When the multi-dimensional target is passed to XGBoost's \texttt{fit(X,Z)} function, XGBoost internally trains an ensemble on each target sequentially while using the same \texttt{DMatrix}, avoiding redundant histogram computations over \texttt{p\_i}. Multi-output trees use the same data structure and training call, so can be integrated seamlessly.\\
\textbf{Benefit 6:} This reduces \texttt{DMatrix} constructions and reduces histogram computations by a factor of \texttt{p}. Additionally, all ensembles over \texttt{p} for a given \texttt{t} and \texttt{y} are contained in the same \texttt{Booster} object, which, in turn, reduces the number of model files and metadata to be stored, and reduces file I/O overhead. 

\begin{minipage}[t]{0.49\linewidth}%
\vspace{-10pt}
\begin{mintedbox}{python}{Issue 6: Original}
# One Booster for each column p_i
Z_tr_i = Z_train[mask[y_i], p_i]
model.fit(X_tr_i, Z_tr_i)
\end{mintedbox}
\end{minipage}%
\hfill
\begin{minipage}[t]{0.51\linewidth}%
\vspace{-10pt}
\begin{mintedbox}{python}{Issue 6: Improvement}
# Single Booster for all columns p_i
Z_tr_i = Z_train[y_slice[y_i], :]
model.fit(X_tr_i, Z_tr_i)
\end{mintedbox}
\end{minipage}%

\textbf{Issue 7:} XGBoost internally uses \texttt{fp32} regardless of the input data type. However, \texttt{numpy.float64} is implicitly used in the original implementation.\\
\textbf{Solution 7:} Use \texttt{fp32} throughout the whole pipeline.\\
\textbf{Benefit 7:} Using lower precision throughout reduces memory usage without losing model accuracy and avoids implicit data type conversions.

\begin{minipage}[t]{0.5\linewidth}%
\begin{mintedbox}{python}{Issue 7: Original}
X0 = inputs()
# loaded as numpy.float64
X1=numpy.random.normal(size=X0.shape)
# default dtype is numpy.float64
\end{mintedbox}
\end{minipage}%
\hfill
\begin{minipage}[t]{0.5\linewidth}%
\begin{mintedbox}{python}{Issue 7: Improvement}
X0 = inputs()
X0 = X0.astype(numpy.float32)
X1=numpy.random.normal(size=X0.shape)
X1 = X1.astype(X0.dtype)
\end{mintedbox}
\end{minipage}%

\clearpage
To summarize, our implementation making use of all our recommended changes is given below.

\begin{mintedbox}{python}{Our Implementation of ForestFlow Training with Single-Output Trees}
from sklearn.preprocessing import MinMaxScaler
import numpy, xgboost, tempfile, os
from joblib import delayed, Parallel, dump, load

X0, y, K, n_t, xgb_kwargs, n_jobs = inputs()
n, p = X0.shape
X0 = X0.astype(numpy.float32) # use XGBoost's native dtype
# Sort by label and slice for class conditioning
y_arg_sort = numpy.argsort(y)
y, X0 = y[y_arg_sort], X0[y_arg_sort]
y_uniq, y_counts = numpy.unique(y, return_counts=True)
mask = {} # Slice of X0's rows that have label y_i
csum = 0
for y_i, count in zip(y_uniq, y_counts):
  mask[y_i] = slice(csum, csum + count)
  csum += count
# Scale each class's data so that range matches noise variance
scalers = []
for y_i in y_uniq:
  scalers.append(MinMaxScaler(feature_range=(-1, 1))
  X0[mask[y_i], :] = scalers[-1].fit_transform(X0[mask[y_i], :])
# Duplicate data and generate noise
X0 = numpy.repeat(X0, K, axis=0)
X1 = numpy.random.normal(size=X0.shape).astype(X0.dtype)
y_slice = {} # adjust slices
for y_i, sl in mask.items():
    y_slice[y_i] = slice(sl.start*K, sl.stop*K)
# Create regression targets (ForestFlow)
Z_train = X1 - X0 # regression target is constant in t, but input is not
t = numpy.linspace(0, 1, num=n_t)
# Create memmap files
temp_folder = tempfile.mkdtemp()
def create_memmap(array, file_name):
  file = os.path.join(temp_folder, file_name)
  dump(array, file)
  return load(file, mmap_mode='r')
X0_mmap = create_memmap(X0, "X0.mmap")
X1_mmap = create_memmap(X1, "X1.mmap")
Z_mmap = create_memmap(Z_train, "Z.mmap")
del X0, X1, Z_train
# Train models in double loop over timesteps and classes
def train_parallel(X0_mmap, X1_mmap, Z_mmap, t_i, y_i):
  X_tr_i = t_i*X1_mmap[y_i, :] + (1-t_i)*X0_mmap[y_i, :]
  Z_tr_i = Z_mmap[y_i, :]
  model = xgboost.XGBRegressor(**xgb_kwargs)
  model.fit(X_tr_i, Z_tr_i) # single Booster for all columns p_i
  model.save_model(f"{model_path}.ubj") # path for t_i, y_i
Parallel(n_jobs)(
  delayed(train_parallel)(
    X0_mmap, X1_mmap, Z_mmap, t_i, y_i,
  ) for t_i in t for y_i in y_slice
)
shutil.rmtree(temp_folder) # clean up memmaps
\end{mintedbox}

For a direct comparison, we show in Figure \ref{fig:usage_1} the memory usage during training using the original implementation as well as ours on the same dataset with $n=1000$, $p=100$, and $n_y=10$. Our implementation does not suffer from the three undesirable behaviours noted in Section \ref{sec:limitations}.

\subsection{Analysis and Improvement of Data Generation}\label{app:generation}

To this point we have focused on improving the implementation of ForestFlow training. For many applications generation speed is also a critical requirement, including hosted generative model services and our running example of calorimeter simulation for experimental particle physics. In this section we turn our attention to improving the implementation of data generation with a trained ForestFlow model, starting with a summary of the existing implementation from \cite{jolicoeur2023generating}.

First, for conditional sampling, labels are created using a multinomial distribution with probabilities equal to the relative prevalence of labels in the training set. Boolean masks are created to indicate the conditioning. Gaussian noise \texttt{X1} is sampled to seed the generation, and Euler's method over uniformly discretized timesteps is used to solve the ODE using the trained models as the vector field. In particular, a triple \texttt{for} loop is used over timesteps, classes, and features in that order.

\begin{mintedbox}{python}{Python Implementation of ForestFlow Generation from \cite{jolicoeur2023generating}}
import numpy

y, n_t, n, p, regressors = inputs()
# Sample labels for conditioning using frequencies from the training dataset
y_uniq, y_counts = numpy.unique(y, return_counts=True)
y_probs = y_counts / numpy.sum(y_counts)
y_sample = numpy.argmax(numpy.random.multinomial(1, y_probs, size=n), axis=1)
# Create Boolean masks for class-conditioning
label_y = y_uniq[y_sample]
mask = {}
for y_i in y_uniq:
  mask[y_i] = (label_y == y_i)
# Solve ODE with Euler's method starting from noise
X1 = numpy.random.normal(size=(n, p))
h = 1 / (n_t-1) # size of timestep
for t_i in numpy.linspace(1, 0, num=n_t):
  out = numpy.zeros(shape=X1.shape)
  for y_idx, y_i in enumerate(y_uniq):
    for p_i in range(p):
      model = regressors[t_i][y_idx][p_i]
      out[mask[y_i], p_i] = model.predict(X1[mask[y_i], :])
  X1 = X1 - h * out
X0 = X1
\end{mintedbox}

Once again, we proceed by pointing out issues, offering solutions, and quantifying the benefits.

\textbf{Issue 8:} XGBoost's core engine is implemented in C++, and there is hidden overhead when the Python wrapper makes a call to its C-API.\\
\textbf{Solution 8:} Reduce the number of calls to the C-API by reducing the number of distinct \texttt{Booster} objects. In our training implementation, all ensembles trained over \texttt{p} for a given \texttt{t} and \texttt{y} are contained in the same \texttt{Booster} object (See Issue 6 in Appendix \ref{app:continued_issues}). Inference on this \texttt{Booster} object produces an output shape with \texttt{[n\_i, p]} containing all features.\\
\textbf{Benefit 8:} A factor of \texttt{p} fewer calls to the XGBoost C-API are made, and we eliminate Python's slow \texttt{for} loop over \texttt{p}. Additionally, cache locality is utilized more aggressively by XGBoost's C++ inference implementation.

\begin{minipage}[t]{0.52\linewidth}%
\begin{mintedbox}{python}{Issue 8: Original}
for p_i in range(p):
  model = regressors[t_i][y_idx][p_i]
  out[mask[y_i], p_i] =
    model.predict(X1[mask[y_i], :])
\end{mintedbox}
\end{minipage}%
\hfill
\begin{minipage}[t]{0.48\linewidth}%
\begin{mintedbox}{python}{Issue 8: Improvement}
model = regressors[t_i][y_idx]
out[mask[y_i], :] =
 model.predict(X1[mask[y_i], :])
\end{mintedbox}
\end{minipage}%

\textbf{Issue 9:} Slow Numpy indexing operations are used in the triple loop.\\
\textbf{Solution 9:} Conditional generation of datapoints with different \texttt{y} labels uses disjoint sets of ensembles. It is not necessary to combine all partially generated datapoints over classes into a single array after every timestep. Instead, we can iterate over \texttt{y\_i} in the outer loop and concatenate all the generated datapoints over classes only at the end.\\
\textbf{Benefit 9:} This eliminates writing intermediate results to non-contiguous memory \texttt{out[mask[y\_i]]}. Data copying is further avoided by replacing the Boolean mask with \texttt{slice} as in Issue 5 from Appendix \ref{app:continued_issues}.

\begin{mintedbox}{python}{Issue 9: Improvement}
results = []
for y_idx, y_i in enumerate(y_uniq):
  X1_i = X1[y_slice[y_i], :]
  for t_i in numpy.linspace(1, 0, num=n_t):
    model = regressors[t_i][y_idx]
    X1_i = X1_i - h * model.predict(X1_i)
  results.append(X1_i)
X0 = numpy.concatenate(results, axis=0)
\end{mintedbox}

To summarize, our implementation for ForestFlow generation making use of our recommended changes is given below.

\begin{mintedbox}{python}{Our Implementation of ForestFlow Generation with Single-Output Trees}
import numpy

y, n_t, n, p, regressors = inputs()
# Sample labels for conditioning using frequencies from the training dataset
y_uniq, y_counts = numpy.unique(y, return_counts=True)
y_probs = y_counts / numpy.sum(y_counts)
y_sample = numpy.argmax(numpy.random.multinomial(1, y_probs, size=n), axis=1)
label_y = y_uniq[y_sample]
# Sort by label and slice each class
label_y.sort()
y_uniq, y_counts = numpy.unique(label_y, return_counts=True)
y_slice = {}
csum = 0
for y_i, count in zip(y_uniq, y_counts):
  y_slice[y_i] = slice(csum, csum + count)
  csum += count
# Solve ODE with Euler's method starting from noise
X1 = numpy.random.normal(size=(n, p)).astype(numpy.float32)
h = 1 / (n_t-1) # size of timestep
results = []
for y_idx, y_i in enumerate(y_uniq):
  X1_i = X1[y_slice[y_i], :]
  for t_i in numpy.linspace(1, 0, num=n_t):
    model = regressors[t_i][y_idx]
    X1_i = X1_i - h * model.predict(X1_i)
  results.append(X1_i)
X0 = numpy.concatenate(results, axis=0)
\end{mintedbox}

\subsection{Analysis of Data Iterator}\label{app:data_iterator}

In Section \ref{sec:scaling} we remarked that we branched off of the Nov. 2, 2023 version of the ForestDiffusion codebase for our analysis.\footnote{See \href{https://github.com/SamsungSAILMontreal/ForestDiffusion/tree/855281b}{\tt{github.com/SamsungSAILMontreal/ForestDiffusion/tree/855281b}}.} The codebase is active, and has seen numerous changes since that date. Due to the nature of the work presented in this paper, we must select a fixed snapshot in order to provide meaningful analysis and reproducible results. There are a few reasons we have not branched off of more recent versions for the main discussion.

The changes immediately following our branching point (see commit hash \texttt{5417806} and following) implement sweeping changes that do not directly address the issues we have noted, but instead move in an orthogonal direction with the use of a data iterator. However, many of the older implementation issues remain, and there is still value in understanding and solving them from an engineering perspective. We will discuss the merits of the data iterator below, but first note that the paper \cite{jolicoeur2023generating} does not use it for their published results, commenting only in their conclusion section that a data iterator could be implemented in future work. Since only the published results are available to us for comparison, we must use the earlier version of the codebase prior to the data iterator's addition.

Second, more recent versions of the codebase (up to hash \texttt{818ac3b} as of writing) do not exhibit such extreme memory consumption issues when the data iterator is used, but this is not because the data iterator itself is a solution to the issues. The code implementing the data iterator inadvertently resolved most of what we described as Issue 2 in Section \ref{sec:analysis}. We say this was inadvertent because Issue 2 remains present in branches of the execution where the data iterator is not enabled. To disentangle the effects of the data iterator from Issue 2, we started from the earlier snapshot.

Third, we have identified unintended issues with the use of the data iterator that go beyond resource inefficiencies and actually lead to incorrectly trained models. As emphasized in Section \ref{sec:forest_recap}, one drawback of XGBoost is that it does not allow for mini-batch training, instead optimizing over the entire labeled training set at once. The data iterator is an optional technique with an evolving implemention since XGBoost 1.2.0 which allows for batched processing of large datasets. XGBoost's histogram training method does not use the dataset during optimization, instead it constructs a summarized version of the dataset prior to optimization. The summary, stored in a \texttt{QuantileDMatrix} object, consists of histogram quantiles of each feature, along with histogram bin indices of each datapoint. Then, during optimization only the stored quantiles are used for split values in trees, while the bin indices indicate whether a data point falls into the left or right child node given a quantile split. The data iterator allows the histogram quantiles and bin indices to be constructed iteratively from batches of data, rather than from the entire dataset at once which can avoid memory bottlenecks with large datasets. After construction of the \texttt{QuantileDMatrix}, XGBoost does not use the data iterator or raw dataset again during training.

Unfortunately, the use of the data iterator after commit hash \texttt{5417806} is flawed since it uses fresh noise to construct $X_t$ from $X_0$ each time it is consumed. XGBoost consumes the data iterator four times in total, all of which are for the creation of the \texttt{QuantileDMatrix}, when it computes the overall data shape\footnote{\href{https://github.com/dmlc/xgboost/blob/release_2.1.0/src/data/iterative_dmatrix.cc\#L172}{\tt{github.com/dmlc/xgboost/blob/release\_2.1.0/src/data/iterative\_dmatrix.cc\#L172}}}, 

constructs the histogram\footnote{\href{https://github.com/dmlc/xgboost/blob/release_2.1.0/src/data/iterative_dmatrix.cc\#L206}{\tt{github.com/dmlc/xgboost/blob/release\_2.1.0/src/data/iterative\_dmatrix.cc\#L206}}}, stores bin indices in row-major order\footnote{\href{https://github.com/dmlc/xgboost/blob/release_2.1.0/src/data/iterative_dmatrix.cc\#L242}{\tt{github.com/dmlc/xgboost/blob/release\_2.1.0/src/data/iterative\_dmatrix.cc\#L242}}}, and stores bin indices in column-major order.\footnote{
\href{https://github.com/dmlc/xgboost/blob/release_2.1.0/src/data/iterative_dmatrix.cc\#L267}{\tt{github.com/dmlc/xgboost/blob/release\_2.1.0/src/data/iterative\_dmatrix.cc\#L267}}} The latter three components must be built from the same datapoints provided in the same order for correct construction, but the injection of noise by the ForestDiffusion implementation means different datasets are used each time. Seeding the randomness of the data iterator when it is initialized and reset would ensure that the same noisy datapoints appear in the same order on each of the four passes.

There is however one benefit of the data iterator relevant to ForestDiffusion and ForestFlow that we have uncovered. When passing the entire input dataset to XGBoost for \texttt{QuantileDMatrix} construction, the input dataset is held in memory by the \texttt{QuantileDMatrix} throughout optimization, although it is not used. When using the data iterator, \texttt{QuantileDMatrix} does not hold a copy of the input throughout optimization. This could become relevant when training many XGBoost regressors in parallel; say we train $n_\text{jobs}$ jobs in parallel each using the appropriate subset of $X_\text{train}$, the subsets being $[n_i\cdot K, p]$ arrays, where $n_i$ is $n/n_y$ for balanced classes. Using the data iterator would avoid holding these $n_\text{jobs}$ copies of size $[n/n_y\cdot K, p]$ within the respective \texttt{QuantileDMatrix} objects. In the spirit of our paper, we can quantify the benefit it would give on the Pions dataset which is $n_\text{jobs}\cdot n/n_y\cdot K\cdot p\cdot 8$ bytes (or 4 bytes after the switch to \texttt{fp32} from Issue 7 in Appendix \ref{app:continued_issues}). With $n_\text{jobs}=40$, $K=100$, $n=120,800$, and $p=533$ we would see a further reduction of 128 GiB (or 64 GiB). While significant, this is lower by orders of magnitude than the benefits we described in Issues 1, 2, and 3. 

There is also a major con to using the data iterator - it is slower. First, for general uses, the data iterator constructs the \texttt{QuantileDMatrix} histogram on one batch, and then iteratively updates it on subsequent batches. This process requires more computation overall than constructing the histogram in a single shot on the entire dataset. Second, in the particular use of the data iterator for ForestDiffusion, (seeded) noise must be generated in each of the four passes through the iterator. This is wasteful compared to the alternative of generating noise once across the entire dataset.

Still, the data iterator can be helpful, so we have included a corrected implementation in our codebase. To demonstrate its merits, we reproduce the data from Figure \ref{fig:single_usage} in Table \ref{tab:data_iterator} with the data iterator option included, where $K$ batches are used such that only one copy of the raw dataset is loaded at a time during \texttt{QuantileDMatrix} construction. We see a marginal slowdown, but also reduction in peak memory usage for the reasons explained above. In particular, for the $n=100,000$ run we note that the expected memory savings of $n_\text{jobs}*(n/n_y*K*p*4)=1.5$ GiB is consistent with the observed memory savings. In conclusion, the data iterator can be beneficial for very large datasets if there is a memory bottleneck. Otherwise, it is better to use the ordinary \texttt{QuantileDMatrix} construction method as it is faster.

\begin{table}[]
    \caption{Data from Figure \ref{fig:single_usage} compared against the data iterator. \textbf{Left:} Training time (s). \textbf{Right:} Peak memory usage (GiB).}
    \label{tab:data_iterator}
    \centering
        \setlength{\tabcolsep}{1pt} 
    \begin{tabular}{l|rrrrr}
    \toprule
        Method / $n$ & 1000 & 3000 & 10,000 & 30,000 & 100,000\\
         \midrule
         Original & 297 & 363 & 587 & 1108 & \emph{crash}\\
         Ours & 172 & 221 & 380 & 880 & 2711\\
         Ours-Iterator & 177 & 230 & 408 & 957 & 2941\\
         \bottomrule
    \end{tabular} \ \
        \begin{tabular}{l|rrrrr}
    \toprule
    Method / $n$ & 1000 & 3000 & 10,000 & 30,000 & 100,000\\
             \midrule
Original & 33 & 36 & 71 & 169 & \emph{crash}\\
         Ours & 18 & 18 & 20 & 25 & 31\\
         Ours-Iterator & 19 & 20 & 20 & 23 & 29\\
\bottomrule
    \end{tabular}
\end{table}

\section{Performance Improvements and Ablation Studies}
\label{app:performance}

In this Appendix, we further details methods to improve the generative quality or resource utilization of ForestDiffusion and ForestFlow \cite{jolicoeur2023generating} that go beyond implementation changes, and evaluate them with ablation studies. This discussion extends Section \ref{sec:alg_improvements} from the main text.

\subsection{Multi-Output Trees}\label{app:multi-output}

One obvious downside of using XGBoost regressors is that they output a scalar, whereas for generative modelling we need to output a vector $\mathbf{x}$. In practice $\mathbf{x}$ is often high dimensional, and its dimension $p$ enters multiplicatively into the number of ensembles needed ($n_t\cdot n_y\cdot p$) for ForestDiffusion and ForestFlow.

We propose replacing single-output trees with multi-output trees, also referred to as vector-leaf trees \cite{zhang2021gbdtmo, ying2022mt, marz2022multi, iosipoi2022sketchboost, schmid2023tree}. Each leaf node in the multi-output tree provides a vector of values, and the training algorithm is modified to fit all output variables at once by minimizing the sum of losses over individual outputs \cite{zhang2021gbdtmo}. Not only does this reduce the number of ensembles we require by a factor of $p$, but it has the potential to increase model performance by better capturing correlations between output variables during generation since generated elements do not come from independent trees. Although single-output trees may be sampling from the marginals, and not the joint distribution at any timestep, we do note that feature interactions can occur \emph{between} timesteps in the ForestDiffusion and ForestFlow algorithms, since during generation (independent) outputs from one timestep affect all inputs in the next timestep of the SDE/ODE solve. This explains why the algorithms can show strong distribution learning abilities even with single-output trees that are limited to sampling marginally.

Past work on discriminative tasks has shown that multi-output trees may require thousands of boosting rounds before their performance surpasses their single-output counterparts \cite{zhang2021gbdtmo}. Similarly, we found for our generative tasks that multi-output trees only become comparable or surpass single-output baselines when scaled up with large $n_{\text{tree}}$ ($\sim2000$) and $K$ ($\sim1000$) with appropriate regularization, see Table \ref{tab:main}. However, we noticed that multi-output trees are less prone to overfitting and generalize better to the test set. This may be because multi-output trees solve a more difficult optimization task involving all outputs at once, meaning they are less prone to learning ``shortcuts'' that minimize the loss on individual outputs but that do not generalize. 

These differences are demonstrated by comparing Table \ref{tab:main}, which used scaled-up ensembles and early stopping, to Table \ref{tab:main_ablation} below without those additions. Without the benefit of wide ensembles the MO method seems to underperform SO, however, we note that both versions use the same $n_{\textrm{tree}}$ and maximum depth hyperparameters across tests, meaning that the SO models essentially use $p$ times more parameters. When scaled up with early-stopping, single-output trees tended to stop earlier than multi-output versions, which may simply be a result of their different capacities. 

Since version 2.0.0, XGBoost has implemented multi-output trees. Although there is a lack of documentation currently, to the best of our knowledge the algorithm reflects the proposal of \citet{zhang2021gbdtmo}. During our testing we identified a bug in the gain computation in the official XGBoost codebase and reported it to the maintainers who implemented our proposed fix. See \href{https://github.com/dmlc/xgboost/issues/9960}{\tt{github.com/dmlc/xgboost/issues/9960}}. Hence, only XGBoost version 2.1.0 or later should be used for multi-output trees. Still, this version's implementation is not yet optimized for time and memory performance, so our measurements in Section \ref{sec:resource-scaling} should be considered preliminary.
\clearpage
\begin{figure}[h!]
    \centering
    \includegraphics[width=1.0\linewidth]{figures/n_tree/SO_ES_flow.eps}
    \includegraphics[width=1.0\linewidth]{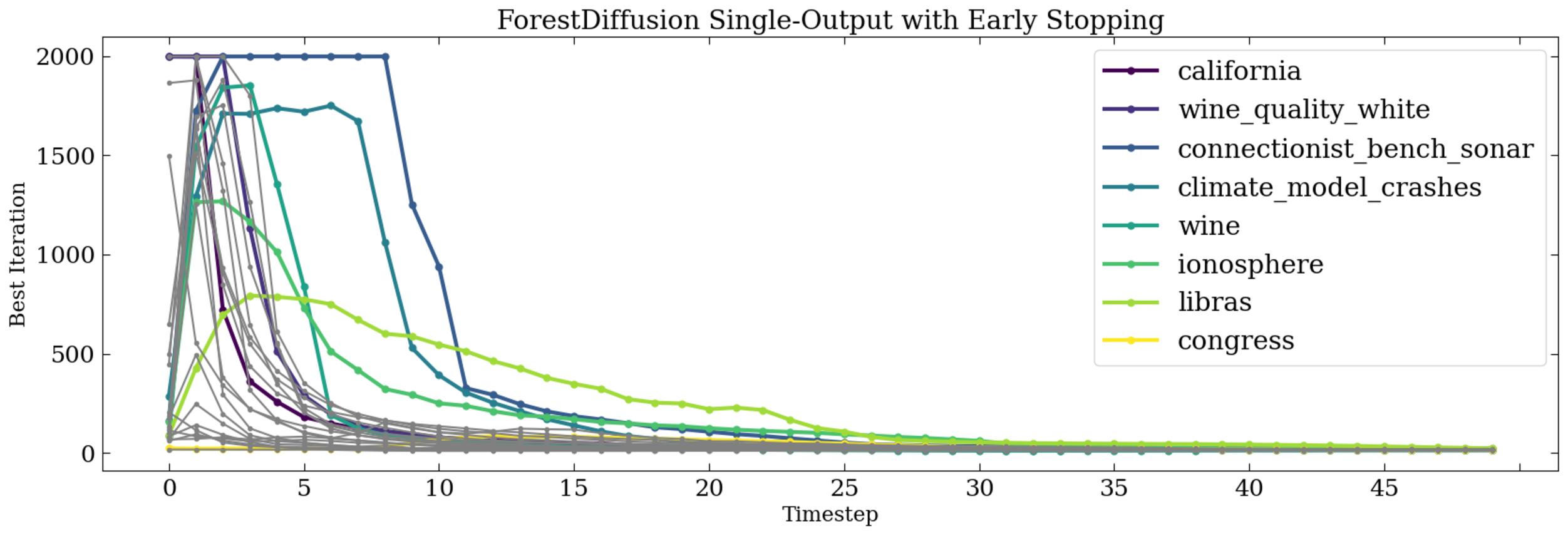}
    \includegraphics[width=1.0\linewidth]{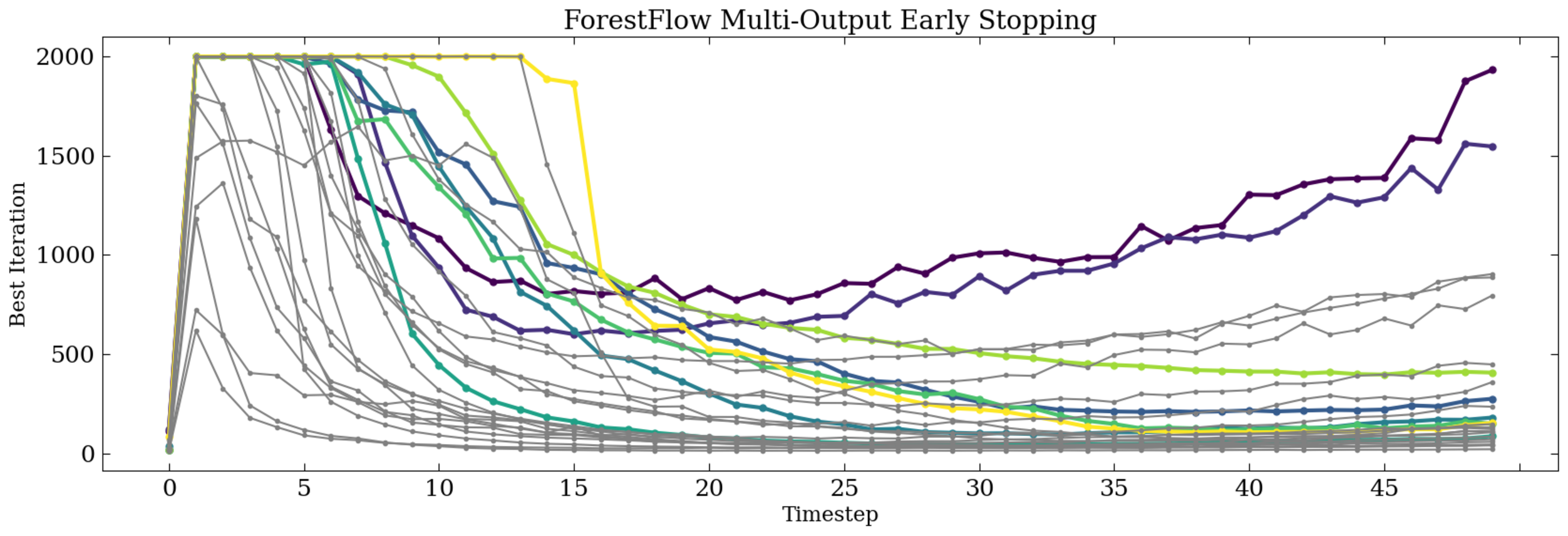}
    \includegraphics[width=1.0\linewidth]{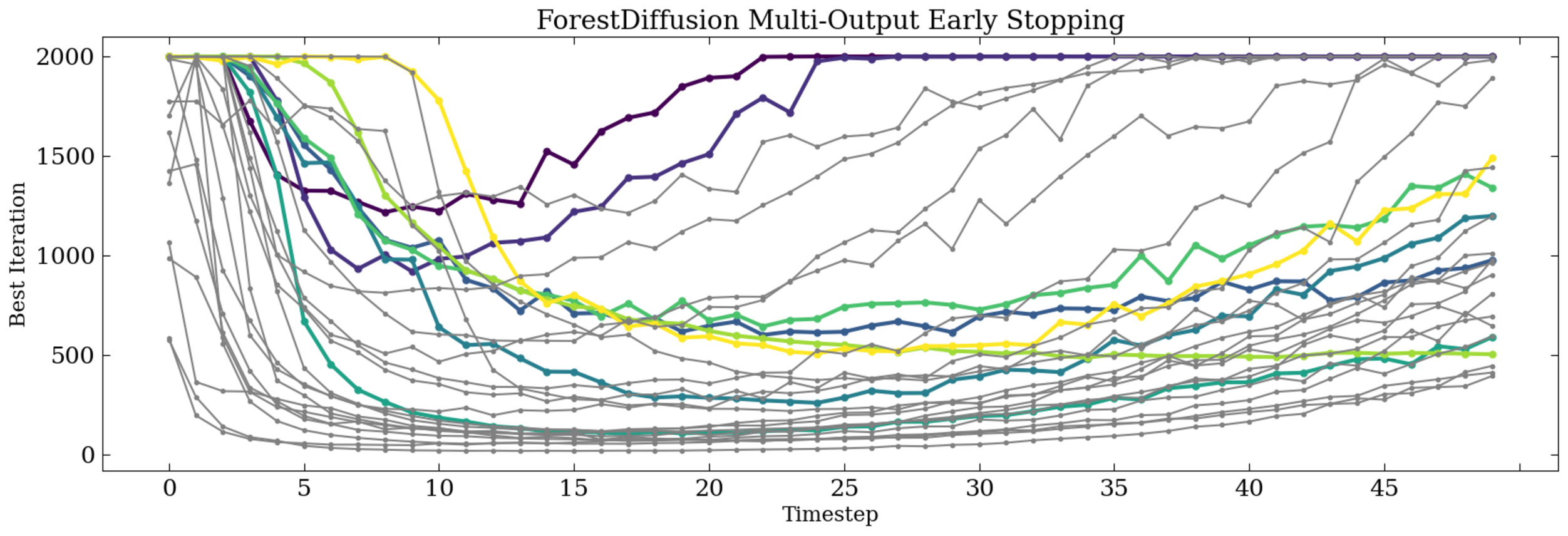}
    \caption{Number of trees at the best iteration of the validation loss by timestep and dataset. The number of trees is averaged over ensembles at the same timestep for different classes $y$ and features $p_i$ (for single-output trees). Selected datasets from the set of all 27 in Table \ref{tab:datasets} are highlighted for comparison between methods. Models are trained with early stopping after $n_{\text{ES}}=20$ rounds. The top figure is reproduced from Figure \ref{fig:n_tree_SO_flow} in the main text for ease of comparison.}
    \label{fig:n_tree}
\end{figure}
\clearpage

\subsection{Early Stopping with Scaled-Up Hyperparameters}\label{app:hypers}

\citet{jolicoeur2023generating} observed that their models were underfitting the training data, and hence used no regularization on their XGBoost ensembles. This has the downside that nearly every tree in every ensemble will reach is maximal size, determined by the depth hyperparameter, resulting in very large models that slow down training and generation. To provide more capacity for models, we scaled up $n_\text{tree}$ from $100$ to $2000$, and increased $K=100$ to $1000$ for better coverage of the loss function expectations. While scaling up the capacity of models can reduce underfitting and increase model performance, it also exacerbates training and generation speed concerns.

We have found that the severity of underfitting varies greatly across timesteps. Models at timesteps closer to $t=1$ (noise) converge quickly, so increasing $n_\text{tree}$ without regularization is computationally wasteful, and in the worst case could lead to severe overfitting. Hence, to provide capacity only where it is needed we propose to apply the well-known regularization technique of early stopping individually on each XGBoost ensemble's training loss. Taking advantage of the nature of ForestDiffusion and ForestFlow, we reused the training set $X_0$ with fresh noise $X_1$ to create the regression features and targets that we validate on. 

In Figure \ref{fig:n_tree_SO_flow} we showed how ensembles trained for $t$ close to 1 stop improving on the validation set very early. We reproduce this figure (which used FF-SO-Scaled) along with the other scaled variants in Figure \ref{fig:n_tree} for comparison. The patterns of early stopping are very similar between ForestDiffusion and ForestFlow, but differ greatly between the single-output and multi-output variants. First, a different subset of datasets ends up using very wide ensembles -- as one example SO stops early at all timesteps on \emph{congress}, whereas MO uses maximally wide ensembles for many timesteps. Second, on many datasets MO also trains wide ensembles for later timesteps. One basic implication is that early stopping provides a greater computational benefit for SO variants as the average stopping round is lower. We suppose this is due to how multi-output trees are better able to capture joint distributions and can pick up on the subtle correlations between elements which are mostly noise. However, as timesteps close to $t=1$ in diffusion sampling are mainly responsible for global structure and not fine details, we find that single-output trees are not at a significant disadvantage here.

Finally, the fact that SO focuses almost entirely on the first fifth of the timesteps indicates that a non-uniform partitioning of the $(0, 1)$ interval could  potentially improve performance. This would be similar to how NN-based diffusion models are often trained with a non-uniform noise schedule \cite{pmlr-v139-nichol21a, chen2023importance}. We leave these explorations for future work.

\begin{figure}[t]
    \centering
    \includegraphics[width=0.9\linewidth]{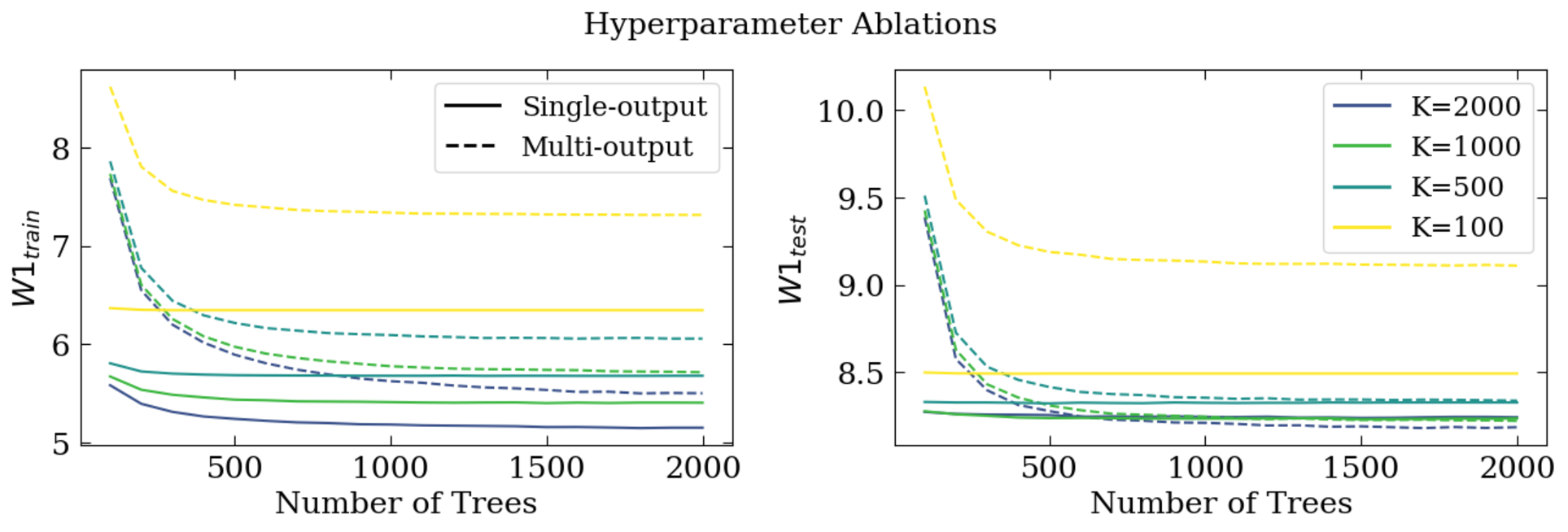}
    \includegraphics[width=0.9\linewidth, trim={0 0 0 30}, clip]{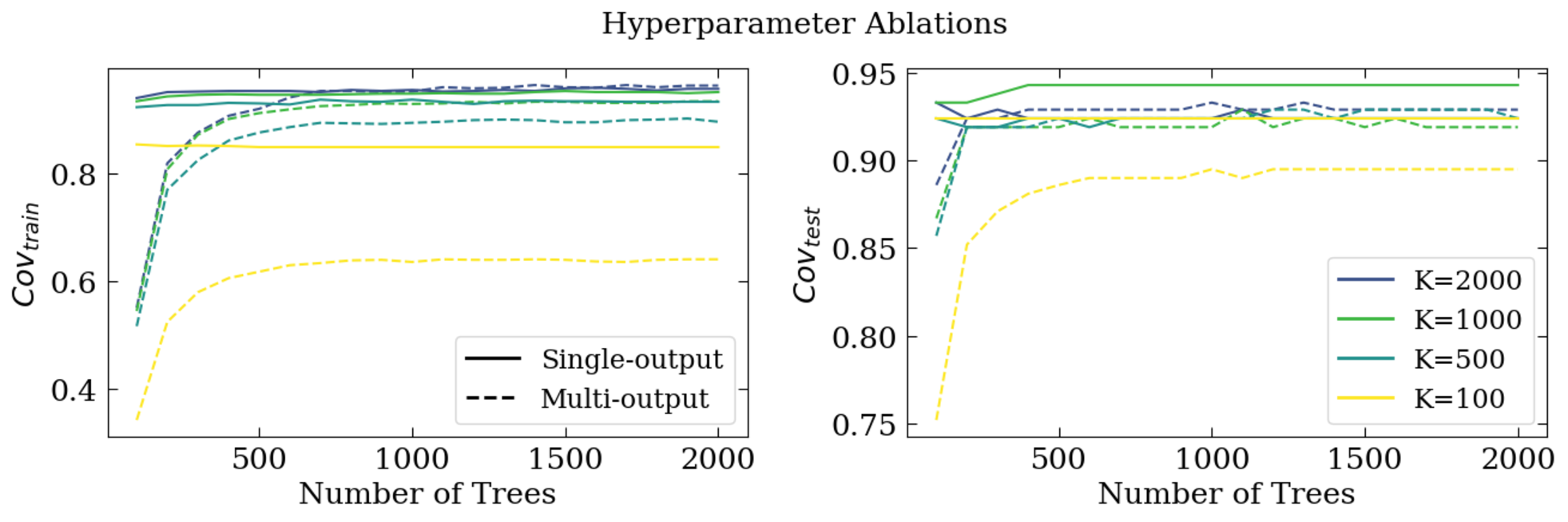}
    \caption{Effect of varying $K$, $n_\text{tree}$, and the tree structure (SO vs. MO) on distributional metrics.}
    \label{fig:hypers}
\end{figure}

To wrap us this discussion, we show hyperparameter ablations over a single dataset, \emph{connectionist\_bench\_sonar}, for $K$, $n_\text{tree}$, and tree structure in Figure \ref{fig:hypers}. The results demonstrate that SO hardly benefits from wide ensembles, while MO can continue to marginally improve its generalization to the test set even up to $n_\text{tree}=2000$. The duplication factor $K$ has a strong effect, and the default setting of $100$ used in \cite{jolicoeur2023generating} is far from enough. While SO outperforms MO for most settings, there is a regime where MO can achieve the best generalization on $W1_\text{test}$, but this requires both high duplication ($K\geq 1000$), and wide ensembles ($n_\text{tree}\geq 1000$). 

\subsection{Class-Conditional Scalers}\label{app:scalers}

One advantage of XGBoost as a function approximator is its robustness to data with varying scales and distributions. This stands in stark contrast to deep NNs which require data to be carefully pre-processed for best results. While XGBoost itself is agnostic, ForestDiffusion and ForestFlow do require input data to be on the same scale as the added noise in Eq. \ref{eq:diffusion_conditional} and \ref{eq:cfm_prob}. \citet{jolicoeur2023generating} achieve this by applying min-max scaling over the entire input dataset. However, when using the class-conditional variant, models are trained on disjoint sets of data belonging to each class. If the classes have distinct distributions, which is often the point of distinguishing data by class, then the data subsets actually provided to the training algorithm may not be properly scaled. To rectify this, we propose class-conditional min-max scaling. This is especially beneficial on the calorimeter datasets as the classes represent particle energies increasing on an exponential scale leading to vastly different per-class distributions. Class-conditional scaling centers the data better making it more easily distinguishable as noise is added, ultimately benefiting the model performance.

\subsection{Sampling with the Training Set Label Distribution}\label{app:label_sampling}

For class-conditional sampling, \cite{jolicoeur2023generating} used the relative prevalence of classes in the training set to define a multinomial distribution and sampled from it to create class labels for conditioning. We found it advantageous to directly use the empirical distribution of class labels from the training set for conditioning, especially for the distributional Wasserstein metrics. For the small datasets used in benchmarking (Table \ref{tab:datasets}), multinomial sampling may lead to a skewed distribution by chance; the law of large numbers may not kick in at these sizes. This type of sampling with training set labels is also mandated in the Fast Calorimeter Simulation Challenge \cite{calochallenge}.

As an ablation, in Table \ref{tab:main_ablation} we show performance results corresponding to Table \ref{tab:main} where SO and MO use the same hyperparameter settings as Original, but with class-conditional scalers and training set label conditioning (i.e. not using scaled up models nor early stopping). The results show that some gains can be made on distributional metrics simply by improving the distributions of scaled data and class labels. However, scaling up the models as done in Table \ref{tab:main} provides more impressive gains, especially for the MO variant.
\begin{table}[t]
\caption{Average rank (standard error) of generated data quality over 27 datasets. Lower is better.}  
\label{tab:main_ablation}
\small
\centering
\resizebox{\textwidth}{!}{
    \setlength{\tabcolsep}{3pt} 
\begin{tabular}{lrrrrrrrr|r}
  \toprule
 & $W1_\textrm{train}$ & $W1_\textrm{test}$ & $\textrm{Cov}_\textrm{train}$ & $\textrm{Cov}_\textrm{test}$ & $R^2_\textrm{gen}$ & $F1_\textrm{gen}$ & $P_\textrm{bias}$ & $\textrm{cov}_\textrm{rate}$ & Avg.\\ \hline
  GaussianCopula \cite{joe2014dependence} & 10.0{\tiny$\pm$0.3} & 10.1{\tiny$\pm$0.3} & 10.0{\tiny$\pm$0.4} & 10.1{\tiny$\pm$0.4} & 9.1{\tiny$\pm$0.1} & 9.4{\tiny$\pm$0.4}& 8.4{\tiny$\pm$1.5}& 10.6{\tiny$\pm$0.5} & 9.7{\tiny$\pm$0.1} \\  
  TVAE \cite{xu2019tvae} & 7.9{\tiny$\pm$0.4} & 7.6{\tiny$\pm$0.4} & 8.3{\tiny$\pm$0.4} &  8.3{\tiny$\pm$0.4} & 9.6{\tiny$\pm$0.7} & 9.0{\tiny$\pm$0.6}& 10.7{\tiny$\pm$0.5} & 10.1{\tiny$\pm$0.3} & 8.9{\tiny$\pm$0.0} \\  
  CTGAN \cite{xu2019tvae} & 11.4{\tiny$\pm$0.2} & 11.3{\tiny$\pm$0.2} & 11.2{\tiny$\pm$0.2} & 11.1{\tiny$\pm$0.2} & 11.6{\tiny$\pm$0.2} & 11.4{\tiny$\pm$0.2}& 7.9{\tiny$\pm$1.3}& 10.6{\tiny$\pm$0.5} & 10.8{\tiny$\pm$0.1} \\  
  CTAB-GAN+ \cite{zhao2024ctabganp} & 9.7{\tiny$\pm$0.3} & 9.6{\tiny$\pm$0.4} & 9.9{\tiny$\pm$0.4} & 9.8{\tiny$\pm$0.4} & 10.0{\tiny$\pm$0.2} & 9.9{\tiny$\pm$0.4}& 10.7{\tiny$\pm$0.6}& 8.9{\tiny$\pm$1.2} & 9.8{\tiny$\pm$0.1} \\  
  STaSy \cite{kim2023stasy} & 9.0{\tiny$\pm$0.2} & 9.2{\tiny$\pm$0.2} & 8.0{\tiny$\pm$0.3} & 8.0{\tiny$\pm$0.4} & 8.3{\tiny$\pm$1.3} & 7.8{\tiny$\pm$0.5}& 6.9{\tiny$\pm$1.1}& 6.5{\tiny$\pm$1.4} & 8.0{\tiny$\pm$0.2} \\  
  TabDDPM \cite{kotelnikov2023tabddpm} & 4.4{\tiny$\pm$0.9} & 5.9{\tiny$\pm$0.8} & 4.1{\tiny$\pm$0.7} & 4.9{\tiny$\pm$0.7} & \better{2.0{\tiny$\pm$0.7}}  & 5.6{\tiny$\pm$0.8}& \better{3.3{\tiny$\pm$1.4}}& \better{3.2{\tiny$\pm$0.7}} & 4.2{\tiny$\pm$0.1} \\  
  \hline
  ForestDiffusion-Original & 5.0{\tiny$\pm$0.3} & 4.8{\tiny$\pm$0.2} & 5.0{\tiny$\pm$0.3} & 4.5{\tiny$\pm$0.4} & 4.0{\tiny$\pm$0.5} & \better{3.2{\tiny$\pm$0.5}}& 6.0{\tiny$\pm$0.7}& 5.1{\tiny$\pm$0.9} & 4.7{\tiny$\pm$0.1} \\  
  ForestDiffusion-SO & 4.4{\tiny$\pm$0.2} & 4.3{\tiny$\pm$0.2} & 5.4{\tiny$\pm$0.2} & 5.1{\tiny$\pm$0.4} & 6.9{\tiny$\pm$0.4} & 4.5{\tiny$\pm$0.5}& 6.0{\tiny$\pm$1.0}& 6.1{\tiny$\pm$0.7} & 5.3{\tiny$\pm$0.1} \\  
  ForestDiffusion-MO & 6.6{\tiny$\pm$0.3} & 6.5{\tiny$\pm$0.4} & 6.9{\tiny$\pm$0.3} & 6.3{\tiny$\pm$0.4} & 6.0{\tiny$\pm$0.6} & 5.0{\tiny$\pm$0.5} & 6.0{\tiny$\pm$0.5} & 5.4 {\tiny$\pm$0.6} & 6.1{\tiny$\pm$0.0} \\  
  ForestFlow-Original & 3.0{\tiny$\pm$0.3} & 2.9{\tiny$\pm$0.3} & 2.4{\tiny$\pm$0.3} & 2.9{\tiny$\pm$0.4} & 2.6{\tiny$\pm$0.6} & 4.0{\tiny$\pm$0.4}& \better{3.3{\tiny$\pm$1.0}} & 3.4{\tiny$\pm$0.8} & \better{3.1{\tiny$\pm$0.1}} \\  
  ForestFlow-SO & \better{2.0{\tiny$\pm$0.3}} & \better{1.8{\tiny$\pm$0.4}} & \better{2.2{\tiny$\pm$0.4}} & \better{2.7{\tiny$\pm$0.5}} & 2.9{\tiny$\pm$0.5} & 4.0{\tiny$\pm$0.4}& 5.3{\tiny$\pm$0.8}& 4.3{\tiny$\pm$0.9} & \better{3.1{\tiny$\pm$0.1}} \\  
  ForestFlow-MO & 4.6{\tiny$\pm$0.3} & 4.1{\tiny$\pm$0.3} & 4.6{\tiny$\pm$0.4} & 4.4{\tiny$\pm$0.4} & 5.1{\tiny$\pm$0.7} & 4.5{\tiny$\pm$0.5}& 3.6{\tiny$\pm$0.6}& 3.9{\tiny$\pm$1.3} & 4.3{\tiny$\pm$0.1} \\  
   \hline
\bottomrule 
\end{tabular}
}
\vspace{-15pt}
\end{table}

\section{Experimental Details}
\label{app:experiments}

In this Appendix we lay out the details of the experiments conducted in Section \ref{sec:results} and Appendix \ref{app:performance}.

\subsection{Datasets}\label{app:datasets}
For the resource scaling experiments in Section \ref{sec:resource-scaling} we used synthetic data that was randomly generated. The input data $X$ of size [$n$, $p$] was simply drawn from a identity covariance Gaussian, while the class label $y$ was randomly drawn from the integers $[0, n_y)$. While this data is meaningless for model performance, it gives us precise control over the dataset size for analysing resource usage. Since the correlations between features are random, unregularized XGBoost regressors will use essentially their entire available capacity in learning which gives us a good upper bound on resource usage. The dataset size parameters were set at $n=1000$, $p=10$, and $n_y=10$ by default, and a single one of these three was modified at a time. We swept over the values $n\in\{100, 300, 1000, 3000, 10000, 30000, 100000, 300000\}$, $p\in\{3, 10, 30, 100, 300\}$, and $n_y\in\{1, 3, 10, 30, 100\}$.

The early stopping regularization used for SO-ES and MO-ES introduces dataset dependence for training time. One may wonder if the randomness of the synthetic data affects when early stopping would typically occur. To check this, we repeated the resource scaling experiments on differently sized subsamples of the Pions data set. We found the same trends for SO-ES and MO-ES as shown in Figure \ref{fig:resources-random} which gives us confidence that our reported results for training time will be typical of real-world datasets.

In Section \ref{sec:benchmarking} we used 27 datasets from the UCI Machine Learning Repository of tabular datasets \cite{uci} and from scikit-learn \cite{scikit-learn} that have previously been studied \cite{muzellec2020missing, jolicoeur2023generating}. These datasets are summarized in Table \ref{tab:datasets}, and showcase a variety of tabular learning tasks with variation in the number of datapoints $n$, features $p$, classes $n_y$, and target types. In each case, we randomly held-out 20\% of the dataset as a test split and trained generative models on the remaining 80\%. Categorical variables are one-hot encoded.

Each of the UCI datasets is covered by a CC BY 4.0 license, while the \emph{iris} dataset has a BSD 3-Clause License,
and \emph{california housing} has no license.

\begin{table}[htp]
\caption{Tabular benchmark datasets. Training dataset sizes $n$ are 80\% of the total number of datapoints. Continuous and integer targets $y$ are treated as an additional feature.}
\label{tab:datasets}
\centering
\setlength{\tabcolsep}{3pt} 
\begin{tabular}{lcrrrl}
\toprule
 Dataset & Citation & \# Datapoints $n$ &   \# Features $p$ & \# Classes $n_y$ & Target $y$ type \\
\midrule
        airfoil self noise & \citep{misc_airfoil_selfnoise_291} &   1503 &   6  & N/A & Continuous \\
        bean & \citep{koklu2020multiclass} & 13611 & 16 & 7 & Categorical \\
          blood transfusion & \citep{misc_blood_transfusion_service_center_176} &    748 &   4 & 2& Binary\\
  breast cancer diagnostic & \citep{misc_breast_cancer_wisconsin_diagnostic_17} &    569 &  30 & 2 & Binary \\
                  california housing & \citep{pace1997sparse} &  20640 &   9 & N/A &  Continuous \\
   car evaluation & \citep{misc_car_evaluation_19} &   1728 &   6 & 4 & Categorical \\
     climate model crashes & \citep{misc_climate_model_simulation_crashes_252} &    540 &  18 & 2 & Binary \\
       concrete compression & \citep{misc_concrete_compressive_strength_165} &   1030 &   9  & N/A &Continuous \\
             concrete slump & \citep{misc_concrete_slump_test_182} &    103 &   8  & N/A & Continuous \\
   congressional voting & \citep{misc_congressional_voting_records_105} &   435 &   16 & 2 & Binary \\
 connectionist bench sonar & \citep{misc_connectionist_bench_sonar_mines_vs_rocks_151} &    208 &  60 & 2 & Binary \\
 connectionist bench vowel & \citep{misc_connectionist_bench_vowel_recognition__deterding_data_152} &    990 &  10 & 2 & Binary \\
                       ecoli & \citep{misc_ecoli_39} &    336 &   7  & 8 & Categorical \\
                       glass & \citep{misc_glass_identification_42} &    214 &   9 & 6 & Categorical \\
                  ionosphere & \citep{misc_ionosphere_52} &    351 &  33 & 2 & Binary \\
                        iris & \citep{misc_iris_53} &    150 &   4 & 3 & Categorical \\
                      libras & \citep{misc_libras_movement_181} &    360 &  90  & 15 & Categorical \\
                  parkinsons & \citep{misc_parkinsons_174} &    195 &  22 & 2  & Binary \\
             planning relax & \citep{misc_planning_relax_230} &    182 &  12  & 2 & Binary \\
        qsar biodegradation & \citep{misc_qsar_biodegradation_254} &   1055 &  41 & 2 & Binary \\
                       seeds & \citep{misc_seeds_236} &    210 &   7   & 3 & Categorical \\
           tic-tac-toe & \citep{misc_tic-tac-toe_endgame_101} &   958 &   9 & 2 & Binary \\
                        wine & \citep{misc_wine_109} &    178 &  13 & 3    & Categorical \\
          wine quality red & \citep{misc_wine_quality_186} &   1599 &  11  & N/A  & Integer \\
        wine quality white & \citep{misc_wine_quality_186} &   4898 &  12  & N/A & Integer \\
        yacht hydrodynamics & \citep{misc_yacht_hydrodynamics_243} &    308 &   7  & N/A & Continuous \\
                       yeast & \citep{misc_yeast_110} &   1484 &   8 & 10 & Categorical \\
\bottomrule
\end{tabular}
\end{table}

\subsection{Metrics}\label{app:metrics}

For a fair and direct comparison in Section \ref{sec:benchmarking}, we use eight performance metrics for generated data from \cite{jolicoeur2023generating} which measure quality along four different axes: distributional distance, diversity, usefulness for training discriminative models, and usefulness for statistical inference.

\textbf{Distributional Distance} \ We measure the Wasserstein-1 distance between the generated data and either the training set ($W1_{\textrm{train}}$) or test set ($W1_{\textrm{test}}$). The Wasserstein distance quantifies similarity in distribution - smaller distance to the test set is always desirable while distance to the training set should be similar in magnitude to the distance between the training and test sets, as a much smaller distance here can indicate memorization. Generally, $W1_{\textrm{train}}$ values are not less than the train-test distance, so we treat lower values as better. Computation of Wasserstein distances was done with the Python Optimal Transport library \cite{flamary2021pot}. These metrics are omitted for the larger \emph{bean} and \emph{california} datasets as they scale quadratically in dataset size which is prohibitively expensive \cite{muzellec2020missing, jolicoeur2023generating}.

\textbf{Diversity} \ Coverage \cite{naeem2020prdc} measures to what extent the generated data covers a reference dataset, where a reference datapoint is covered if there is at least one generated datapoint in its neighbourhood. Hence, generated data must be as diverse as the reference data to achieve high Coverage. Coverage is computed as the ratio of covered points to all  points \cite{stein2023exposing}
\begin{equation}
    \mathrm{Coverage}\left(\{x_i^g\}_{i=1}^n, \{x_j^r\}_{j=1}^m\right) = \dfrac{1}{m} \displaystyle \sum_{j=1}^m \max_{i=1,\dots,n} \mathds{1}\left(x_i^g \in B(x_j^r, \mathrm{NND}_k(x_j^r))\right),
\end{equation}
where $x^g$ are the generated datapoints, $x^r$ are the reference datapoints, $\mathds{1}(\cdot)$ denotes the indicator function, $B(x, r)$ denotes a ball centered at $x$ with radius $r$, and $\mathrm{NND}_k(x_j^r)$ is the nearest-neighbour distance between $x_j^r$ and its $k^{th}$ nearest neighbour in $\{x_j^r\}_{j=1}^m$. We use an L1 ball to compute distances as it is more suited for mixed data types typical of tabular data. $k$ is chosen automatically as the smallest value such that the training data has at least $95\%$ Coverage of the test data. We calculate the Coverage using either the training ($\textrm{Cov}_\textrm{train}$) or test ($\textrm{Cov}_\textrm{test}$) dataset as the reference. $\textrm{Cov}_\textrm{train}$ helps to address ``mode dropping'' where some parts of the training dataset are ignored, while $\textrm{Cov}_\textrm{test}$ helps measure the ability to generalize with sufficient diversity. These metrics were computed for all datasets.

\textbf{Usefulness for Training Discriminative Models} \ Tabular generative models are often motivated as a way to replace or extend training data for downstream tabular discriminative models \cite{xu2019tvae, kotelnikov2023tabddpm}. Available training data may be considered private and not suitable for directly training a discriminative model, whereas synthetic data derived from a generative model may be more palatable. Alternatively, synthetic data may be used with the hope that it leads to better performing downstream models. Hence, we measure the usefulness of generative models by training downstream discriminative models on generated data, and evaluating discriminative performance on the test set. Performance is measured either by the F1-score for classification tasks (20 datasets), or the $R^2$-coefficient for regression tasks (7 datasets), where higher is better. Since these metrics are highly dependent on the type of discriminative model used, we average the performance metrics over four different methods that are commonly used for tabular discriminative modelling: linear/logistic regression, AdaBoost \cite{freund1995adaboost}, Random Forests \cite{ho1995randomforest}, and, of course, XGBoost \cite{chen2016xgb}.

\textbf{Usefulness for Statistical Inference} \ Whereas the above metrics take a machine learning point of view in aiming to optimize the performance of a model, we can also consider a statistical point of view and measure the usefulness of synthetic data for inferring the importance of features \cite{vanbuuren2018flexible}. By training a linear model on either the training data or generated data we can compare the regression parameters $\beta$. If the generated data accurately represents the training data, the learned regression coefficients should be similar. If these coefficients are not similar, one might conclude from the generated data that a given feature is statistically important when the same conclusion would not be reached using the training data. The percent bias measures this difference and is defined as $P_\textrm{bias} = \vert \mathbb{E}\frac{\hat \beta - \beta}{\beta} \vert$ using the estimated coefficients $\hat \beta$ on generated data and actual coefficients $\beta$ from training data, with the expectation taken over generated data. From another direction, it is desirable for confidence intervals on the estimated coefficients $\hat \beta$ to contain the true coefficients $\beta$. This is quantified by the coverage rate $\textrm{cov}_\textrm{rate}$, the fraction of $\beta$ that are contained in the confidence intervals around $\hat \beta$. These metrics were computed only for the regression tasks (7 datasets). Lower is better for $P_\textrm{bias}$, but higher is better for $\textrm{cov}_\textrm{rate}$. Coverage rate is not to be confused with Coverage used above as a diversity metric.

\subsection{Baseline Methods}\label{app:methods}

In addition to comparing our approach to the original implementation of ForestDiffusion and ForestFlow in Section \ref{sec:benchmarking}, we also compare our improved models to 6 popular baseline methods for tabular generative modelling, including state-of-the-art deep learning methods, as done in \cite{jolicoeur2023generating}.

\textbf{GaussianCopula} \ Many deep learning-based generative models learn a mapping between a simple distribution on latent space and the data distribution \cite{loaiza2024deep}. Generation is done by sampling from the latent distribution and mapping the sample to data space. This overarching idea hearkens back to copula methods \cite{sklar1959fonctions} which model any multivariate joint distribution by its univariate marginals along with a copula describing the dependence structure. We use Gaussian copulas \cite{joe2014dependence} implemented by Synthetic Data Vault (SDV) \cite{patki2016sdv} using default hyperparameters.

\textbf{TVAE} \ Variational autoencoders (VAE) \cite{kingma2013auto} learn an encoder and decoder with a low dimensional latent space through variational inference. As a typical example we use the tabular VAE (TVAE) from \cite{xu2019tvae}, again implemented by SDV using default hyperparameters.

\textbf{CTGAN and CTAB-GAN+} \ Generative adversarial networks \cite{goodfellow2014generative} train a generator which produces synthetic datapoints, and a discriminator that tries to classify real and synthetic datapoints. The two networks are trained simultaneously in an adversarial manner. As a typical example we use the conditional tabular GAN (CTGAN) from \cite{xu2019tvae}, also as implemented by SDV. We also employ a more modern tabular GAN called CTAB-GAN+ \cite{zhao2024ctabganp} as implemented by its authors. Both methods use default hyperparameters from their respective implementations.

\textbf{StaSy and Tab-DDPM} \ More recently score-based \cite{song2019} and diffusion models \cite{ho2020denoising, song2021scorebased} have eclipsed VAEs and GANs for generative quality on the image modality. To represent these classes, we use STaSy \cite{kim2023stasy}, a score-based method, and Tab-DDPM \cite{kotelnikov2023tabddpm} a denoising diffusion model \cite{ho2020denoising} adapted for tabular settings. For the former we use hyperparameters found by \cite{jolicoeur2023generating}, and the latter uses default hyperparameters from the author's implementation.

\subsection{Experimental Setup and Hyperparameters}\label{app:exp_setup}

When measuring resource usage in Section \ref{sec:resource-scaling} training time was clocked starting once data had been loaded and pre-processed, and stopped once all models had been trained (i.e. generation and evaluation are not included). For models that trained successfully (i.e. did not fail due to memory issues), we measure the time to generate five batches of data equal in size to the dataset a given model was trained on. We also monitor the used CPU memory every second (or every 10 seconds for long runs taking more than one hour), including at the beginning before any data is loaded to capture the background memory usage by the operating system. The peak memory that we report is the single highest measurement over training, with the single lowest measurement subtracted off to remove the effects of the background. The peak memory burden is more important than alternatives, like the average memory, since the peak burden determines if a job can successfully complete on a given machine.

Throughout our experiments we made efforts to keep comparisons as fair as possible by using the same hyperparameters as \cite{jolicoeur2023generating} where sensible. A summary of our settings is given in Table \ref{tab:hypers}. In particular, for the resource scaling experiments in Section \ref{sec:resource-scaling}, all methods use the same learning and XGBoost hyperparameters, other than the number of early stopping rounds $n_{\text{ES}}$ which is set to $20$ when enabled. The data was duplicated $K=100$ times with $n_t=50$ discrete time steps as recommended by \cite{jolicoeur2023generating}, and models were trained conditionally on $y$ whenever $n_y>1$. Computationally, 40 parallel training jobs were used (equal to the number of CPUs on our machine), with one CPU assigned to each job. For the sake of benchmarking, we did not reduce the number of parallel jobs when methods began failing due to memory issues. XGBoost hyperparameters were left at their defaults, other than L2 regularization which was set to $\lambda=0$. Notably, this means $n_{\textrm{tree}}=100$ trees were trained per ensemble of max depth 7. 

The same hyperparameters were used for our performance benchmarking in Section \ref{sec:benchmarking} and additional tests in Appendix \ref{app:performance}, with the exception of $K=1000$ and $n_\text{tree}=2000$ for the scaled up models. Each method was trained with 3 different random seeds on each dataset, and for each training run 5 sets of data the size of the training dataset were generated. Each set of data was used to compute the performance metrics independently, and the results were averaged across the 5 generations per 3 seeds. These averaged performance metrics were then used to compute the relative rankings between methods. For each metric, we computed the ranking of methods on each dataset and then took the mean and standard deviation of rankings across datasets. As discussed in Appendix \ref{app:metrics}, not all metrics could be used for all datasets, so the averages over datasets only include applicable datasets where the metric could actually be computed.

\begin{table}[]
    \centering
        \setlength{\tabcolsep}{3pt} 
    \caption{Hyperparameter settings}
    \label{tab:hypers}
    \small
    \begin{tabular}{lrrrrrrrrr}
      \toprule
        Method & $n_t$ & $K$ & $n_\text{tree}$ & $n_\text{ES}$ & $\eta$ &$\lambda$& $\epsilon$ & Scaler & Sampler\\
        \hline
         Original & 50 & 100 & 100 & 0 & 0.3 & 0 & 0.001 (FD), 0 (FF) & Single & Multinomial\\
         SO & 50 & 100 & 100 & 0 & 0.3 & 0 & 0.001 (FD), 0 (FF)& Per-class & Label\\
         MO & 50 & 100 & 100 & 0 & 0.3 & 0 & 0.001 (FD), 0 (FF)& Per-class & Label \\
         SO-ES (Figure \ref{fig:resources-random}) & 50 & 100 & 100 & 20 & 0.3 & 0 & 0 (FF) & Per-class & Label\\
         MO-ES (Figure \ref{fig:resources-random}) & 50 & 100 & 100 & 20 & 0.3 & 0 & 0 (FF) & Per-class & Label\\
         SO-Scaled (Table \ref{tab:main})& 50 & 1000 & 2000 & 20 & 0.3 & 0 & 0.001 (FD), 0 (FF) & Per-class & Label\\
         MO-Scaled (Table \ref{tab:main})& 50 & 1000 & 2000 & 20 & 0.3 & 0 & 0.001 (FD), 0 (FF) & Per-class & Label\\
         CaloForest & 100 & 20 & 20 & 0 & 1.5 & 1 & 0 (FF) & Per-class & Label \\
           \bottomrule
    \end{tabular}
\end{table}

\subsection{Additional Performance Benchmarking Results}\label{app:extra_benchmarking_plots}

Here we complement the summarized results of Table \ref{tab:main} by plotting the raw metric values averaged over three seeds for each metric, method, and dataset. We remark that the \emph{bean} and \emph{california} datasets were not evaluated with the Wasserstein metrics due to their size. Other plots show missing information for datasets when the metric is suited for either classifcation or regression tasks, but not both. In addition to the 6 baseline methods and variants of FD and FF, we also show the Oracle metric values which are obtained by using the training set as if it were generated data. Since the training set is truly from the underlying distribution, it can serve as a ``gold standard'' for some metrics. Notably, $W1_\text{train}$ will always be zero while $\text{Cov}_\text{train}$ will always be one -- this is not desirable for generated data as it indicates memorization of the training data. However, the oracle values of $W1_\text{test}$ and $\text{Cov}_\text{test}$ give an estimate of the best possible expected value that would be achieved with data from the correct distribution.

\begin{figure}[h]
    \centering
    \includegraphics[scale=0.5, trim={5 10 5 5}, clip]{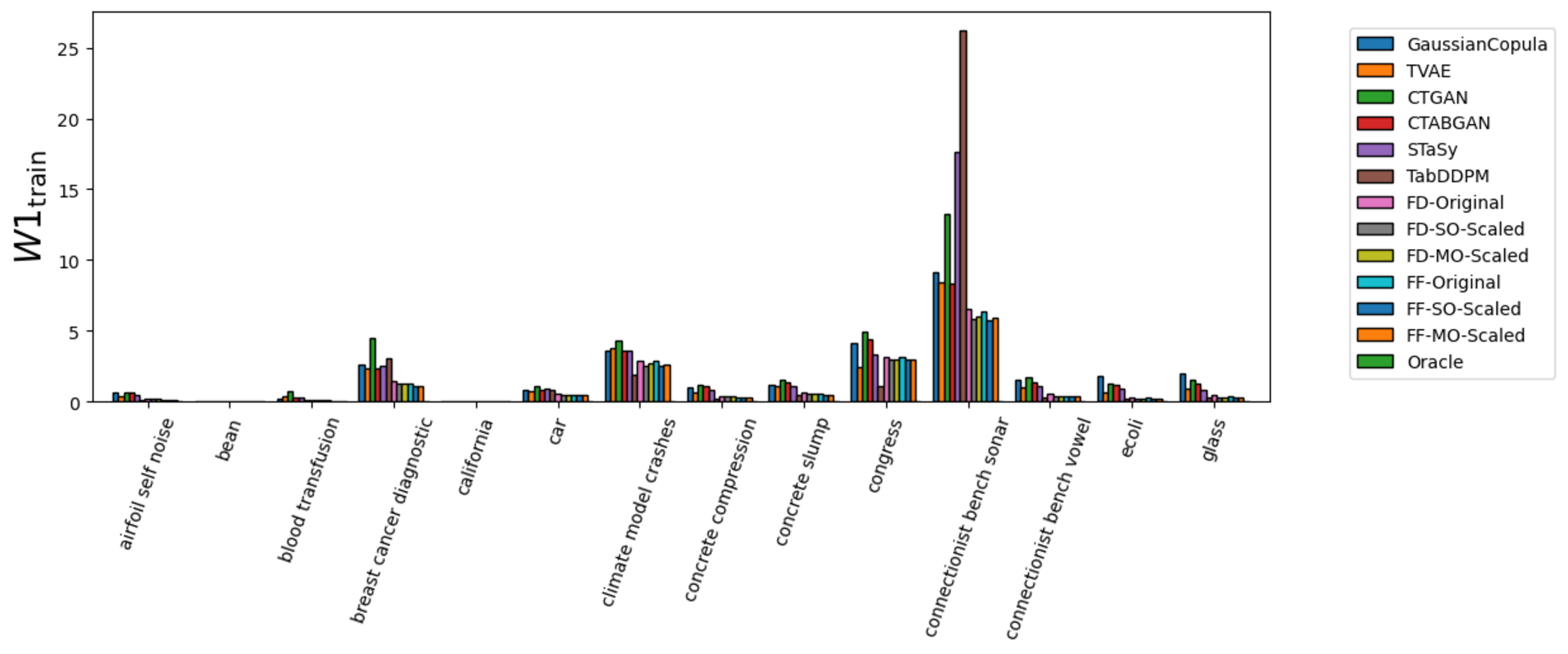}
\end{figure}

\begin{figure}[h]
    \centering
    \includegraphics[scale=0.5, trim={5 10 5 5}, clip]{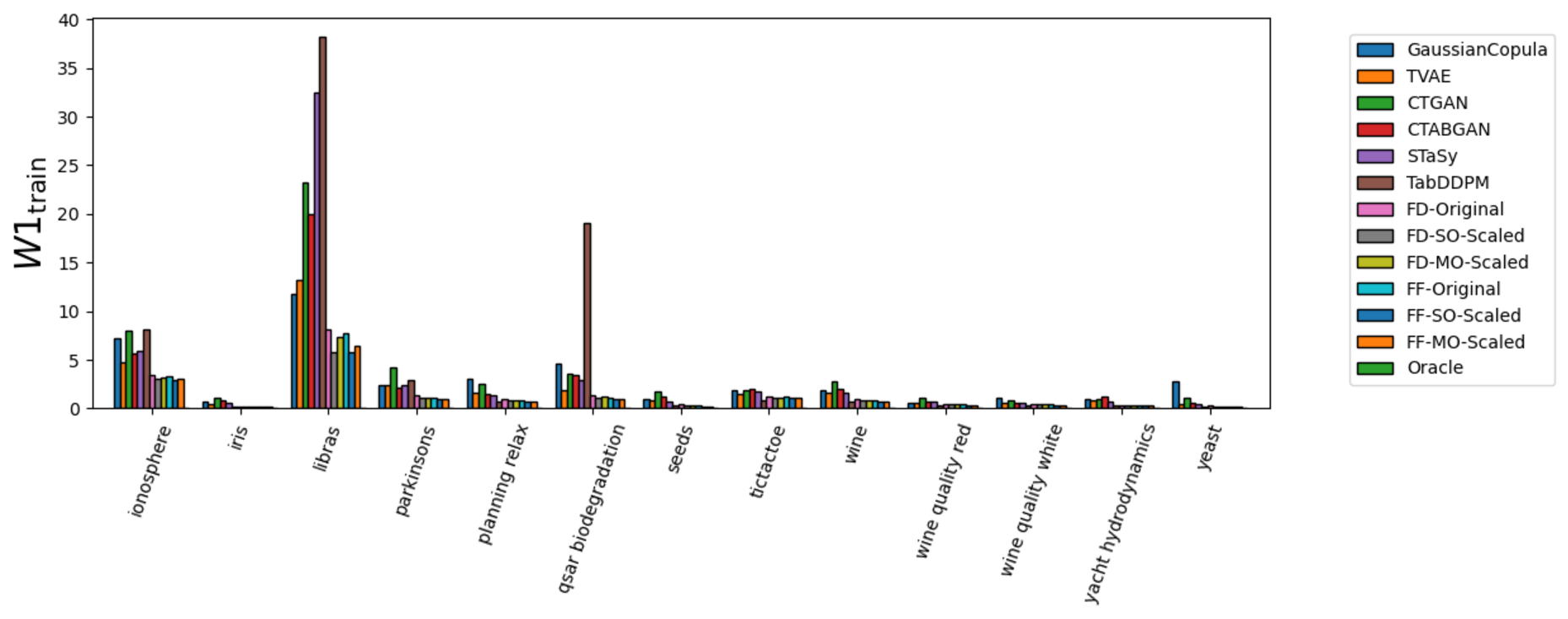}
\end{figure}

\begin{figure}[h]
    \centering
    \includegraphics[scale=0.5, trim={5 10 5 5}, clip]{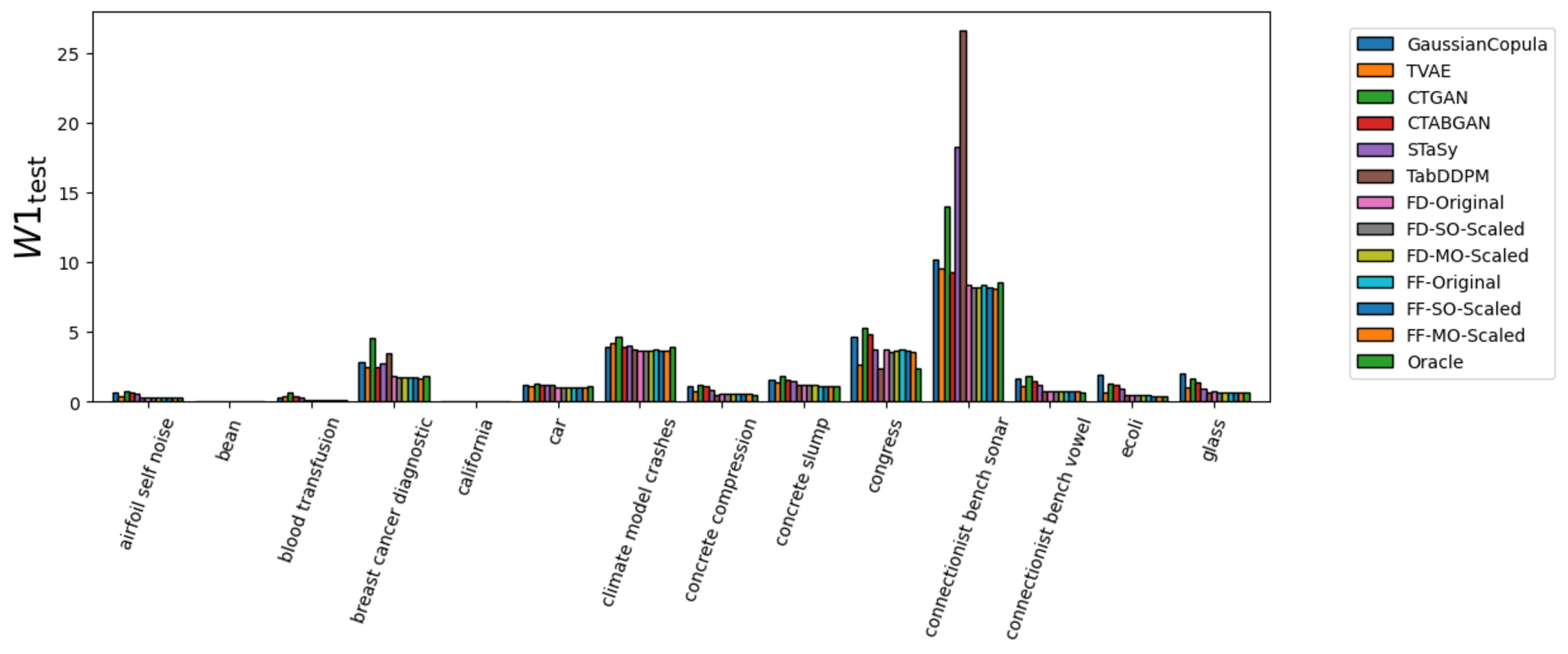}
\end{figure}

\begin{figure}[h]
    \centering
    \includegraphics[scale=0.5, trim={5 10 5 5}, clip]{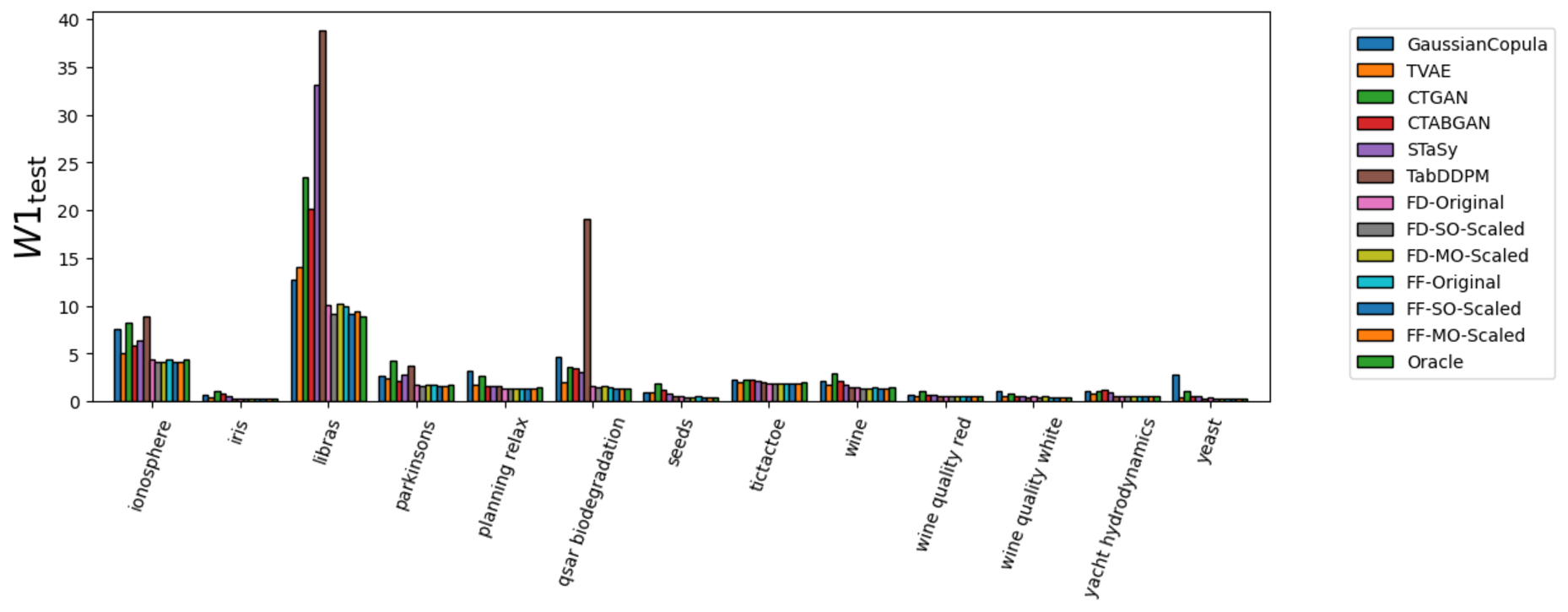}
\end{figure}

\begin{figure}[h]
    \centering
    \includegraphics[scale=0.5, trim={5 10 5 5}, clip]{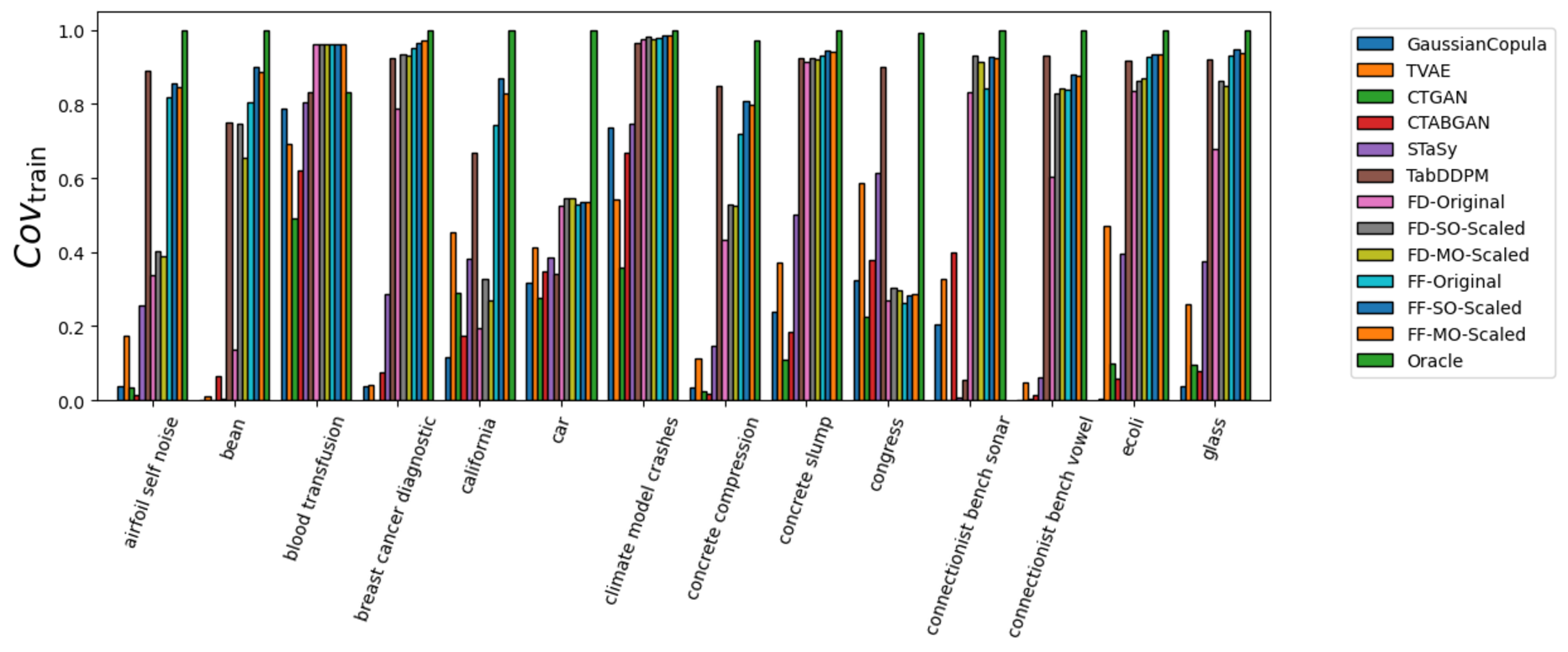}
\end{figure}

\begin{figure}[h]
    \centering
    \includegraphics[scale=0.5, trim={5 10 5 5}, clip]{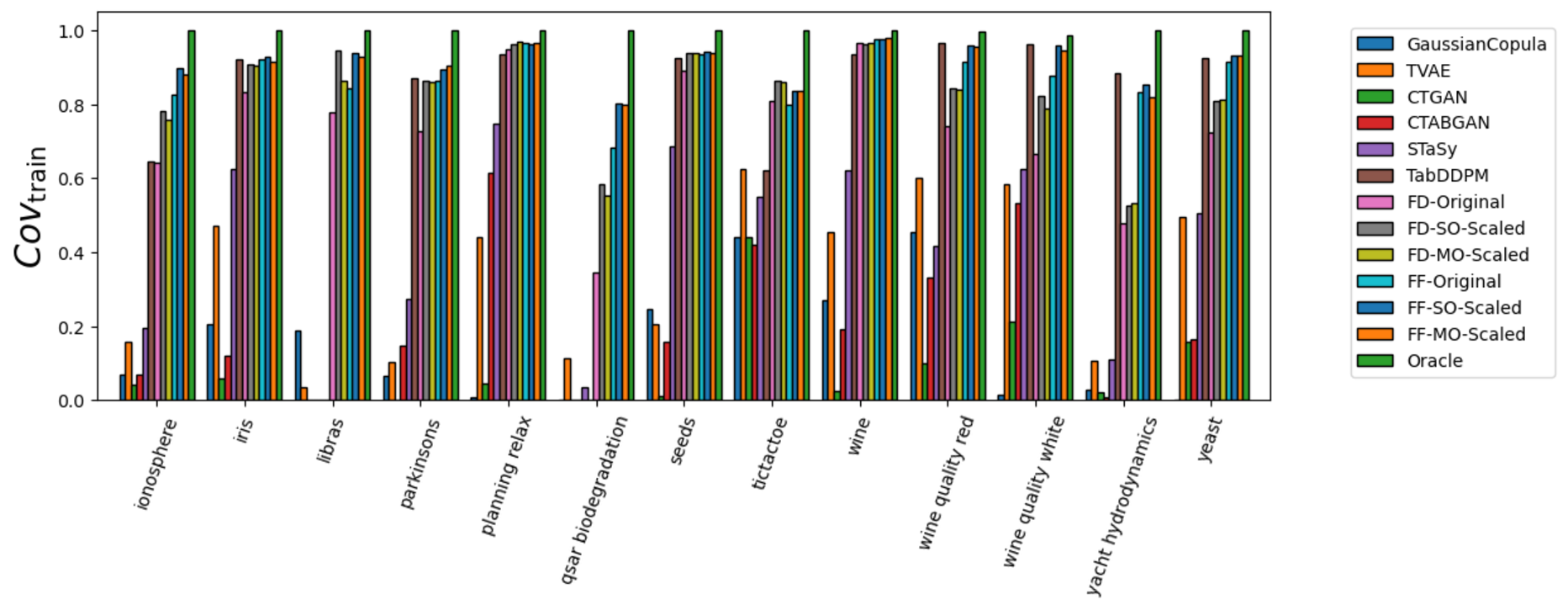}
\end{figure}

\begin{figure}[h]
    \centering
    \includegraphics[scale=0.5, trim={5 10 5 5}, clip]{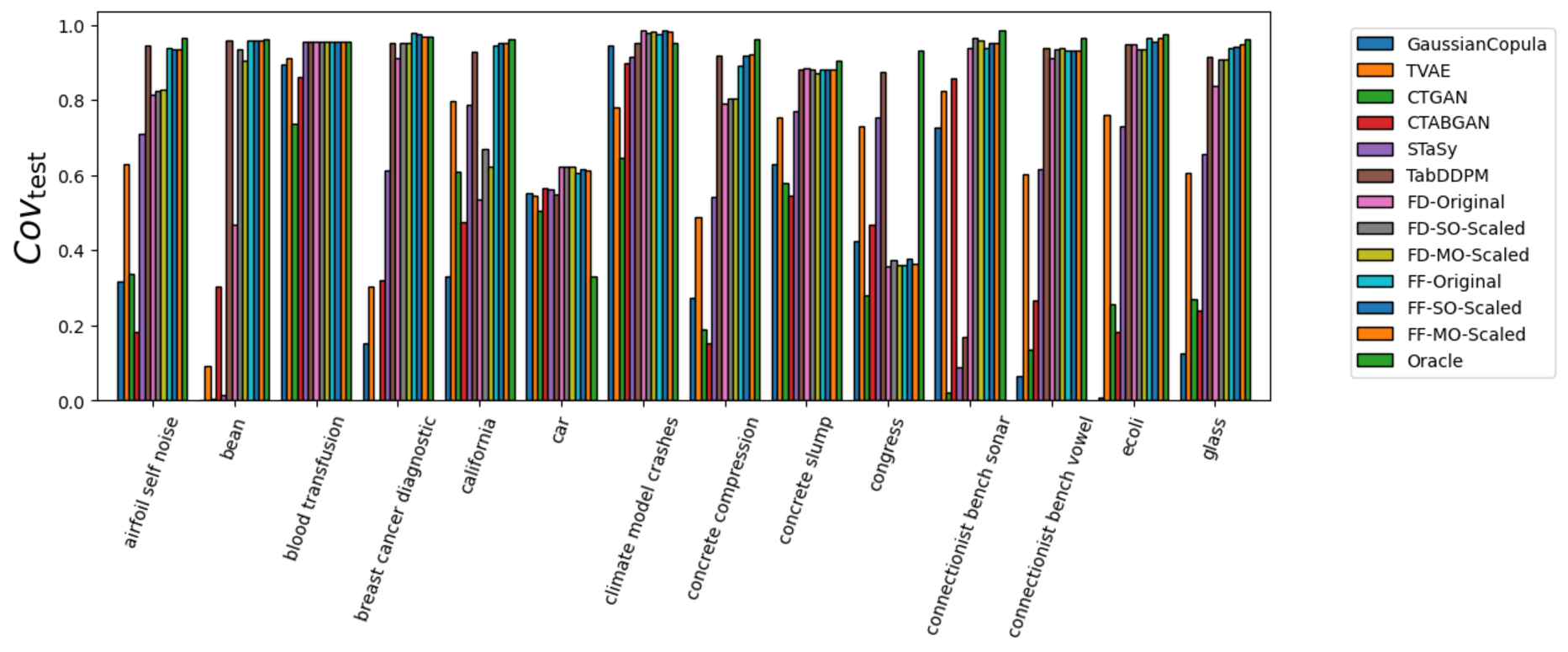}
\end{figure}

\begin{figure}[h]
    \centering
    \includegraphics[scale=0.5, trim={5 10 5 5}, clip]{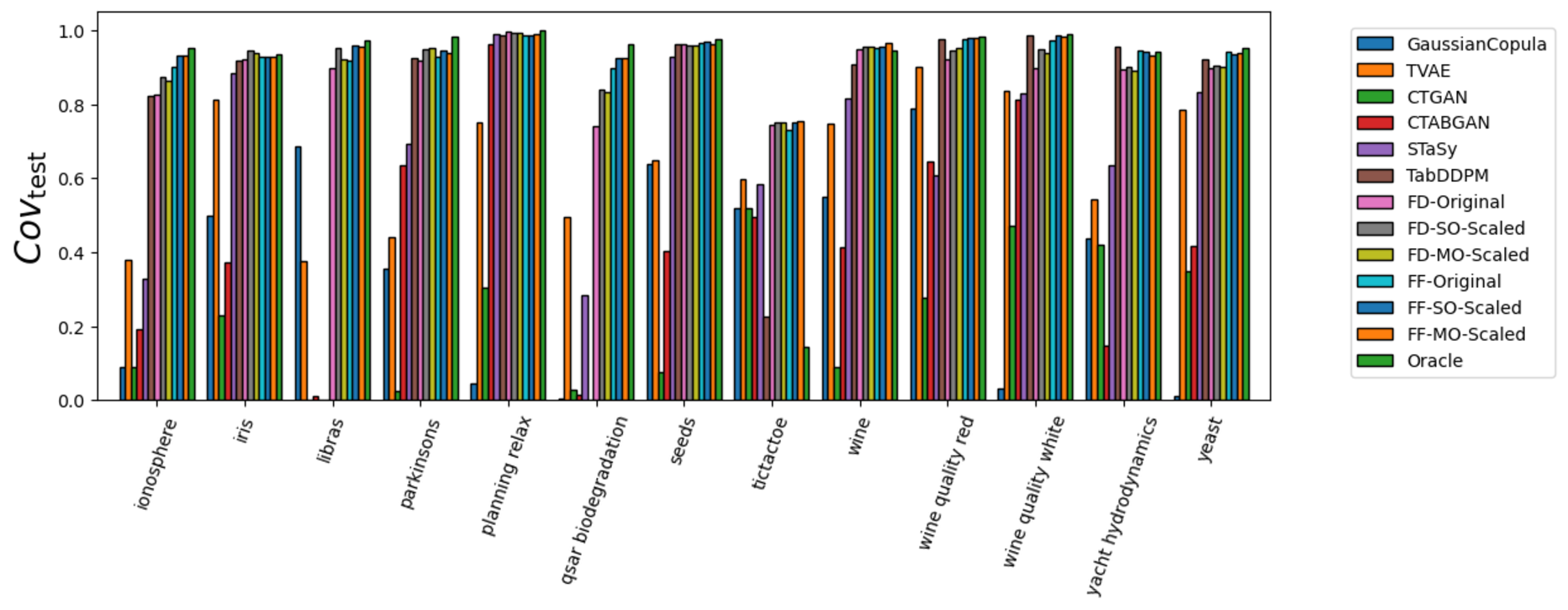}
\end{figure}

\begin{figure}[h]
    \centering
    \includegraphics[scale=0.5, trim={5 10 5 5}, clip]{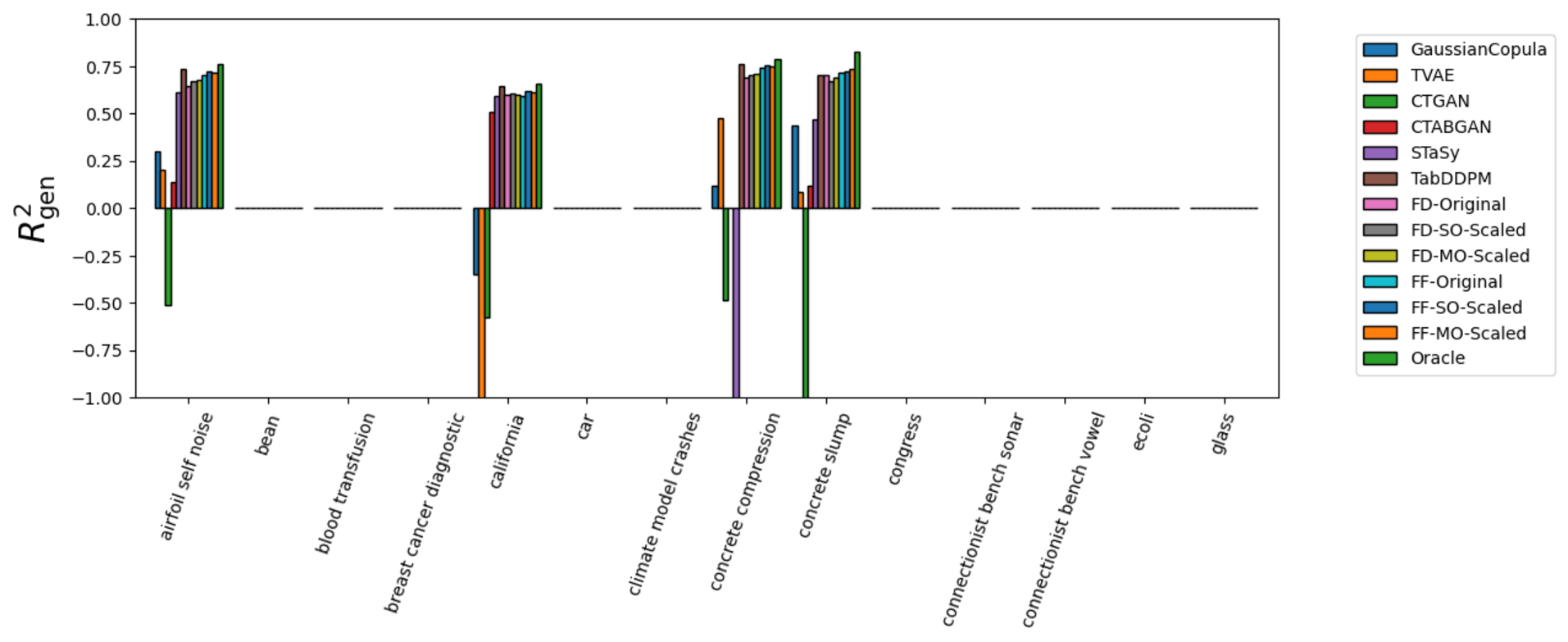}
\end{figure}

\begin{figure}[h]
    \centering
    \includegraphics[scale=0.5, trim={5 10 5 5}, clip]{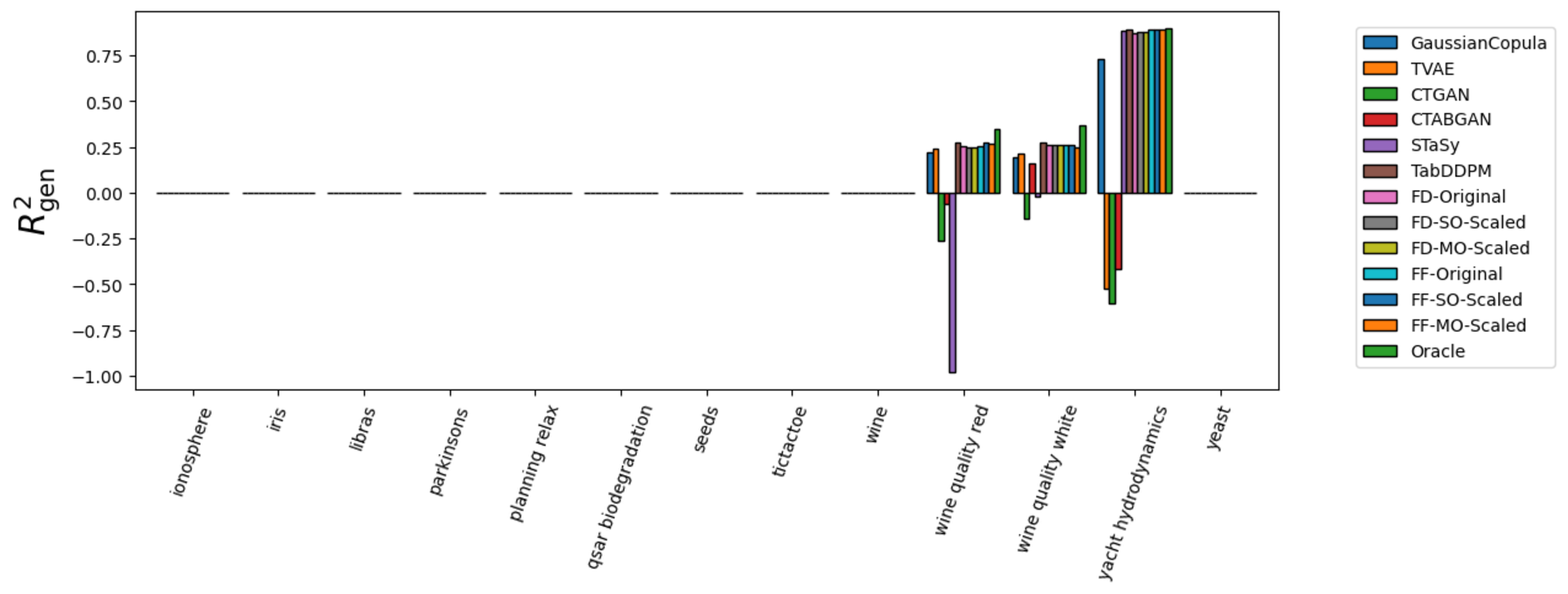}
\end{figure}

\begin{figure}[h]
    \centering
    \includegraphics[scale=0.5, trim={5 10 5 5}, clip]{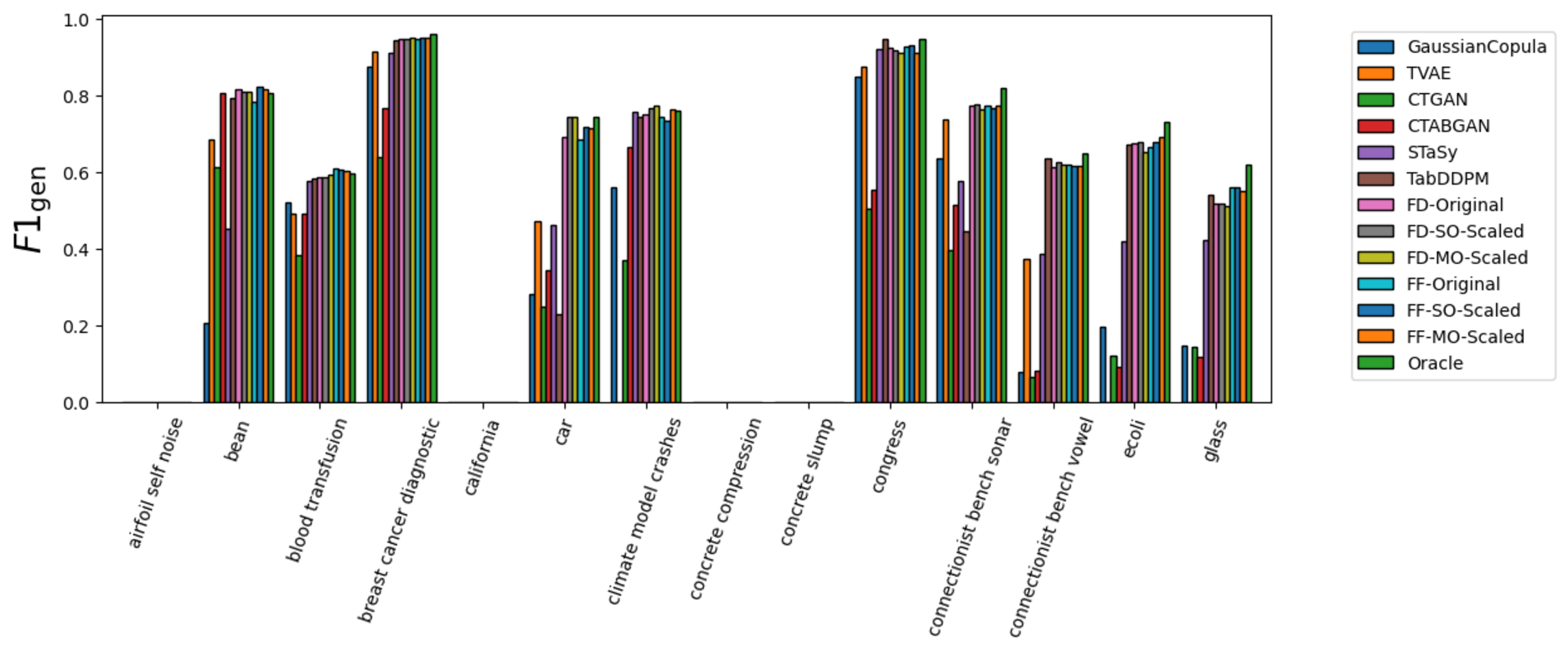}
\end{figure}

\begin{figure}[h]
    \centering
    \includegraphics[scale=0.5, trim={5 10 5 5}, clip]{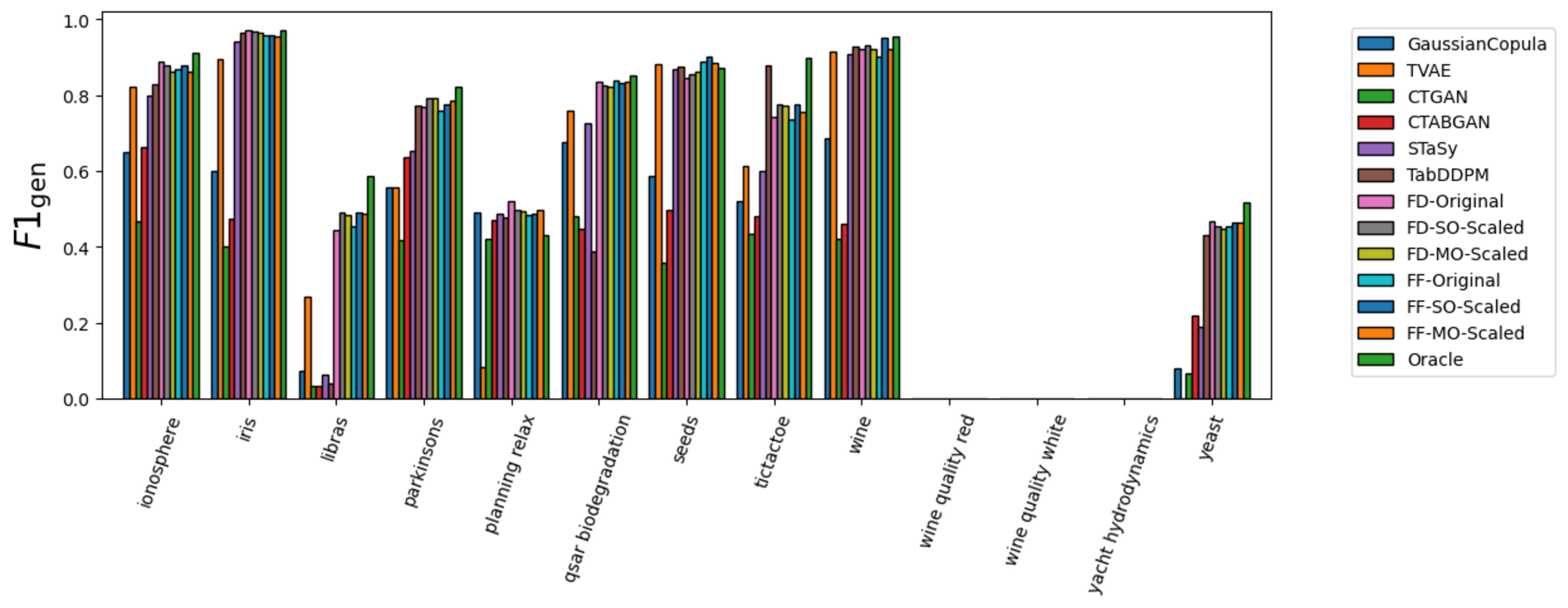}
\end{figure}

\begin{figure}[h]
    \centering
    \includegraphics[scale=0.5, trim={5 10 5 5}, clip]{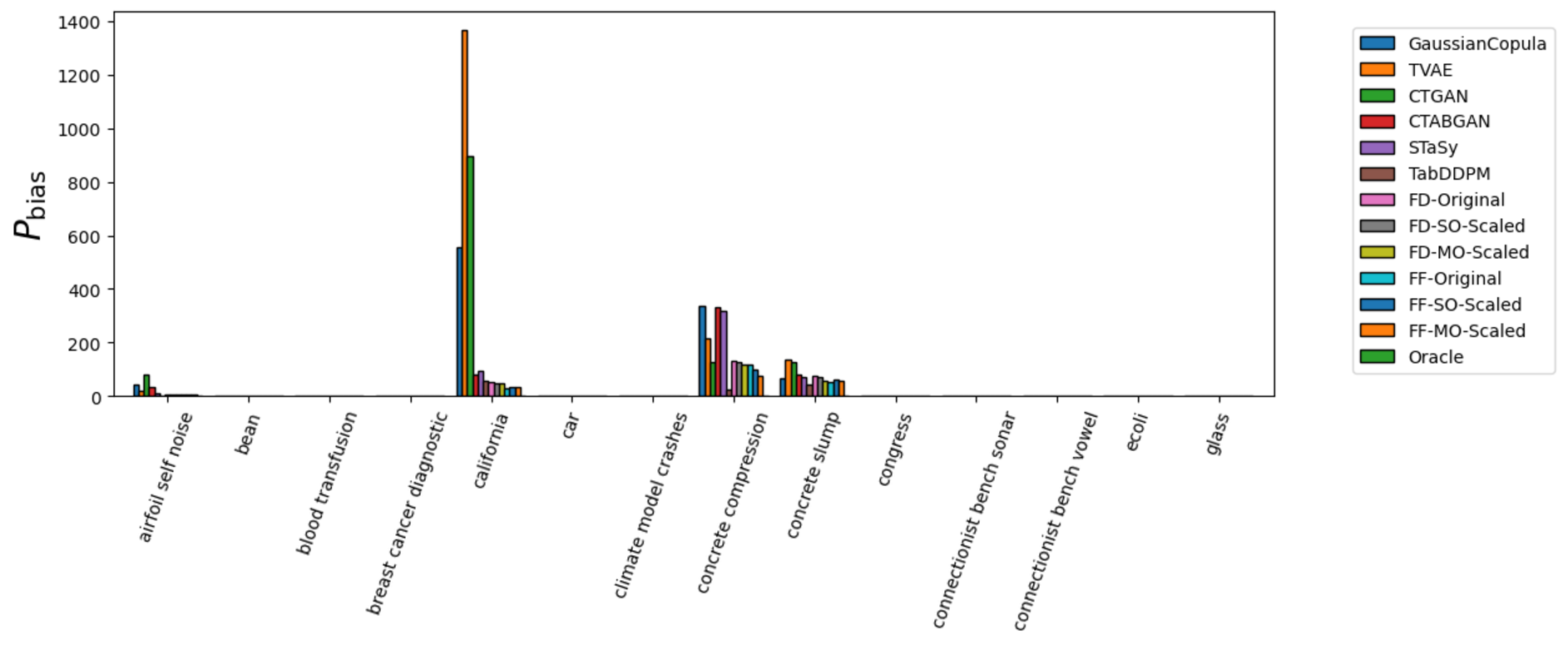}
\end{figure}

\begin{figure}[h]
    \centering
    \includegraphics[scale=0.5, trim={5 10 5 5}, clip]{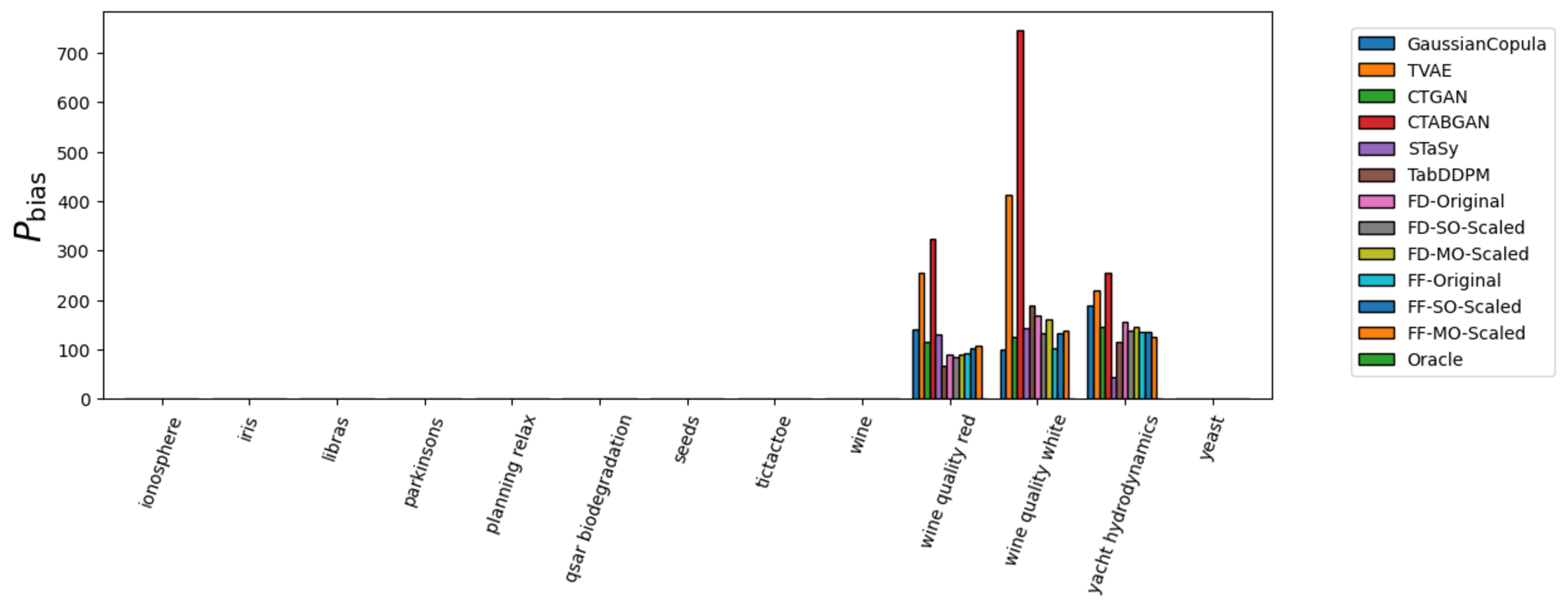}
\end{figure}

\begin{figure}[h]
    \centering
    \includegraphics[scale=0.5, trim={5 10 5 5}, clip]{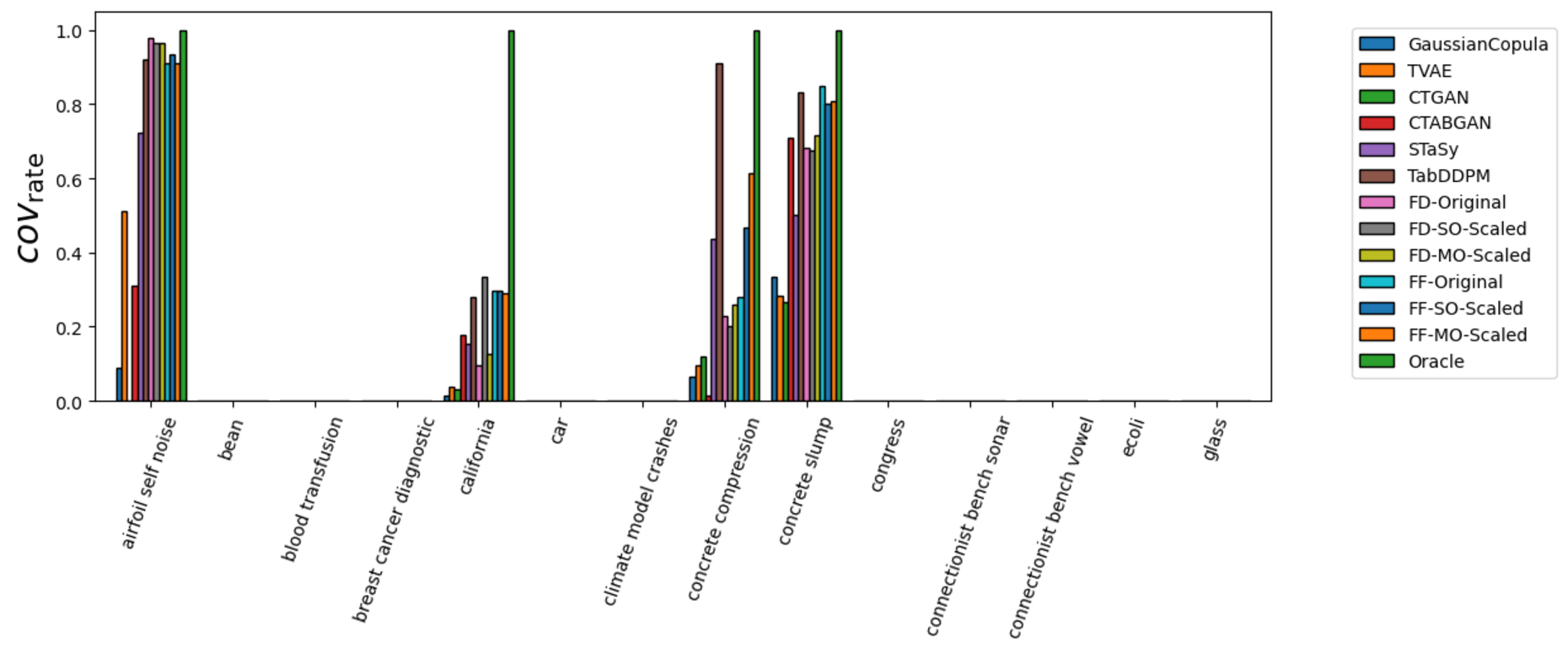}
\end{figure}

\begin{figure}[h]
    \centering
    \includegraphics[scale=0.5, trim={5 10 5 5}, clip]{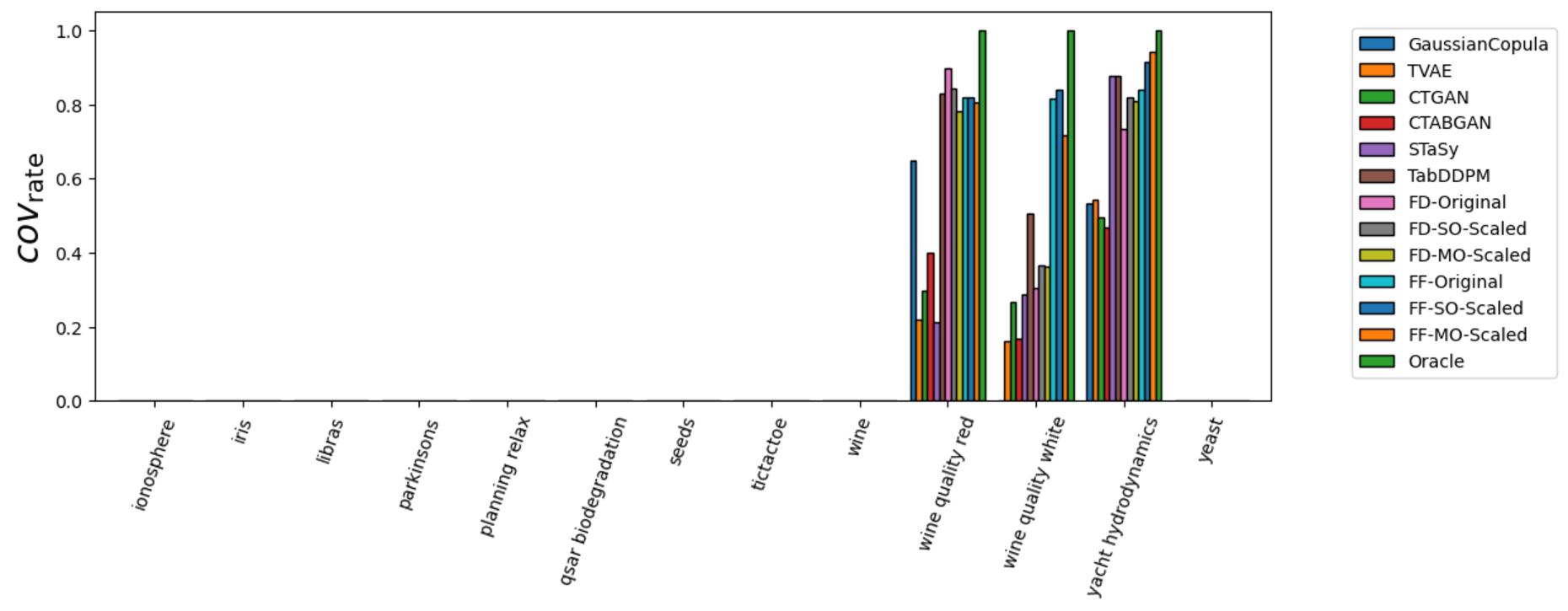}
\end{figure}


\end{document}